\documentclass[10pt,twocolumn,letterpaper]{article}

\usepackage{wacv}                       %

\usepackage{graphicx}
\usepackage{amsmath}
\usepackage{amssymb}
\usepackage{booktabs}
\usepackage{multirow}
\usepackage{makecell}
\usepackage{xcolor}
\usepackage{colortbl}
\usepackage{pifont}
\usepackage{capt-of}   %

\newcommand{\cags}{\textsc{cags}}
\newcommand{\cas}{\textsc{cas}}

\newcommand{\EM}{\ensuremath{\mathrm{EM}}}
\newcommand{\EMs}{\mathrm{EM}_{\mathrm{soft}}}
\newcommand{\dpr}{d^{\prime}}
\newcommand{\accat}[1]{\mathrm{acc}@#1}

\newcommand{\ctax}{C1}
\newcommand{\cattr}{C2}
\newcommand{\cspat}{C3}
\newcommand{\creid}{C4}

\newcommand{\cmark}{\ding{51}}
\newcommand{\xmark}{\ding{55}}
\newcommand{\pmark}{$\circ$}

\newcommand{\biovqa}{\textsc{Bio-VQA}} 
\newcommand{\best}[1]{\textbf{#1}}
\definecolor{rowtint}{RGB}{238,243,236}

\definecolor{cbest}{RGB}{176,190,232}   \definecolor{csnd}{RGB}{221,228,247}
\definecolor{cwst}{RGB}{242,178,178}    \definecolor{cswst}{RGB}{251,222,222}
\newcommand{\bst}[1]{\colorbox{cbest}{#1}}
\newcommand{\snd}[1]{\colorbox{csnd}{#1}}
\newcommand{\wst}[1]{\colorbox{cwst}{#1}}
\newcommand{\swst}[1]{\colorbox{cswst}{#1}}

\usepackage{xcolor}
\usepackage{orcidlink}
\definecolor{taskCountBG}{HTML}{0072B2}
\definecolor{taskOCRBG}{HTML}{B35900}
\definecolor{taskRelPosBG}{HTML}{00785A}
\newcommand{\hlbox}[2]{%
  {\setlength{\fboxsep}{1pt}%
   \raisebox{0pt}[0pt][0pt]{\colorbox{#1!60}{\vphantom{Ag}#2}}}%
}

\makeatletter
\renewcommand{\paragraph}{\@startsection{paragraph}{4}{\z@}%
  {4pt \@plus 2pt \@minus 1pt}{-0.5em}%
  {\normalfont\normalsize\bfseries}}
\makeatother

\newcommand{\dwn}[1]{\textcolor{red}{$\downarrow$}#1}
\newcommand{\upp}[1]{\textcolor{teal}{$\uparrow$}#1}

\definecolor{wacvblue}{rgb}{0.21,0.49,0.74}

\usepackage{siunitx}
\usepackage{newtxtext}
\usepackage[accsupp]{axessibility}
\usepackage{tabularx}
\usepackage{enumitem}
\hypersetup{breaklinks,colorlinks,allcolors=wacvblue}

\def\wacvPaperID{3034}
\def\confName{WACV}
\def\confYear{2027}

\title{How Finely Can Vision–Language Models See Biology? Bio-VQA: A Benchmark for Taxonomic, Morphological, Spatial and Individual Reasoning}

\author{%
S~M~Rayeed$^{*}$\,\orcidlink{0000-0001-7173-3429} \qquad
Qian~Ma\,\orcidlink{0009-0006-5194-7642} \qquad
Charles~V.~Stewart\,\orcidlink{0000-0001-6532-6675} \qquad
Yao~Ma$^{*}$\,\orcidlink{0000-0002-4985-8724}\\[0.2em]
Rensselaer Polytechnic Institute \qquad
{\small\texttt{\href{https://smrayeed.github.io/Bio-VQA/}{{smrayeed.github.io/Bio-VQA}}}} \qquad
{\small\texttt{\{rayees, may13\}@rpi.edu}}
}

\begin{document}
\AddToShipoutPicture*{\AtPageLowerLeft{\raisebox{20pt}{\makebox[\paperwidth]{\parbox{0.85\paperwidth}{\centering\scriptsize
\copyright\,2027 IEEE. Personal use of this material is permitted. Permission from IEEE must be obtained for all other uses, in any current or future media, including reprinting/republishing this material for advertising or promotional purposes, creating new collective works, for resale or redistribution to servers or lists, or reuse of any copyrighted component of this work in other works.}}}}}

\twocolumn[{%
  \renewcommand\twocolumn[1][]{#1}%
  \maketitle
  \begin{center}
    \includegraphics[width=0.9\textwidth]{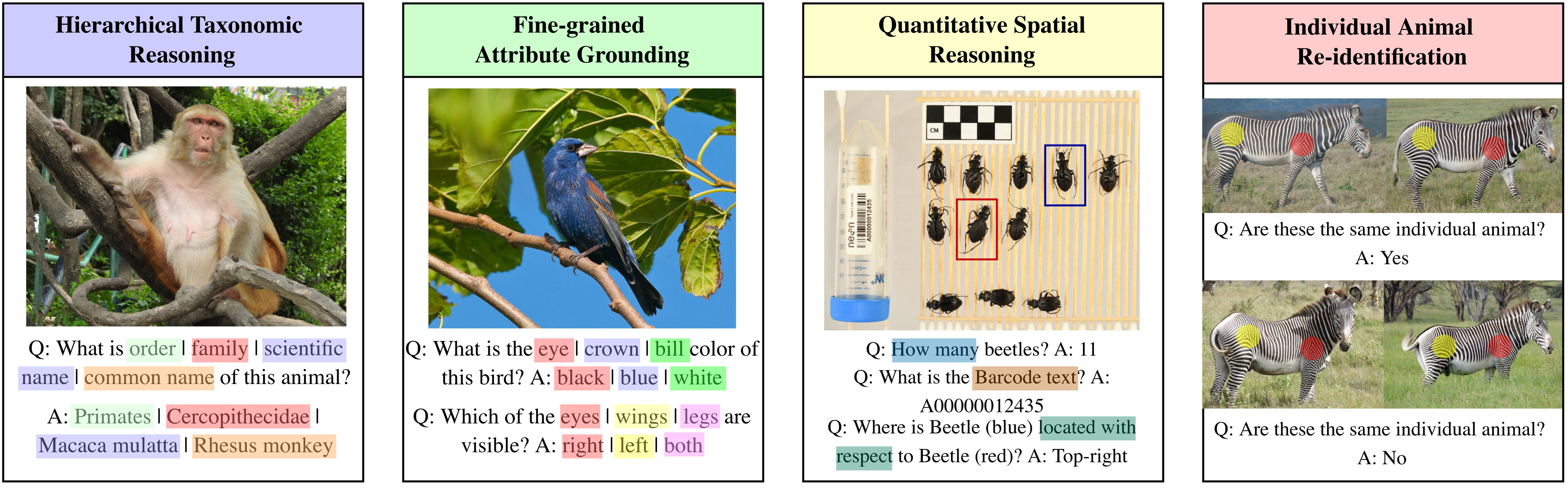}
    \captionsetup{hypcap=false}
    \captionof{figure}[Overview of \biovqa.]{\biovqa~is a benchmark on biological domain, consisting of four components: \textbf{(a)}~hierarchical taxonomic reasoning, \textbf{(b)}~fine-grained attribute grounding, \textbf{(c)}~quantitative spatial reasoning (3 of 8 tasks shown: \protect\hlbox{taskCountBG}{counting}, \protect\hlbox{taskOCRBG}{OCR}, \protect\hlbox{taskRelPosBG}{relative position}), \textbf{(d)}~re-identification.
    }
    \label{fig:teaser}
  \end{center}%
  \vspace{0.1em}
}]

\begin{abstract}
In biodiversity monitoring, a single image raises several questions: which taxon, how many, which trait, and whether the individual is seen before. In practice, each is answered by a specialist model, or by field experts. Vision--language models (VLMs) offer a way past this by answering questions from images. However, their ability is largely untested in biological domains, as existing works focus mostly on species identification within a limited range of taxonomic groups. To address this, we present \biovqa, the largest biological visual question answering (VQA) benchmark, and the only one to span four complementary aspects of biological visual understanding: hierarchical taxonomic reasoning, fine-grained attribute grounding, quantitative spatial reasoning, and individual re-identification. \biovqa~contains 1.86M questions over 217K images extracted from 11 public datasets. Prior works mark an answer right or wrong, we extend that to graded scoring, so a near-miss counts for more than a distant one. Evaluating 18 open-weight models from 3.75B to 38.4B parameters, we find that general visual capabilities do not transfer to biological organisms: the best model counts animals exactly on 41.7\% of images with 2 to 10 individuals, less than half its accuracy on everyday objects at the same counts. Taxonomic accuracy falls to one sixth, from 0.72 at phylum to 0.12 at scientific name, and scaling helps mostly at coarse ranks. The gap to a domain specialist widens accordingly, from 0.731 against BioCLIP-2's 0.874 at class ($1.2\times$) to 0.269 against 0.544 at scientific name ($2\times$). On trait size measurement, 15 of 18 models do no better than an image-blind baseline that predicts one fixed number for every specimen. And on re-identification, 17 models fall below chance under adversarial conditions. Under zero-shot evaluation, these failures persist across a tenfold range of model size. By offering a comprehensive benchmark, we hope \biovqa~becomes a standard test of how well VLM capabilities transfer to biological domain. 
\end{abstract}

\section{Introduction}
\label{sec:intro}

Images are the primary record for modern biodiversity
monitoring~\cite{gbif2024,inaturalist, east2025imaging}, but the large volume of data from camera traps and digitization efforts has created an annotation bottleneck that specialist taxonomic models and experts cannot clear~\cite{stevens2024bioclip,vendrow2024inquire, rayeed2026carabids}. VLMs solve this by answering questions from images, and are being applied across domains like medical imaging, autonomous driving and document understanding. Yet we lack the benchmarks to measure how well VLMs answer biological questions~\cite{miao2024frontiers,fabian2023wildmatch,rayeed2025beetleverse,rayeed2025finegrained}. Current evaluations fall short in four ways: \textbf{\emph{Narrow taxonomic coverage.}} They restrict themselves to a few animal groups. VLM4Bio~\cite{vlm4bio2024}, the closest prior work in scope, draws its questions from three animal groups, while others focus only on crop species~\cite{plantexpertvqa2025,agrocot2025} or a single habitat at a small scale~\cite{jediqa2024}. \textbf{\emph{Narrow question scope.}} They focus heavily on basic taxonomic naming, ignoring major biological aspects like morphological traits, spatial quantification or re-identification. \textbf{\emph{Fixed scoring.}} They mark an answer either right or wrong, which penalizes a near-miss in the correct genus exactly the same as a complete miss in the wrong kingdom. \textbf{\emph{Weaker baselines.}} They report random chance for multiple-choice items, and compares one task against a specialist~\cite{vlm4bio2024}. Neither bound is available per task, and random chance is weaker than an image-blind predictor that exploits answer distribution. A weaker baseline understates the floor and inflates the models' scores. These shortfalls leave a large part of VLMs' biological visual reasoning capabilities unmeasured.

In \biovqa, we address the gaps with 24 tasks across four components, graded scoring, and a baseline for each task. The components are: \textbf{\emph{hierarchical taxonomic reasoning}~(\ctax{})}, \textbf{\emph{fine-grained attribute grounding}~(\cattr{})}, \textbf{\emph{quantitative spatial reasoning}~(\cspat{})}, \textbf{\emph{individual re-identification}~(\creid{})}. \ctax{} probes eight ranks across 10 organismal groups drawn from TreeOfLife-200M~\cite{stevens2025bioclip2}. \cattr{} tests visual traits, attributes and anatomic localization on CUB-200-2011~\cite{wah2011caltech}. \cspat{} decomposes spatial understanding into atomic tasks. \creid{} tests whether two photographs show the same individual. Our contributions are:
\begin{enumerate}
  \itemsep0.1em
  \item \textbf{The Benchmark:} \biovqa~is the largest biological VQA dataset to date, with 1.86M questions across 217K images, and we uniquely combine taxonomic, morphological, spatial, and re-identification tasks.
  \item \textbf{Biology-Aware Metrics:} Besides strict exact matches, we report taxonomy-aware soft match and tree-distance scoring for \ctax{}; chance-adjusted graded score for \cattr{}, validated against three human annotators and an 11-configuration sensitivity sweep; bounded count agreement for \cspat{}; and signal-detection scoring for \creid{}.
  \item \textbf{Rigorous Baselines:} We bound every task below by a computed image-blind floor, or above by a measured specialist ceiling, placing generalist performance between guessing and genuine domain expertise.
  \item \textbf{Cross-Component Analysis:} We run correlation study across all components to test whether a model's ability on one transfers to another. For the strongest models it does, lower down the rank, it does not.
\end{enumerate}
 
The results from 18 open-weight VLMs reveal a domain gap. VLMs handle coarse categories well but fail on fine-grained tasks, and within the models we test, larger models mostly improve at coarse ranks they already know. Outside taxonomy, they fail to count conspecifics exactly or to distinguish individuals when appearance is controlled. In short, biological imagery requires a strict granularity of discrimination that general pretraining corpora do not reward.

\section{Related Work}
\label{sec:related}

\paragraph{General VLM benchmarks.}
Multi-task suites like MMBench~\cite{liu2024mmbench}, SEED-Bench~\cite{li2023seedbench}, BLINK~\cite{fu2024blink}, MMVP~\cite{tong2024eyes}, and CV-Bench~\cite{tong2024cambrian} evaluate VLMs by breaking overall accuracy into per-ability scores (\eg. counting, spatial relations, depth, OCR), and AVA-Bench~\cite{avabench2026} extends the decomposition to 14 atomic visual abilities (\eg. absolute distance, relative distance). These suites establish that decomposing one score into per-ability components exposes failures that an aggregate score would mask. Biological organisms appear in these suites, but incidentally rather than as a focus. So we adopt their per-ability component decomposition and apply it to a structured set of biological tasks.

\paragraph{Biological VQA benchmarks.}
The closest reference of our work is VLM4Bio~\cite{vlm4bio2024}. It evaluates 12 models on 469K QA pairs across 30K images of fishes, birds, and butterflies. We extend this line of work along three axes, not just in scale. First, we cover ten organismal groups, testing underrepresented branches like arachnids and bryophytes. Second, we apply open-ended and multiple-choice questions on the exact same items to measure the recognition-recall gap. Finally, we explore tasks on spatial reasoning and individual re-identification, which are untested in prior biological VQA benchmarks. Other domain benchmarks are also limited in scope: J-EDI QA~\cite{jediqa2024} focuses on 100 deep-sea images, PlantExpertVQA~\cite{plantexpertvqa2025} is limited to crop diseases, and AgroCoT~\cite{agrocot2025} targets agricultural chain-of-thought reasoning. Other biodiversity evaluations either lack imagery~\cite{iucnllm2026} or focus on retrieval rather than VQA~\cite{vendrow2024inquire}. \Cref{tab:related} compares \biovqa~against these prior works.

\begin{table}[t]
  \centering
  \footnotesize
  \setlength{\tabcolsep}{3pt}
  \begin{tabular}{clrrccccc}
    \toprule
    & \multirow{2}{*}{Benchmark}
      & \multirow{2}{*}{QA} & \multirow{2}{*}{Img.}
      & \multirow{2}{*}{\makecell{Org.\\grp.}}
      & \multicolumn{4}{c}{Question type} \\
    \cmidrule(lr){6-9}
    & & & & & Tax & Att & Spa & Ind \\
    \midrule
    \multirow{4}{*}{\rotatebox[origin=c]{90}{General}}
      & MMBench~\cite{liu2024mmbench}   & 3.0K & 3.0K & --- & \xmark & \pmark & \pmark & \xmark \\
      & BLINK~\cite{fu2024blink}        & 3.8K & 7.4K & --- & \xmark & \pmark & \cmark & \xmark \\
      & CV-Bench~\cite{tong2024cambrian}& 2.6K & 2.6K & --- & \xmark & \pmark & \cmark & \xmark \\
      & AVA-Bench~\cite{avabench2026}   & 218K & 218K & --- & \xmark & \cmark & \cmark & \xmark \\
    \midrule
    \multirow{4}{*}{\rotatebox[origin=c]{90}{Biology}}
      & J-EDI QA~\cite{jediqa2024}               & 0.1K & 0.1K & 1  & \cmark & \xmark & \xmark & \xmark \\
      & PlantExpertVQA~\cite{plantexpertvqa2025} & 765K & 151K & 1  & \pmark & \cmark & \xmark & \xmark \\
      & VLM4Bio~\cite{vlm4bio2024}               & 469K & 30K  & 3  & \cmark & \cmark & \xmark & \xmark \\
    \cmidrule(l{2pt}r{2pt}){2-9}
    \rowcolor{rowtint}
      & \biovqa~(ours) & \best{1.86M} & \best{217K} & \best{10} & \cmark & \cmark & \cmark & \cmark \\
    \bottomrule
  \end{tabular}
  \caption{Comparison with prior benchmarks. Columns report the number of question-answer pairs (QA), images (Img.) and organismal groups (Org. grp.); and whether they cover question types: taxonomic (Tax), attribute (Att), spatial (Spa) and individual identity (Ind). \cmark{} means the question type is a primary task family, \pmark{} that it appears incidentally, and \xmark{} that it is absent.}
  \label{tab:related}
\end{table}

\paragraph{Domain specialists.} For taxonomic identification, contrastive biological encoders are the strongest references: BioCLIP~\cite{stevens2024bioclip} and BioCLIP-2~\cite{stevens2025bioclip2} are trained at tree-of-life scale, and we use the latter as the specialist baseline for \ctax{}. Individual re-identification has a longer, separate line of work, from hand-crafted local features~\cite{crall2013hotspotter} to learned embeddings~\cite{moskvyak2019pie,nepovinnykh2024alfreid,he2021transreid} and multi-species descriptors~\cite{cermak2024wildlifedatasets,otarashvili2024miewid}; all are purpose-built, and no survey of the area~\cite{animalreid2026} reports VLMs. \creid{} asks whether VLMs can re-identify individuals under controlled difficulty levels, scored with signal detection theory~\cite{green1966signal,macmillan2004detection} against a specialist, MegaDescriptor.

\paragraph{Evaluation methodology.}
Exact-match scoring penalizes near-misses (such as identifying correct genus with wrong species) identically to complete failures (wrong family). Hierarchical scoring with Wu--Palmer~\cite{wu1994verb} and Lin~\cite{lin1998information} similarity is standard in taxonomy-aware evaluation~\cite{ging2024openended}, and chance correction via Cohen's $\kappa$~\cite{cohen1960coefficient} is standard in annotation agreement. However, prior VQA benchmarks do not combine these. In \biovqa, we combine them and report every graded metric along with the exact match.

\section{Curation of \textsc{\biovqa}}
\label{sec:biovqa}

\paragraph{Design principles.}
We curate \biovqa~on three principles. First, we base our questions on actual biological workflows: naming organisms, describing traits, counting and measuring specimens, and re-identifying individuals. Second, we generate a question only where an existing annotation supplies the ground truth, so every answer can be checked against its source. Third, within each component, we stratify questions into difficulty tiers along an axis that component defines (\eg common vs rare taxa on \ctax{}). The four subsections below give the reasoning and the construction for each component.

\subsection{\ctax{}: Hierarchical taxonomic reasoning}
\label{sec:c1}

\begin{figure}[t]
  \centering
  \fbox{\includegraphics[width=0.97\columnwidth]{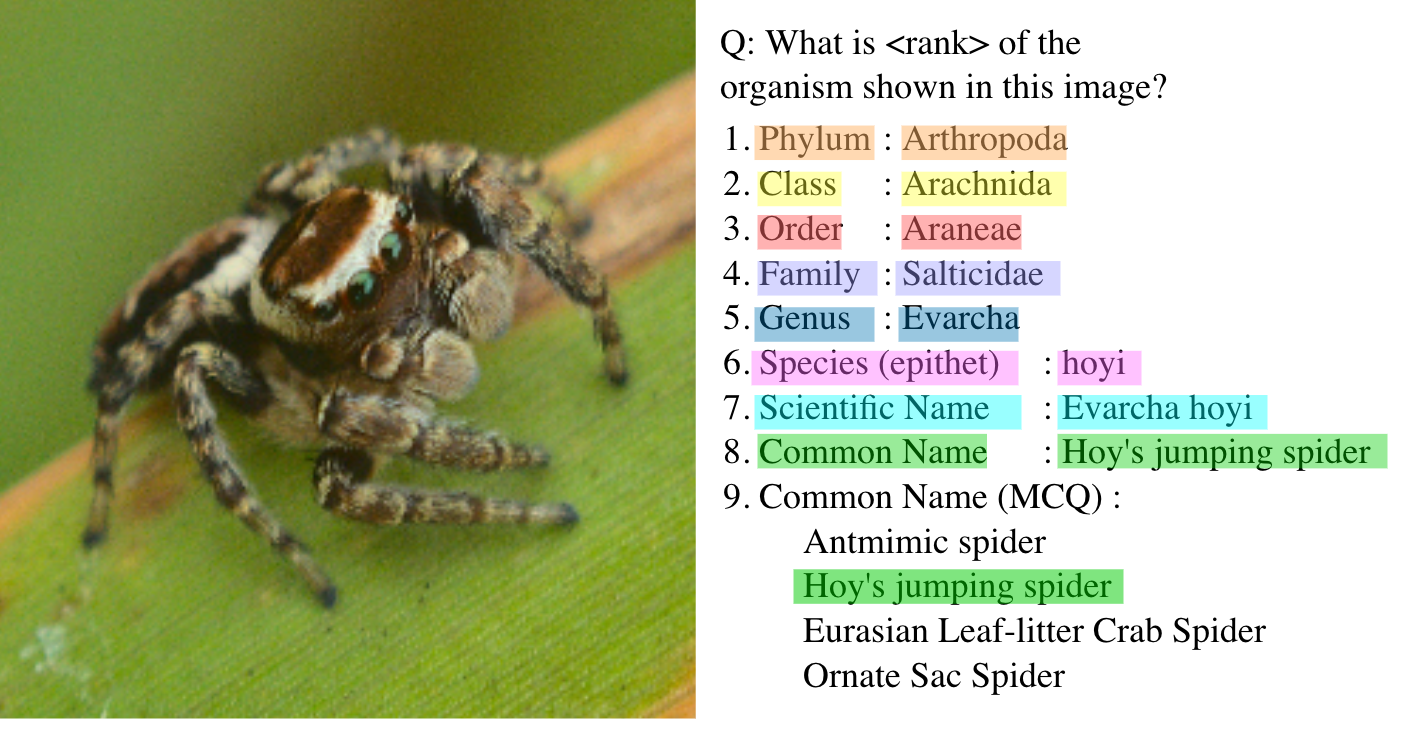}}
  \caption{\ctax{} asks one image for its identity at eight ranks, from phylum down to common name, and repeats common name as four-way multiple choice. Species is the epithet and scientific name is the binomial, so we score the two separately.}
  \label{fig:c1}
\end{figure}

Taxonomic identification is the most common biological visual task, and contrastive encoders like BioCLIPs~\cite{stevens2024bioclip,stevens2025bioclip2} and BioTrove-CLIP~\cite{biotrove2024} are specialists at this. We ask \emph{how far a generalist VLM stands from the specialists on the same task}. Current works evaluate species identification on a limited taxa~\cite{vlm4bio2024,plantexpertvqa2025}. We extend this along two axes: (a) we ask the question at eight ranks rather than at species alone, (b) we ask it of ten organismal groups; six of which do not appear in prior works.

We sample a 100K subset of TreeOfLife-200M~\cite{stevens2025bioclip2} spanning \num{626} families, \num{2311} genera (6$\times$VLM4Bio) and \num{3403} species (4$\times$VLM4Bio). The source is long-tailed, so a random sample would be dominated by a few abundant taxa. We therefore select 10K images from each organismal group, on three constraints: (a) complete lineage from phylum to species and common names, so that no question is generated without a ground truth; (b) a minimum resolution threshold; (c) a per-species cap, to avoid abundant species within a group. Then, we ask each image for its phylum, class, order, family, genus, species, scientific name and common name\footnote{Reptiles are asked at seven ranks. GBIF reclassified Squamata from order to class in 2022, and left the order-level empty.} (~\cref{fig:c1}). We test two framings on the common name questions, open-ended and four-way multiple choice, on the same image and same target. This tests what a model recalls from what it recognizes once given the options, and evaluate the recognition-recall gap. For the multiple choice, we draw the three distractors uniformly from the taxa of the same class as the target, at whichever rank the question asks.

Then we assign difficulty tier labels on questions by how common the target taxon is in TreeOfLife-200M. At each rank, we take the top third of taxa by image count as Easy, the bottom third as Hard, and the rest as Medium; so the tiers separate taxa a model has seen often from ones it has barely seen. Finally, we address two properties of the source labels. The GBIF vernacular records are not filtered by language, and contains Swedish, Danish and Norwegian names in the common-name field. A model answering with the correct English name would be scored wrong, so we resolve every non-English common name to English against the iNaturalist and Wikidata APIs~\cite{inaturalist,wikidata}. The second is synonymy: a cougar answered as \emph{Felis concolor} is the same animal as \emph{Puma concolor}. We build a synonym map from GBIF Backbone~\cite{gbif2024} and accept any synonym of the target.

\subsection{\cattr{}: Fine-grained attribute grounding}
\label{sec:c2}

\begin{figure*}[!t]
  \centering
  \includegraphics[width=0.9\textwidth]{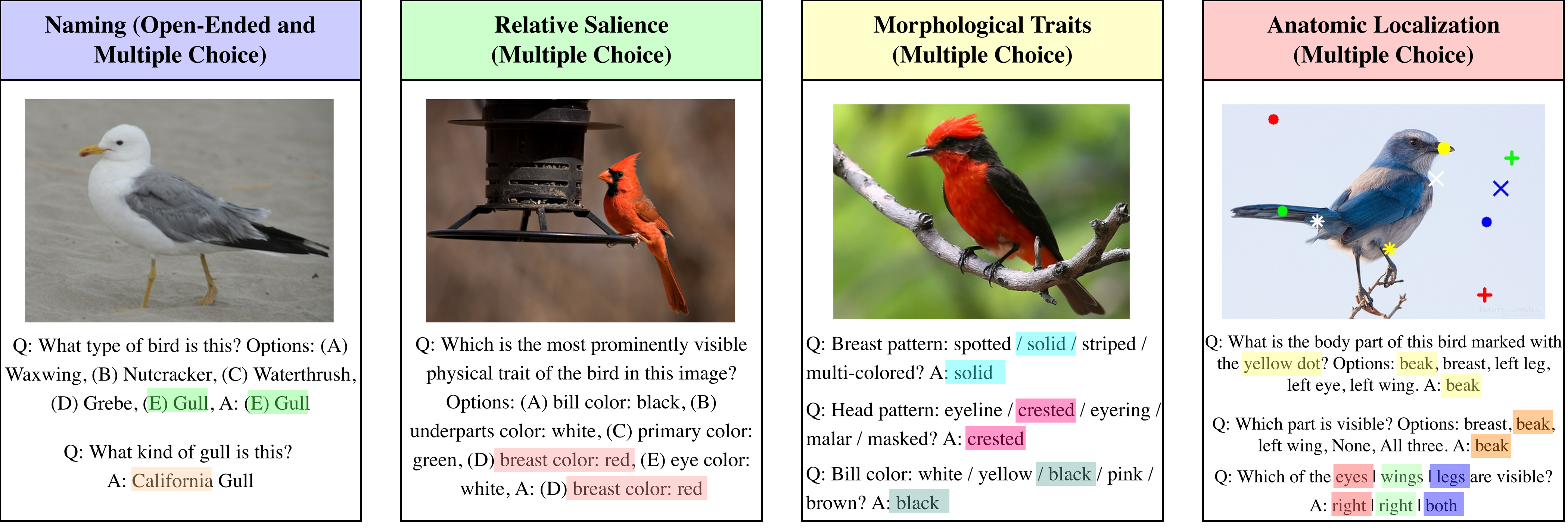}
  \caption{\cattr{} poses four categories of question, with one example item each. We ask naming both open-ended and as multiple choice on the same image. Anatomical localization is visually prompted, so the question refers to a rendered marker rather than to a named part.}
  \label{fig:c2}
\end{figure*}

Fine-grained traits play a key role in trait-based ecology, and recognizing them differs from naming a taxon. In \cattr{}, we evaluate \emph{how well VLMs recognize these fine-grained visual attributes}. We use CUB-200-2011~\cite{wah2011caltech}, a well-known bird dataset that provides 312 binary attributes per image, each named as a part, a property, and a value (\emph{has\_wing\_color::brown}). Evaluating a model on 312 sparse yes/no questions is uninformative, because most flags are absent and a constant predictor that answers \emph{no} scores well. So we invert the mapping: we combine a part and a property (\emph{wing color}) into a single question and use the marked value as the ground truth (\emph{brown}).

This gives us four types of questions: morphological attributes, relative salience, anatomical localization and naming (\cref{fig:c2}). For multiple choice, we draw distractors from the other values attested for that same part and property, and exclude any value that is itself true of the image, so every option is a real trait of some bird and exactly one is correct. A model must therefore ground its answer rather than use the process of elimination. We use the annotator certainty on each label to select and filter candidate traits, and report that handling in the appendix. For anatomical localization, we render a marker onto the image and ask \emph{which body part it is}, and add distractors using two controls: one distractor shares the marker's color but not its shape, another shares its shape but not its color. Neither cue identifies the marker alone, so a model must bind color and shape together to answer. Because we render all the \num{11788} source images, the component comprises \num{23576} images and \num{824104} questions.

\subsection{\cspat{}: Quantitative spatial reasoning}
\label{sec:c3}

\begin{figure*}[t]
  \centering
  \includegraphics[width=\textwidth]{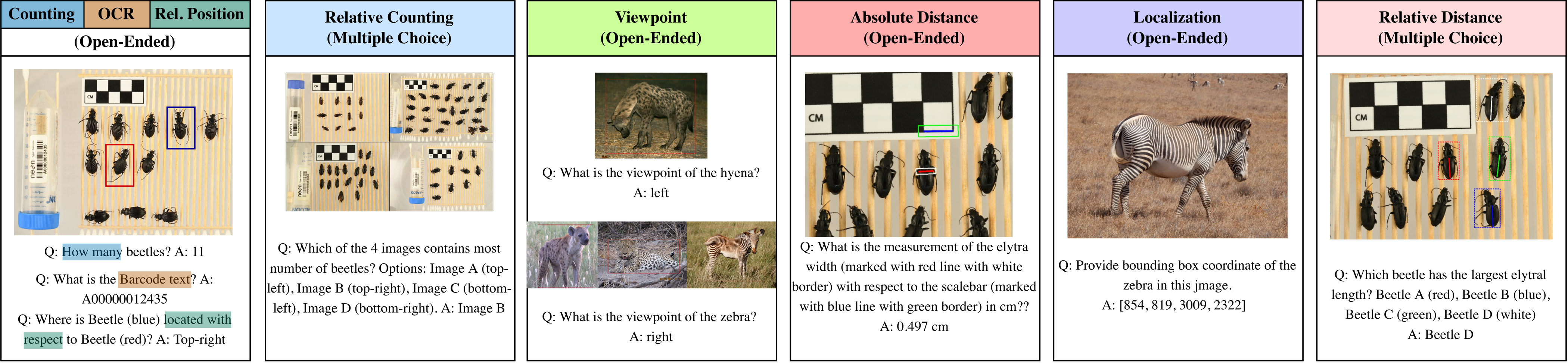}
  \caption{\cspat{}'s eight spatial tasks, with one example item each, its answer format and its ground truth. The first panel asks counting, OCR and relative position of the same beetle plate. The relative distance and counting tasks are multiple-choice, rest are open-ended.}
  \label{fig:c3}
\end{figure*}

Spatial reasoning is widely studied in general VQA benchmarks~\cite{tallyqa2019,paiss2023countbench,liu2023vsr,fu2024blink,chen2024spatialvlm}, and it also covers the most common operations in biodiversity fieldwork. In \cspat{}, we ask \emph{how well VLMs perform these tasks on biological subjects}. Following the task-decomposition approach of AVA-Bench~\cite{avabench2026}, we break spatial reasoning into seven tasks (\cref{fig:c3}). The decomposition attributes a failure to a specific ability rather than to spatial reasoning in general. We add optical character recognition (OCR) as an eighth task and a negative control. Locating the label is a spatial task, but the answer depends on transcribing characters.

We use five ecological datasets (\cref{tab:stats}). We select these for multi-organism images, which tasks like counting and relative position require, and which prior biological VQA works do not hold~\cite{vlm4bio2024,plantexpertvqa2025}. We aggregate them into ten 500-image subsets, on two constraints. First, we sample using count stratification for a balanced mix of sparse and dense scenes. Second, we discard images whose annotation completeness we cannot verify, since a partial annotation penalizes a correct count.

Next, we evaluate each task only on subsets where the ground truth exists. Absolute and relative distance run on the beetle dataset, as that is the only one with recorded trait and scalebar measurements. Relative position also runs on beetles, because the specimens lie flat in a lattice at one focal depth, so \emph{top-right} and \emph{left} are genuine two-dimensional relations and not perspective artifacts. Localization and viewpoint need a single dominant subject, otherwise a model has to locate the animal before it can answer and the tasks stop being isolated. So we leverage the wildlife catalogs from \creid{} (\cref{sec:c4}), where annotations explicitly record the bounding box and the viewing angle. In total, \cspat{} holds \num{132504} questions over \num{78139} images across eight tasks.

\subsection{\creid{}: Individual re-identification}
\label{sec:c4}

Animal re-identification is a highly specialized task, handled by purpose-built matching models~\cite{crall2013hotspotter,cermak2024wildlifedatasets,otarashvili2024miewid} and field experts. We are not aware of a benchmark that asks it of a generalist VLM. We are interested to know if \emph{VLM focuses on identity (\eg, zebra's stripe pattern, leopard's spots) or appearance similarity}, on re-id tasks. We test this with a minimal probe. We place two images side by side and ask whether these are the same individual or not (\cref{fig:teaser}d).

We build pairs from five wildlife catalogs (\cref{tab:stats}), and set their difficulty using DINOv2~\cite{oquab2024dinov2} embeddings. \textbf{Easy} pairs an animal with its nearest match, often frames taken seconds apart, and pairs different animals that look nothing alike, so appearance and identity agree. \textbf{Random} draws both without using appearance, so appearance carries no signal. \textbf{Hard} reverses easy: different individuals are drawn from each other's nearest neighbors, and the same individual from its furthest, so appearance and identity disagree. We retrieve within viewpoint groups, since viewpoint dominates the embedding space. Each catalog contributes 1000 balanced pairs per condition, and we sample same-individual pairs first, since the catalogs hold fewer repeat sightings than distinct animals. The hard condition is adversarial by construction. In these pairs the two animals share same pose, lighting, background and coat tone. They differ only in the markings that identify them. Therefore a model that decides by appearance yields the wrong answer.

\begin{table}[t]
  \centering
  \footnotesize
  \setlength{\tabcolsep}{3pt}
  \resizebox{\columnwidth}{!}{%
  \begin{tabular}{llrrrr}
    \toprule
    & Component & Tasks & Data & Images & Questions \\
    \midrule
    \ctax{}  & Hierarchical taxonomic reasoning &  9 & 1 & 100{,}000 & 890{,}000 \\
    \cattr{} & Fine-grained attribute grounding &  4 & 1 &  23{,}576 & 824{,}104 \\
    \cspat{} & Quantitative spatial reasoning   &  8 & {8$^\dagger$} &  78{,}139 & 132{,}504 \\
    \creid{} & Individual re-identification    &  3 & 5 &  15{,}000 &  15{,}000 \\
    \midrule
    \multicolumn{2}{l}{\textbf{Total}} & \textbf{24} & \textbf{11$^\dagger$}
      & \textbf{216{,}715} & \textbf{1{,}861{,}608} \\
    \bottomrule
  \end{tabular}}
  \caption{Bio-VQA statistics: We draw \ctax{} from TreeOfLife-200M~\cite{stevens2025bioclip2}; \cattr{} from CUB-200-2011~\cite{wah2011caltech}; \cspat{} from WAID~\cite{waid2023}, Bees~\cite{boxesonbees}, 2018-NEON-Beetles~\cite{neonbeetles}, DePondFi~\cite{depondfi}, GZCD~\cite{gzcd}; and \creid{} from WhaleSharkID~\cite{whalesharkid}, BelugaID~\cite{belugaid}, HyenaID~\cite{hyenaid}, LeopardID~\cite{leopardid}, GZCD~\cite{gzcd}. $^\dagger$C3 uses WhaleSharkID, HyenaID, LeopardID, GZCD on two tasks: so the total count becomes 11.}
  \label{tab:stats}
\end{table}

\section{Evaluation Pipeline}
\label{sec:pipeline}

\subsection{Models and Decoding}

We evaluate 18 open-weight models from 3.75B to 38.4B parameters: Qwen2.5/3-VL~\cite{qwen25vl,qwen3vl}, InternVL3/3.5~\cite{internvl3,internvl35}, Gemma-3/4~\cite{gemma3,gemma4}, Molmo~2~\cite{molmo2}, Pixtral~\cite{pixtral}, Phi-3.5-Vision~\cite{phi35}, Llama-3.2-Vision~\cite{llama32}, and two LLaVA-NeXT variants~\cite{llavanext}. Two 30B models are mixture-of-expert variants that activate 3B parameters per token. All run under one protocol: greedy decoding, the same question and answer-format instruction for every model within a question family~\cite{sclar2024}, no chain-of-thought, one attempt per item, and one scorer per component. We report parse and letter-text mismatch rates to separate formatting failures from knowledge errors~\cite{liu2024mmbench}. The format instructions show an example answer; the supplement lists the exact strings.

\subsection{Scoring}
Prior biological VQA scores by strict match~\cite{vlm4bio2024,plantexpertvqa2025,jediqa2024}, which credits a near miss and a random guess alike. We add one graded metric per component.

\paragraph{\ctax{}: quantifying taxonomic error.}
On a warbler, exact match scores \emph{Turdidae} and \emph{Fungi} alike, although one is near miss and one is wrong by far. To partially credit the near miss, we measure taxonomic distance. With $\ell = \mathrm{LCA}(\hat{y}, y)$ the deepest clade containing both answers, Wu--Palmer similarity~\cite{wu1994verb} scores the depth of that shared ancestor:
\begin{equation}
  \mathrm{wup}(\hat{y},y) = \frac{2\,\mathrm{depth}(\ell)}
       {\mathrm{depth}(\hat{y}) + \mathrm{depth}(y)}
  \label{eq:wup}
\end{equation}
We also report soft exact match, which accepts any synonym of the gold answer from the accepted-name relations of the GBIF Backbone~\cite{gbif2024}, and give Lin similarity~\cite{lin1998information} and severity in the appendix.

\paragraph{\cattr{}: near misses and uneven answer spaces.}
On a brown wing, \emph{red} is closer than \emph{green}, and the option count runs from 3 for bill length to 15 for color, so a raw graded score confounds ability with option count. Our Chance-Adjusted Graded Score (\cags{}) corrects both. Let $s(\hat{y},y) \in [0,1]$ be the similarity between two answers and $\mathbb{E}_0[s]$ its value under a random guess:
\begin{equation}
  \cags{} = \frac{\mathbb{E}\big[s(\hat{y}, y)\big] - \mathbb{E}_{0}[s]}
                 {1 - \mathbb{E}_{0}[s]}
  \label{eq:cags}
\end{equation}
\cags{} is 0 at chance and 1 at perfect for any option count. We take $s$ from CIEDE2000 distance for colors and annotation co-occurrence for parts, and validate it in \cref{sec:exp-c2}.

\paragraph{\cspat{}: grading error and establishing baselines.}
Predicting 14 beetles for 15 beats predicting 20, so we score counting by relative error, macro-averaged over corpora whose count distributions differ:
\begin{equation}
  \cas{}(\hat{n}, n) = \max\!\Big(0,\; 1 - \tfrac{|\hat{n} - n|}{n}\Big)
  \label{eq:cas}
\end{equation}
Localization uses mean IoU and the categorical tasks balanced accuracy, with macro-averaged agreement for viewpoint. Every task also carries a floor $\phi$, the best score an image-blind rule reaches over the modal count, a per-type constant and majority class. We report lift $(S-\phi)/(1-\phi)$, which is 0 when a model matches that rule and negative when it does worse.

\paragraph{\creid{}: same or not same.}
\creid{} is a verification task~\cite{lfw2007}: balanced same and not-same pairs, scored by accuracy. A VLM returns a decision rather than a similarity score, so it sits at one operating point where accuracy conflates discrimination with bias, and answering \emph{yes} to every pair scores 50\%. We separate the two using signal detection theory~\cite{green1966signal,macmillan2004detection}, writing $H$ for the \emph{yes} rate on same pairs and $F$ for the rate on different ones:
\begin{equation}
  \dpr = z(H) - z(F), \qquad
  c = -\tfrac{1}{2}\big[z(H) + z(F)\big]
  \label{eq:dprime}
\end{equation}
where $z$ is the inverse normal CDF, $\dpr$ is discriminability and $c$ is model bias. We apply the log-linear correction first, since $z$ diverges when $H$ or $F$ reaches 0 or 1~\cite{hautus1995}, and report 95\% bootstrap intervals for accuracy over 2000 resamples per cell. Balanced accuracy, Matthews correlation and Cohen's $\kappa$ are reported in the appendix.

\section{Experiment Results}
\label{sec:exp}

\begin{table*}[t]
  \centering
  \scriptsize
  \setlength{\tabcolsep}{1.1pt}
  \setlength{\fboxsep}{1.2pt}
  \renewcommand{\arraystretch}{1.08}
  \begin{tabular*}{\textwidth}{@{\extracolsep{\fill}} rl cccccccc cccc cccccccc cc @{}}
    \toprule
    \multirow{2}{*}{\#} & \multirow{2}{*}{Model}
      & \multicolumn{8}{c}{\ctax{}~~(Exact match)}
      & \multicolumn{4}{c}{\cattr{}~~(\cags{})}
      & \multicolumn{8}{c}{\cspat{}}
      & \multicolumn{2}{c}{\creid{}} \\
    \cmidrule(lr){3-10}\cmidrule(lr){11-14}\cmidrule(lr){15-22}\cmidrule(lr){23-24}
    & & phy & cls & ord & fam & gen & spc & sci & com
      & sal & mrp & loc & nam
      & cnt & rcnt & adst & rdst & lcl & rpos & view & ocr
      & $\dpr$ & acc \\
    \midrule
    \rowcolor{rowtint}
    & BioCLIP-2 (tol)$^{\ast}$ & .925 & .874 & .821 & .775 & .691 & .545 & .544 & .500
      & --- & --- & --- & --- & --- & --- & --- & --- & --- & --- & --- & --- & --- & --- \\
    \rowcolor{rowtint}
    & MegaDescriptor$^{\ast\ast}$ & --- & --- & --- & --- & --- & --- & --- & ---
      & --- & --- & --- & --- & --- & --- & --- & --- & --- & --- & --- & --- & .733 & .683 \\
    \midrule
     1 & Qwen3-VL-32B         & .831 & .615 & .712 & \bst{.612} & \bst{.444} & \snd{.066} & \snd{.244} & \snd{.078} & \snd{.584} & \snd{.644} & \bst{.475} & \bst{.731} & .789 & .555 & \bst{.622} & \bst{.443} & \bst{.866} & \bst{.550} & \snd{.427} & .900 & \bst{.696} & \bst{.636} \\
     2 & Qwen3-VL-30B-A3B     & .864 & \bst{.731} & \bst{.757} & \bst{.612} & .417 & \bst{.072} & .221 & \bst{.107} & .558 & .631 & .453 & .658 & \snd{.795} & .483 & \snd{.528} & \snd{.423} & .861 & .355 & .391 & \bst{.979} & .518 & .602 \\
     3 & InternVL3-38B        & \bst{.911} & .683 & .638 & .484 & .355 & .054 & .189 & .044 & .562 & \bst{.645} & .413 & .669 & \bst{.799} & \bst{.567} & .367 & .363 & .744 & .205 & .373 & .105 & .463 & .581 \\
     4 & Gemma-4-12B          & .880 & .686 & .578 & .363 & .183 & .021 & .065 & .043 & \bst{.592} & .621 & .431 & .374 & .753 & \snd{.563} & .385 & \bst{.443} & .589 & \snd{.495} & .331 & .345 & .351 & .567 \\
     5 & Qwen3-VL-8B          & \snd{.887} & .680 & \snd{.720} & \snd{.589} & .406 & .046 & .202 & .065 & .531 & .624 & .441 & .643 & .739 & .414 & .303 & .345 & \snd{.862} & .421 & .373 & .711 & .492 & .572 \\
     6 & InternVL3.5-38B      & .746 & .636 & .624 & .363 & .271 & .038 & .128 & .025 & .559 & .638 & .375 & .536 & .765 & .504 & .503 & .360 & .412 & .219 & .340 & .192 & .258 & .542 \\
     7 & Molmo2-8B            & .823 & .563 & .469 & .286 & .137 & .021 & .038 & .016 & .503 & .643 & \snd{.474} & .330 & .623 & .416 & .290 & .358 & \swst{.055} & .457 & \bst{.483} & .692 & .491 & .578 \\
     8 & Qwen2.5-VL-7B        & .763 & .504 & .673 & .541 & \snd{.420} & .047 & \bst{.269} & .062 & .541 & .580 & .356 & \snd{.713} & .724 & .410 & .334 & .337 & .308 & .213 & .395 & \snd{.916} & .442 & .541 \\
     9 & InternVL3.5-30B-A3B  & .486 & .469 & .545 & .310 & .240 & .034 & .118 & .024 & .537 & .609 & .350 & .539 & .727 & .489 & .378 & .333 & .289 & .252 & .309 & .188 & .568 & \snd{.612} \\
    10 & InternVL3-8B         & .827 & .497 & .478 & .322 & .262 & .032 & .124 & .020 & .508 & .584 & .346 & .581 & .738 & .320 & .297 & .304 & .246 & .203 & .335 & .314 & \snd{.684} & .601 \\
    11 & InternVL3.5-8B       & .819 & .531 & .421 & .266 & .191 & .023 & .077 & .015 & .495 & .603 & .372 & .457 & .747 & .408 & .263 & .299 & .236 & .224 & .264 & .460 & .236 & .544 \\
    12 & Gemma-3-12B          & .790 & \snd{.688} & .626 & .429 & .225 & .048 & .093 & .047 & .466 & .551 & .410 & .511 & .606 & .415 & .352 & .280 & .329 & .145 & .235 & .352 & .120 & \swst{.505} \\
    13 & Qwen2.5-VL-3B        & \swst{.416} & .457 & .477 & .454 & .358 & .040 & .206 & .064 & .427 & .500 & .389 & .661 & .640 & .321 & \wst{.194} & .300 & .307 & \wst{.110} & .230 & .904 & .655 & .559 \\
    14 & Pixtral-12B          & .688 & .501 & .530 & .349 & .187 & .029 & .079 & .027 & .447 & .495 & .320 & .465 & \swst{.599} & .352 & .307 & .304 & .229 & .266 & .246 & .065 & \swst{.119} & .519 \\
    15 & LLaVA-NeXT-Vicuna    & .662 & \wst{.343} & \swst{.290} & \wst{.100} & \wst{.028} & \wst{.003} & \wst{.015} & \wst{.003} & .457 & \wst{.454} & .364 & \wst{.198} & \wst{.510} & \wst{.250} & .217 & \wst{.250} & .701 & \swst{.125} & \swst{.187} & \wst{.000} & \wst{.000} & \wst{.500} \\
    16 & Phi-3.5-Vision       & .748 & \swst{.359} & \wst{.267} & .191 & .066 & .013 & \swst{.023} & .009 & .454 & .533 & \wst{.154} & \swst{.217} & .613 & .316 & \swst{.198} & \swst{.256} & \wst{.030} & .160 & .221 & \swst{.044} & .562 & .576 \\
    17 & Llama-3.2-Vision     & .545 & .482 & .518 & .434 & .234 & .031 & .104 & .054 & \wst{.317} & \swst{.483} & .286 & .520 & .724 & .286 & .260 & .317 & .061 & \swst{.125} & .250 & .722 & \wst{.000} & \wst{.500} \\
    18 & LLaVA-NeXT-Mistral   & \wst{.364} & .392 & .408 & \swst{.162} & \swst{.029} & \swst{.005} & \wst{.015} & \swst{.004} & \swst{.372} & .484 & \swst{.259} & .238 & .616 & \swst{.259} & .332 & \wst{.250} & .089 & .131 & \wst{.173} & \wst{.000} & .476 & .508 \\    \midrule
    \multicolumn{2}{@{}l}{Mean} & .725 & .545 & .541 & .382 & .247 & .035 & .123 & .039
      & .495 & .574 & .370 & .502
      & .695 & .407 & .341 & .331 & .401 & .259 & .309 & .438
      & .396 & .558 \\
    \multicolumn{2}{@{}l}{Floor} & .500 & .100 & .111 & .060 & .035 & .005 & .001 & .011
      & .000 & .000 & .000 & .000
      & .501 & .250 & .410 & .250 & .415 & .250 & .170 & .000
      & .000 & .500 \\
    \bottomrule
  \end{tabular*}
  \caption{Bio-VQA results for 18 models across 24 tasks. Colors mark the \bst{best}, \snd{second best}, \swst{second worst} and \wst{worst} model in each column, over the 18 models only. \textbf{\ctax{}} is exact match at eight ranks. \textbf{\cattr{}} is \cags{} for relative salience (sal), morphological attributes (mrp), anatomical localization (loc) and naming (nam). \textbf{\cspat{}} reports one metric per task: \cas{} for counting (cnt); balanced accuracy for relative counting (rcnt), relative distance (rdst) and relative position (rpos); the share of measurements within 20\% of the true value for absolute distance (adst); mean IoU for localization (lcl) with unparsed replies scored as zero; agreement over gold labels for viewpoint (view); and exact match for OCR. \textbf{\creid{}} is discriminability ($\dpr$) and accuracy under random pairing. \emph{Floor} is the score an image-blind rule reaches, and the most frequent taxon at each rank for \ctax{}. \cags{} and $\dpr$ are chance-corrected, so floors are zero. $^{\ast}$Domain specialist, taxonomy, with the full TreeOfLife-200M label set (tol). $^{\ast\ast}$Domain specialist, re-identification, on Grevy's zebra only.}
  \label{tab:main}
\end{table*}

\Cref{tab:main} reports all 18 models on 24 tasks across 4 components. We read it against two reference points. The floor row gives the score an image-blind rule reaches. The two specialist rows give the score a purpose-built model reaches.

\subsection{\ctax{}: VLMs fail at finer ranks}
\label{sec:exp-c1}

Exact match falls at every step down the hierarchy. The 18 models average .725 at phylum, .545 at class, .382 at family, .247 at genus and .123 at scientific name. This is a fall to one sixth over the same images. However, it is not ordered for every model: 9 models score higher at order than at class.

\paragraph{Coarse accuracy is inflated.}
Half our images are chordates, so an image-blind model answering \emph{Chordata} to every image scores .500 at phylum. Three models score below that: InternVL3.5-30B at .486, Qwen2.5-VL-3B at .416 and LLaVA-NeXT-Mistral at .364. After subtracting the floor, models recover .450 of the headroom\footnote{Share of the gap between floor and a perfect score that a model recovers.} at phylum and .494 at class, although raw accuracy favors phylum by 1.3$\times$. So the accuracy is inflated for phylum, partly because the answer distribution is narrower. 

\paragraph{The collapse starts at finer ranks.}
The most common family covers 6\% of the images, the most common genus 3.5\% and the most common species 0.5\%, so whatever a VLM scores here, it earns from the image. Even the best model reaches .612 at family and .269 at scientific name.

\paragraph{The specialist gap widens as the answer sharpens.} The best model reaches .731 at class against BioCLIP-2's .874, and .269 at scientific name against .544. The specialist advantage grows from $1.2\times$ to $2.0\times$. BioCLIP-2 is trained on TreeOfLife-200M, the corpus our \ctax{} images come from, and ranks a closed label set while VLMs answer open-ended, so we treat it as an optimistic upper bound than a baseline.

\paragraph{Scale helps mostly at coarse ranks.} From 8B to 38B, InternVL3 gains 18.7 points at class and 6.5 at scientific name. InternVL3.5 gains 10.5 and 5.0 respectively. Both families gain 2$\times$~to 3$\times$ more at class than at scientific name\footnote{Qwen3-VL ranks differently: from 8B to 32B it drops from .680 to .615 at class and rises from .202 to .244 at scientific name}. Extra parameters can widen the coverage of categories these models partly knew, but they can not sharpen the finer distinctions they missed.

\paragraph{Models recognize what they cannot recall.} On open-ended common name questions, models average .039. On the same images with four options, they average .645 ($16.4\times$). The images and targets are identical, so the knowledge is present. \emph{VLMs can recognize the correct name among four options, but cannot retrieve it unprompted}.

\subsection{\cattr{}: VLMs see traits but cannot localize}
\label{sec:exp-c2}
 
Of the four tasks in \cattr{}, models average .574 on morphological attributes, .502 on naming, .495 on relative salience and .370 on anatomical localization. Naming a bird and finding its traits is easier than identifying a body part and its laterality. The two hardest sub-tasks are both from anatomical localization: part-set visibility (\emph{which parts of the bird are visible?}) averages .341 and laterality (\emph{which of its eyes is visible?}) .411. \textbf{Why are these the hardest?} A possible explanation is that these questions require more fine-grained reasoning. For example, from a coarser look, a model may answer the bird as \emph{Cardinal}. And once it knows this, it may answer that the wing is red. Answering which of its two eyes is visible needs a finer look at the image.

\paragraph{Models confuse left and right.}
We take only the items where a model names the correct part, then ask which side it is on. 17 of the 18 models score at or below chance and 9 fall significantly below. Mirror confusion runs from 21.8\% to 54.0\% against a 20\% baseline and rises with model strength ($\rho$=.75), so stronger models rule out the wrong part but then choose the mirror side. The one exception is Molmo~2-8B, the only model trained with explicit pointing supervision. So, \emph{VLMs find it difficult to localize a body part. Even when they can, they cannot tell which side it is.}

\paragraph{The \cags{}~metric holds up.}
We sweep eleven kernel and hyperparameter settings and the model ranking barely moves ($\rho\geq$.996 on the base grid, and $\rho$=.959 when anatomy similarity enters the visibility kernel). Three annotators rate 306 items and agree with \cags{}~at .908 of the rate they agree with each other. On the items exact match is zero, \cags{}~separates credited answers from rejected ones (AUC .676).

\paragraph{Models top on one task and tank on another.}
Molmo~2-8B leads marker localization at $.743$ and laterality at .746, the latter ahead of the next model by $.220$. But on naming it ranks 15th, scoring $.330$. Qwen2.5-VL-7B is the exact opposite, 2nd on naming at $.713$ and 12th on anatomical localization at $.356$. So, \emph{one finds the parts but cannot name the bird; the other names the bird but cannot find its parts.}

\subsection{\cspat{}: VLMs see objects but not their geometry}
\label{sec:exp-c3}

All eight tasks in \cspat{} carry a floor, and models fall below the floor on two of them: localization and absolute distance.
 
\paragraph{Localization is the hardest and only Qwen3 does it well.}
Models average .401 mean IoU against a mean gold box at .415, and 12 of the 18 fall below it. Only Qwen3-VL-32B (.866), Qwen3-VL-8B (.862) and Qwen3-VL-30B (.861) find the animal reliably. Molmo~2-8B and Phi-3.5-Vision score lowest because most of their replies repeat the example box in the prompt.

\paragraph{Models recall a size instead of measuring one.}
Answering one fixed size per measurement type (elytral length and width) would score .410. Out of 18 models, 15 fall below this floor; only Qwen3-VL-32B (.622), Qwen3-VL-30B (.528) and InternVL3.5-38B (.503) beat it. The others answer in the right units but with far too little spread across. They \emph{recall a measurement instead of reading the bar.}

\paragraph{Counting is close but rarely accurate.}
Models average .695 count agreement against a floor of .501, and InternVL3-38B leads at .799, but exact match never exceeds .242: \emph{VLMs find the right ballpark but not the exact count.} The gap is specific to biological subjects. Qwen3-VL-32B counts 2 to 10 everyday objects exactly 94.9\% of the time on CountBench~\cite{paiss2023countbench,qwen3vl}, but only 41.7\% of single-image \biovqa~items in the same range, $2.3\times$ lower.

\paragraph{Relative tasks barely clear their floors.}
Relative counting averages .407 against a floor of .250 and relative distance .331 against .250. Relative position averages .259 against a guessing floor of .250 over its four options, so it stays at chance. \emph{VLMs see objects but cannot relate one to another.}

\paragraph{OCR works as the negative control we designed.}
The seven spatial tasks correlate at mean $\rho = +.703$ and none correlates with OCR. InternVL3-38B tops counting at .799 but drops to 14th on OCR at .105, so \emph{spatial composite measures spatial ability rather than general model competence.}

\subsection{\creid{}: VLMs answer from resemblance}
\label{sec:exp-c4}

\paragraph{The specialist leads the field.}
We run MegaDescriptor on Grevy's zebra and compare it against the VLMs. Under random pairing, it reaches $\dpr$ .733 against a VLM mean of .391 ($1.9\times$), and under easy pairing 2.055 against 1.346 ($1.5\times$). \emph{A specialist re-identification model is about twice as good as the average VLM}, although the two strongest VLMs exceed it on this dataset.

\paragraph{Under adversarial condition, models fall below chance.}
We pair different animals that sit closest in the DINOv2 embedding space, so the two match on pose, lighting and background. 17 of the 18 models then score negative $\dpr$, from $-.932$ to $-.091$. Scoring below chance is not the error a guesser makes. It means a model reads a cue that we inverted. MegaDescriptor fails the same way at $-.969$. So, \emph{when the two animals that look most alike are the different ones, models keep calling them the same. Even specialists fail under adversarial conditions.}

\paragraph{Difficulty ordering holds across models and species.}
77 of 90 model-dataset cells (18$\times$5) show the easy $\geq$ random $\geq$ hard ordering, against a $1/6$ chance of monotone order. Response bias moves with it: criterion $c$ is negative for 17 of 18 models and grows more negative as pairs harden. \emph{Models answer \textnormal{yes} more often as the task gets harder, not less.}

\subsection{Cross-component correlation}
\label{sec:exp-cross}

\begin{table}[t]
  \centering
  \footnotesize
  \setlength{\tabcolsep}{4pt}
  \begin{tabular}{lcccc}
    \toprule
    Spearman $\rho$ & \ctax{} & \cattr{} & \cspat{} & \creid{} \\
    \midrule
    \ctax{}  \emph{($\alpha$ .981)} & --- & .689 & .787 & .433$^{\dagger}$ \\
    \cattr{} \emph{($\alpha$ .901)} &     & ---  & \best{.927} & .514$^{\dagger}$ \\
    \cspat{} \emph{($\alpha$ .925)} &     &      & ---  & .610 \\
    \creid{} \emph{($\alpha$ .889)} &     &      &      & --- \\
    \bottomrule
  \end{tabular}
  \caption{Correlation between components over the 18 models scored on all four. $\alpha$ is each component's internal consistency. $\dagger$~not significant after Holm correction. Confidence intervals are wide at $n = 18$ (median width .66), enough to separate correlated from uncorrelated pairs but not to rank one pair above another.}
  \label{tab:cross}
\end{table}

We correlate all 24 tasks over the 18 models to ask whether the four components measure four things. \Cref{tab:cross} reports the component-level correlations.
 
\paragraph{Tasks agree more within a component than across one.}
Task pairs within a component correlate at mean $\rho$ .749, and pairs drawn from different components at .534. A label permutation puts $p = .0005$ on that gap. Per component, the internal agreement is .841 for \ctax{}, .662 for \cspat{}, .645 for \cattr{} and .617 for \creid{}. \emph{Tasks within a component agree enough for the component score to summarize them, and differ enough to be worth reporting on their own.}
 
\paragraph{\creid{} is the stand-alone component.}
It correlates with \ctax{} at .433 and with \cattr{} at .514, and neither survives multiple-comparison correction. \emph{Knowing how well a model names an organism tells us nothing about whether it can tell two individuals of that organism apart.}
 
\paragraph{\cattr{} and \cspat{} overlap.}
The two correlate at .927. At the reliability we can measure, they are one construct. Both ask a model to find something fine-grained, so fine-grained attributes and spatial reasoning meet at the same skill. We keep them apart because they diagnose different failures, and we do not treat them as independent evidence.
 
\paragraph{A strong model is strong everywhere.}
A single factor explains 62.2\% of the variance. InternVL3-38B, Qwen3-VL-30B and Qwen3-VL-32B place top-five in all components, so any of them is a safe pick whatever the task. Below them the ranking reorders: the median model moves 5.5 of 18 places between components, and Phi-3.5-Vision is at the 8th percentile on \ctax{} and 64th on \creid{}. \emph{A mid-tier model is good at one component and poor at another.}

\section{Discussion}
\label{sec:discussion}

\paragraph{Implications.}
Multiple choice says little about how a model answers on its own (.645 with four options, .039 without). Without a floor, we cannot tell whether a model looked at the image: on two tasks an image-blind rule beats most models. Where a specialist exists, it stays ahead of most VLMs.

\paragraph{Limitations.}
We evaluate open-weight VLMs only, so we cannot say if proprietary models like GPT-5 behave the same way. We run one decoding protocol, so we cannot say whether chain-of-thought or a reasoning model would change the results. Our images are from public datasets that a model may have seen in pretraining, including the BioCLIP-2 training corpus for \ctax{}. We report no human performance reference, and human raters validate \cattr{} scoring only, with no human audit of \ctax{}, \cspat{} and \creid{} ground truth. Our floors are rule-based, and a text-only VLM may set a higher one. Two models copy example answers on localization. Species-epithet answers are truncated, so we analyze species results with scientific name. Lastly, \cags{} anatomy kernel over-credits laterality confusions (See \cref{sec:exp-c2}).

\paragraph{Conclusion.}
In \biovqa, we ask 1.86M questions about 217K images, and after scoring the components against a floor or a specialist, we find that VLMs are good at coarse categories but poor at the finer ones, and scaling helps at coarse ranks. The gap to a specialist grows as the task gets more fine-grained. We release the benchmark\footnote{\url{https://huggingface.co/datasets/imageomics/Bio-VQA}}, the generators, the scorers\footnote{\url{https://github.com/Imageomics/Bio-VQA}} and every per-item answer, along with \biovqa-Lite, a stratified sample of 500 questions per task (12K in total) for low-cost evaluation.

\paragraph{Acknowledgments.} This research was supported by the Imageomics Institute, funded by the US National Science Foundation's (NSF) Harnessing the Data Revolution (HDR) program under Award No. 2118240. And, Q.~M. and Y.~M. are supported by NSF under grant number NSF2406648.

{
    \small
    \bibliographystyle{ieeenat_fullname}
    \bibliography{main}
}

\clearpage
\begin{center}
  {\Large\bfseries Supplementary Material}
\end{center}
\vspace{0.5em}
\appendix

\section*{Contents}
\label{sec:supp-contents}

We organize this supplement into thirteen sections. Every number here comes from the same scored files that produce the tables in the main paper. We add no new experiments and no new references.

\begin{itemize}\itemsep0.5em

  \item[\textbf{A}] \textbf{Benchmark statistics.} We give counts for every component, group, family and task.

  \item[\textbf{B}] \textbf{Models and specialists.} We list the 18 models with their parameter counts and result-file keys. We give the decoding protocol and the parse rates. We then describe the two specialists. BioCLIP-2 bounds \ctax{} from above. It cannot read a prompt, so we derive all eight rank answers from prediction. MegaDescriptor bounds \creid{} from above. We fit its threshold on the random condition and apply it unchanged to easy and hard.

  \item[\textbf{C}] \textbf{Metric definitions.} We give all 29 equations with a worked example for each. They cover exact match, the synonym relation, severity, \cags{} and its six kernels, \cas{}, the image-blind floor, mean IoU and $\dpr$.

  \item[\textbf{D}] \textbf{\ctax{} full results.} We show why phylum accuracy is inflated, and we name the three models an image-blind rule beats there. We report the recognition--recall gap of $16.4\times$ on identical images. We show that scale helps mostly at the coarse ranks. We give the full $10 \times 10$ cross-group matrix.

  \item[\textbf{E}] \textbf{\ctax{} difficulty stratification.} We tier every \ctax{} item from the taxonomy and the answer key alone, so no prediction file ever changes a tier. We use prevalence for open recall and confusability for choice items.

  \item[\textbf{F}] \textbf{\cattr{} full results.} We give \cags{} and exact match for every model, category and family. We report three anomalies. One model answers option A two thirds of the time. One model misses farther from the ground truth than a random distractor. One 8B model beats every 38B competitor at grounding and ranks 15th at naming.

  \item[\textbf{G}] \textbf{\cspat{} full results.} We give all eight tasks with their floors beside their scores. We show the two-panel counting penalty, which holds for every model, and we show that models compress absolute distance. We close with the OCR control.

  \item[\textbf{H}] \textbf{\creid{} full results.} We report $\dpr$ and criterion $c$ by condition, and we test the difficulty manipulation. We give per-species results and the MegaDescriptor baseline. We then explain why scoring below chance is not deliberate error.

  \item[\textbf{I}] \textbf{Metric validation.} We sweep \cags{} across eleven configurations and report how far the ranking moves. We run a human study on 306 items with three annotators, and we report where the metric and the humans agree and disagree.

  \item[\textbf{J}] \textbf{Cross-component analysis.} We ask whether four components measure four things. We show that \cattr{} and \cspat{} are one construct, and that \creid{} is the real dissociation.

  \item[\textbf{K}] \textbf{Prompts and parsing.} We give the prompt and template for every question family. We specify the parser, which both evaluators use, so the two metrics partition the items identically.

  \item[\textbf{L}] \textbf{Reproducibility.} We say what reproduces exactly and what does not. We give the verification contract and the bootstrap procedure.

  \item[\textbf{M}] \textbf{Extended limitations.} We give five benchmark-level limitations, then six for \ctax{}, six for \cattr{}, five for \cspat{} and nine for \creid{}.
  
\end{itemize}

\paragraph{How to read the tables.}
We report exact match for every task. We put each graded metric next to the exact match, so no claim rests on a graded score alone. Where a task has an image-blind floor, we report the floor in the same table. Where a model sits below its floor, we say so.

\paragraph{The bounds.}
We bound every task from at least one side. \ctax{} has the most frequent taxon at each rank as its floor and BioCLIP-2 as its ceiling. \cattr{} and \creid{} have a floor of zero, because \cags{} and $\dpr$ are chance-corrected, and \creid{} has MegaDescriptor as its ceiling. \cspat{} has one image-blind rule per task and no ceiling, since no specialist covers all eight tasks. \cattr{} also has three annotators, but they bound the metric and not the task.

\paragraph{A note on model names.}
The main paper uses display names. The result files use short keys. We give both in \cref{tab:supp-models} and use display names everywhere else.

\section{Benchmark Statistics}
\label{sec:supp-stats}

\subsection{\ctax{}: groups and question types}

We draw \num{10} organismal groups from TreeOfLife-200M and take \num{10000} images from each. \cref{tab:supp-c1groups} shows the filter that defines each group. Three groups need more than one filter value, because the GBIF Backbone holds no single accepted node that covers them.

\textbf{Fish spans six classes.} Fish is a paraphyletic concept and the backbone never treats it as one taxon. \emph{Actinopterygii} is a valid class, but it covers the ray-finned fishes only and leaves out sharks, chimaeras, coelacanths and lungfish. We therefore assemble the group from six class values. \emph{Teleostei} sits inside \emph{Actinopterygii} in most treatments, and the backbone assigns both as class on different records, so we take both.

\textbf{Reptiles carry no order.} The backbone places \emph{Squamata}, \emph{Testudines}, \emph{Crocodylia} and \emph{Sphenodontia} at class rank, and puts \emph{Reptilia} as a synonym of those four. Family sits directly under class, so the order column is empty for every reptile row. So we ask reptiles seven questions.

\textbf{Angiosperms span two classes.} The backbone splits the flowering plants into \emph{Magnoliopsida} and \emph{Liliopsida}, both at class rank under \emph{Tracheophyta}. There is no node above them that holds the angiosperms alone, so we take both.

\begin{table}[t]
  \centering\footnotesize
  \setlength{\tabcolsep}{4pt}
  \begin{tabular}{lll}
    \toprule
    Group & Filter rank & Value \\
    \midrule
    Amphibians  & class  & \emph{Amphibia} \\
    Angiosperms & class  & \emph{Magnoliopsida}, \emph{Liliopsida} \\
    Arachnids   & class  & \emph{Arachnida} \\
    Birds       & class  & \emph{Aves} \\
    Bryophytes  & phylum & \emph{Bryophyta} \\
    Fish        & class  & \emph{Actinopterygii}, \emph{Teleostei}, \\
                &        & \emph{Elasmobranchii}, \emph{Holocephali}, \\
                &        & \emph{Coelacanthi}, \emph{Dipneusti} \\
    Gymnosperms & class  & \emph{Pinopsida} \\
    Insects     & class  & \emph{Insecta} \\
    Mammals     & class  & \emph{Mammalia} \\
    Reptiles    & class  & \emph{Squamata}, \emph{Testudines}, \\
                &        & \emph{Crocodylia}, \emph{Sphenodontia} \\
    \bottomrule
  \end{tabular}
  \caption{10 organismal groups in \ctax{} and the backbone filter that defines each. Six of them do not appear in prior biological VQAs.}
  \label{tab:supp-c1groups}
\end{table}

\ctax{} spans \num{626} families, \num{2311} genera and \num{3403} species. This is $6\times$ genus coverage and $4\times$ species coverage of VLM4Bio. We ask nine question types per image. \cref{tab:supp-c1types} shows each type, the rank it scores, and an example answer. Nine groups contribute nine questions and Reptiles contributes eight, which gives \num{890000} questions in total.

\begin{table}[t]
  \centering\footnotesize
  \setlength{\tabcolsep}{4pt}
  \begin{tabular}{lll}
    \toprule
    Question type & Scored rank & Example answer \\
    \midrule
    phylum          & phylum          & \emph{Chordata} \\
    class           & class           & \emph{Amphibia} \\
    order           & order           & \emph{Anura} \\
    family          & family          & \emph{Hylidae} \\
    genus           & genus           & \emph{Dryophytes} \\
    species epithet & species         & \emph{cinereus} \\
    scientific name & scientific name & \emph{Dryophytes cinereus} \\
    common name     & common name     & green treefrog \\
    common name MCQ & common name     & one of four candidates \\
    \bottomrule
  \end{tabular}
  \caption{The nine \ctax{} question types. We ask the common name twice on the same image, once open-ended and once as four-way multiple choice to measure the recognition--recall gap.}
  \label{tab:supp-c1types}
\end{table}

\subsection{\cattr{}: four categories, ten question families}
We build \cattr{} from CUB-200-2011, which records \num{312} binary attributes of \num{11788} annotated birds. CUB-200-2011 stores every annotation as a part, a property and a value, such as \textit{has\_wing\_color::brown}. We invert that triplet into a question and take the marked value as the answer. Based on the triplet, we divide our questions into four categories and 10 sub-categories. \cref{tab:supp-c2families} shows the ten templates and \cref{tab:supp-c2examples} reports an example for each one.
 
\begin{table}[t]
  \centering\footnotesize
  \setlength{\tabcolsep}{3pt}
  \begin{tabular}{llrl}
    \toprule
    Category & Family & Questions & Format \\
    \midrule
    Morphological & \textit{attr\_bare\_value}      & \num{280433} & MCQ \\
    \cmidrule(lr){1-4}
    Rel.\ salience& \textit{attr\_group\_value}     & \num{321073} & MCQ \\
    \cmidrule(lr){1-4}
    Anatomical    & \textit{marker\_localization}   & \num{11788}  & MCQ \\
                  & \textit{visibility\_partset}    & \num{117880} & MCQ \\
                  & \textit{visibility\_side}       & \num{35364}  & MCQ \\
    \cmidrule(lr){1-4}
    Naming        & \textit{category\_mcq}          & \num{11191}  & MCQ \\
                  & \textit{open\_common\_name}     & \num{11788}  & OE \\
                  & \textit{open\_scientific\_name} & \num{11668}  & OE \\
                  & \textit{open\_genus}            & \num{11728}  & OE \\
                  & \textit{open\_kind\_of}         & \num{11191}  & OE \\
    \midrule
    \multicolumn{2}{l}{\textbf{Total}} & \textbf{\num{824104}} & \\
    \bottomrule
  \end{tabular}
  \caption{\cattr{} categories and question families. About 94\% of items are multiple choice questions (MCQ) with two to five distractors. The other 6\% are open-ended (OE) naming.}
  \label{tab:supp-c2families}
\end{table}
 
\begin{table*}[t]
  \centering\footnotesize
  \setlength{\tabcolsep}{4pt}
  \renewcommand{\arraystretch}{1.15}
  \begin{tabular}{@{}lp{7.3cm}p{4.0cm}l@{}}
    \toprule
    Family & Question & Options & Answer \\
    \midrule
    \textit{attr\_bare\_value}
      & What is the bill color of this bird?
      & white / yellow / black / pink / brown
      & black \\
    \textit{attr\_group\_value}
      & Which of the following is the most prominently visible physical trait of
        the bird in this image?
      & bill color: black / underparts color: white / primary color: green /
        breast color: red / eye color: white
      & breast color: red \\
    \textit{marker\_localization}
      & What is the body part of this bird marked with the yellow dot?
      & beak / breast / left leg / left eye / left wing
      & beak \\
    \textit{visibility\_partset}
      & Which of the following parts of the bird are visible in this image?
      & breast / beak / left wing / None / All three
      & beak \\
    \textit{visibility\_side}
      & Which of the eyes of the bird are visible in this image?
      & Left / Right / Both / None
      & Both \\
    \textit{category\_mcq}
      & What type of bird is this?
      & Waxwing / Nutcracker / Waterthrush / Grebe / Gull
      & Gull \\
    \textit{open\_common\_name}
      & What is the common name of the bird?
      & open-ended
      & California Gull \\
    \textit{open\_scientific\_name}
      & What is the scientific name of the bird?
      & open-ended
      & \emph{Larus californicus} \\
    \textit{open\_genus}
      & What is the genus of the bird?
      & open-ended
      & \emph{Larus} \\
    \textit{open\_kind\_of}
      & What kind of gull is this?
      & open-ended
      & California \\
    \bottomrule
  \end{tabular}
  \caption{One example per \cattr{} family, all asked against the same photograph of a California gull. The four naming families ask the same bird at four levels of specificity, and only one of the four gives the model options.}
  \label{tab:supp-c2examples}
\end{table*}
 
\textbf{The families do not share an answer space.} \textit{color} takes one of fifteen attested values, \textit{bill\_length} takes one of three, and \textit{visibility\_side} takes one of four. A raw graded score pays a model more for a lucky guess on a narrow item than on a wide one. This is what we correct per item in \cref{sec:supp-metrics}.
 
\textbf{A near miss implies something different in each family.} Answering \emph{purple} for \emph{blue} is close on a perceptual scale. Answering \emph{nape} for \emph{crown} is close on an anatomical layout. Answering \emph{Left} for \emph{Both} is half of a set. No single kernel expresses all three, so we dispatch one kernel per family.
 
The four naming families allow us to ask naming questions at four levels of specificity on one image, from \emph{gull} to \emph{Larus californicus}. Only \textit{category\_mcq} supplies options, so the gap between it and \textit{open\_common\_name} isolates recognition from recall without changing the image or the target.
 
We render a marker onto each of the \num{11788} source images for anatomical marker localization questions, so \cattr{} holds
\num{23576} images and \num{824104} questions.

\subsection{\cspat{}: eight spatial tasks}
 
We generate a \cspat{} item only where an annotation carries a count, a measurement, a box or an angle. \cref{tab:supp-c3tasks} reports the eight spatial tasks, and the dataset each one runs on. \cref{tab:supp-c3examples} gives an example per task.
 
\begin{table}[t]
  \centering\footnotesize
  \setlength{\tabcolsep}{3pt}
  \begin{tabular}{llrl}
    \toprule
    \# & Task & Items & Dataset \\
    \midrule
    1 & Counting          & \num{53498} & all ten subsets \\
    2 & Viewpoint         & \num{13000} & wildlife datasets \\
    3 & Absolute distance & \num{14108} & beetles \\
    4 & Relative distance & \num{3976}  & beetles \\
    5 & Relative counting & \num{42412} & all ten subsets \\
    6 & Localization      & \num{3988}  & wildlife datasets \\
    7 & Relative position & \num{1000}  & beetles \\
    8 & OCR    & \num{522}   & beetles \\
    \midrule
    \multicolumn{2}{l}{\textbf{Total}} & \textbf{\num{132504}} & \\
    \bottomrule
  \end{tabular}
  \caption{\cspat{} spatial tasks with the dataset each runs on. A task runs only where an existing annotation provides the ground truth, which is why five of them run on beetles.}
  \label{tab:supp-c3tasks}
\end{table}
 
\begin{table*}[t]
  \centering\footnotesize
  \setlength{\tabcolsep}{4pt}
  \renewcommand{\arraystretch}{1.15}
  \begin{tabular}{@{}lp{7.2cm}p{3.4cm}l@{}}
    \toprule
    Task & Question & Answer form & Answer \\
    \midrule
    Counting
      & How many beetles are there in this image?
      & integer & 11 \\
    Relative counting
      & Which of the 4 images contains the most number of beetles?
      & one of four panels & Image B \\
    Absolute distance
      & What is the measurement of the elytra width, marked with the red line,
        with respect to the scalebar, marked with the blue line, in cm?
      & number in cm & 0.497 \\
    Relative distance
      & Which beetle has the largest elytral length?
      & one of four specimens & Beetle D \\
    Localization
      & Provide the bounding box coordinate of the zebra in this image.
      & four integers & [854, 819, 3009, 2322] \\
    Relative position
      & Where is Beetle (blue) located with respect to Beetle (red)?
      & one of eight directions & Top-right \\
    Viewpoint
      & What is the viewpoint of the hyena?
      & one of ten orientations & left \\
    OCR (control)
      & What is the barcode text?
      & alphanumeric string & A00000012435 \\
    \bottomrule
  \end{tabular}
  \caption{One example per \cspat{} task. Five of the eight take an open-ended answer, so a model cannot recover the answer form from a list of options. The answer form column is what the parser expects, and \cref{sec:supp-prompts} gives the full template for each.}
  \label{tab:supp-c3examples}
\end{table*}
 
\paragraph{Counting runs on three canvas tiers.}
A counting item is a canvas at a fixed pixel budget, so the encoder receives the same number of tokens whatever the tier. The \emph{Easy} tier fills the canvas with one image. The \emph{Medium} tier splits the canvas into two panels drawn from different datasets. The \emph{Hard} tier splits it into two panels drawn from the same dataset. A target therefore occupies about half the area in Medium and Hard that it occupies in Easy. The three tiers isolate that area change from any change in
content, and \cref{sec:supp-c3} reports what the split costs every model.

\paragraph{We macro-average counting over the ten subsets.}
The ten datasets do not contribute equal numbers of questions, they differ tenfold. GZCD contributes 2.9 times the median, because it produces three nested questions per image. Under micro-averaging, this one dataset supplies about 29\% of the counting headline. Under macro-averaging it supplies 10\%. So we report macro-average.
 
Absolute distance, relative distance and relative position run on the beetle trays, because that is the only dataset with recorded trait and scalebar measurements and where the specimens lie flat at one focal depth. Localization and viewpoint run on the \creid{} wildlife datasets, where one animal is in the foreground of the frame.

\subsection{\creid{}: five datasets, three conditions}
 
We place two photographs side by side on one canvas and ask a single question.
 
\begin{quote}\footnotesize
Are these two photographs of the same individual animal? Answer Yes or No.
\end{quote}

A model has to parse two subjects in one frame before it can compare them. This configuration is identical across models and conditions, so it does not confound the difficulty.
 
\begin{table}[t]
  \centering\footnotesize
  \setlength{\tabcolsep}{4pt}
  \begin{tabular}{lll}
    \toprule
    dataset & Species & Identity signal \\
    \midrule
    WhaleSharkID & whale shark    & spot pattern \\
    BelugaID     & beluga whale   & scarring, dorsal ridge \\
    HyenaID      & spotted hyena  & spot pattern \\
    LeopardID    & leopard        & rosette pattern \\
    GZCD         & Grevy's zebra  & stripe pattern \\
    \bottomrule
  \end{tabular}
  \caption{The five \creid{} datasets and the marking that identifies an individual in each. Four carry a pattern that is unique to the animal and stable across sightings. Beluga is the exception.}
  \label{tab:supp-c4cats}
\end{table}
 
\cref{tab:supp-c4cats} reports the five datasets with identity signal. Beluga is the one species with no unique pattern, and identifying a beluga requires reading accumulated scarring, which changes between sightings. This reflects on the results as reported in \cref{sec:supp-c4}, where beluga is the hardest dataset for every model we test. Each dataset contributes 1000 pairs per condition, balanced 500 same-individual against 500 different-individual, which gives \num{15000} pairs in total. We draw the same-individual pairs first, because every dataset holds far fewer repeat sightings than distinct animals, so this is the scarce resource.
 
\subsection{Source datasets}
 
Every image and ground-truth annotation in \biovqa{} comes from a public dataset. We generate a question only where an existing annotation supplies the answer, so no answer key depends on our own labeling. \cref{tab:supp-sources} gives the eleven sources and the component each serves.
 
\begin{table}[t]
  \centering\footnotesize
  \setlength{\tabcolsep}{3pt}
  \begin{tabular}{lll}
    \toprule
    Dataset & Used by & Content \\
    \midrule
    TreeOfLife-200M & \ctax{} & 100K subset, ten groups \\
    CUB-200-2011    & \cattr{} & \num{11788} annotated birds \\
    WAID            & \cspat{} & aerial wildlife \\
    Boxes on Bees   & \cspat{} & dense hive frames \\
    2018-NEON-beetles & \cspat{} & trays with scalebars \\
    DePondFi        & \cspat{} & pond fish \\
    GZCD            & \cspat{}, \creid{} & Grevy's zebra \\
    WhaleSharkID    & \cspat{}, \creid{} & whale shark \\
    HyenaID         & \cspat{}, \creid{} & spotted hyena \\
    LeopardID       & \cspat{}, \creid{} & leopard \\
    BelugaID        & \creid{} & beluga whale \\
    \bottomrule
  \end{tabular}
  \caption{The eleven public datasets behind \biovqa{}. Four of them serve two components.}
  \label{tab:supp-sources}
\end{table}
 
\paragraph{What we release.}
Every image keeps the license of its source, so what we ship depends on where the ground truth sits. Most Bio-VQA items ask about a source photograph. For these we ship an identifier and not the image. The identifier is a TreeOfLife-200M 
{\small\texttt{UUID}} for C1, a CUB filename for C2, and a catalog record id for C3 viewpoint, localization and OCR. This covers C1 in full and 9 of the 10 C2 families, which together hold most of the questions. We pin the {\small\texttt{UUID}} and the content hash of every C1 image, because our download runs concurrently under a budget per group and a fresh run would select a different 10,000 images per group. The rest of the items ask about an image we render, and for those the rendered image is the ground truth. C2 marker localization asks which part carries a dot we draw. C3 counting asks about a canvas we compose from one or two panels, and absolute and relative distance ask about lines we overlay on a beetle tray. Every C4 pair is two photographs we place side by side. In each case, the answer is correct only for the image we produced, so we redistribute those images under the terms of their upstream sources, with per-source attribution in the release.

\section{Models and Protocol}
\label{sec:supp-models}

\subsection{The 18 open-weight models}
 
\cref{tab:supp-models} lists the 18 models we evaluate, with the checkpoint, the language backbone and the vision path of each.
 
\begin{table*}[t]
  \centering\scriptsize
  \setlength{\tabcolsep}{4pt}
  \renewcommand{\arraystretch}{1.12}
  \begin{tabularx}{\textwidth}{@{}llllX@{}}
    \toprule
    Model & Checkpoint & Params & LLM backbone & Vision path \\
    \midrule
    Qwen2.5-VL-3B  & \texttt{Qwen/Qwen2.5-VL-3B-Instruct}  & 3.75B & Qwen2.5-3B & Native-resolution ViT (0.67B), window attention \\
    Qwen2.5-VL-7B  & \texttt{Qwen/Qwen2.5-VL-7B-Instruct}  & 8.29B & Qwen2.5-7B & Native-resolution ViT (0.67B), window attention \\
    \cmidrule(lr){1-5}
    Qwen3-VL-8B    & \texttt{Qwen/Qwen3-VL-8B-Instruct}    & 8.8B  & Qwen3-8B      & Qwen3-VL ViT, DeepStack fusion \\
    Qwen3-VL-30B-A3B & \texttt{Qwen/Qwen3-VL-30B-A3B-Instruct} & 30.5B / 3B active & Qwen3-30B-A3B & Qwen3-VL ViT, DeepStack fusion \\
    Qwen3-VL-32B   & \texttt{Qwen/Qwen3-VL-32B-Instruct}   & 32B   & Qwen3-32B     & Qwen3-VL ViT, DeepStack fusion \\
    \cmidrule(lr){1-5}
    InternVL3-8B   & \texttt{OpenGVLab/InternVL3-8B-Instruct}  & 8B   & Qwen2.5-7B  & InternViT-300M-448px-V2.5 \\
    InternVL3-38B  & \texttt{OpenGVLab/InternVL3-38B-Instruct} & 38B  & Qwen2.5-32B & InternViT-6B-448px-V2.5 \\
    \cmidrule(lr){1-5}
    InternVL3.5-8B & \texttt{OpenGVLab/InternVL3\_5-8B-Instruct} & 8B & Qwen3-8B & InternViT-300M-448px-V2.5 \\
    InternVL3.5-30B-A3B & \texttt{OpenGVLab/InternVL3\_5-30B-A3B-Instruct} & 31B / 3B active & Qwen3-30B-A3B & InternViT-300M-448px-V2.5 \\
    InternVL3.5-38B & \texttt{OpenGVLab/InternVL3\_5-38B-Instruct} & 38.4B & Qwen3-32B & InternViT-6B-448px-V2.5 \\
    \cmidrule(lr){1-5}
    Gemma-3-12B    & \texttt{google/gemma-3-12b-it} & 12B & Gemma 3 12B & SigLIP-So400m/14-896, pan-and-scan \\
    Gemma-4-12B    & \texttt{google/gemma-4-12B-it} & 12B & Gemma 4 12B & Encoder-free. A 35M embedder projects raw
      $48{\times}48{\times}3$ patches straight into the LLM embedding space \\
    \cmidrule(lr){1-5}
    Molmo2-8B      & \texttt{allenai/Molmo2-8B} & 8.66B & Qwen3-8B & SigLIP 2 \\
    Pixtral-12B    & \texttt{mistral-community/pixtral-12b} & 12.4B & Mistral NeMo 12B & Pixtral-ViT (0.4B), trained from scratch, variable resolution \\
    Phi-3.5-Vision & \texttt{microsoft/Phi-3.5-vision-instruct} & 4.2B & Phi-3.5-mini (3.8B) & CLIP ViT-L/14-336, multi-crop \\
    Llama-3.2-Vision & \texttt{meta-llama/Llama-3.2-11B-Vision-Instruct} & 10.6B & Llama 3.1 8B, frozen & ViT-H/14 (0.63B) with a cross-attention adapter \\
    \cmidrule(lr){1-5}
    LLaVA-NeXT-Vicuna  & \texttt{llava-hf/llava-v1.6-vicuna-7b-hf}  & 7.06B & Vicuna-7B-v1.5 & CLIP ViT-L/14-336, AnyRes tiling \\
    LLaVA-NeXT-Mistral & \texttt{llava-hf/llava-v1.6-mistral-7b-hf} & 7.57B & Mistral-7B-Instruct-v0.2 & CLIP ViT-L/14-336, AnyRes tiling \\
    \bottomrule
  \end{tabularx}
  \caption{The 18 open-weight models, from 3.75B to 38.4B parameters. Two are mixture-of-expert variants that activate 3B parameters per token.}
  \label{tab:supp-models}
\end{table*}
 
Two groupings in \cref{tab:supp-models} give us controlled comparisons:
 
\textbf{Three models share one backbone.} Qwen3-VL-8B, InternVL3.5-8B and Molmo2-8B all run Qwen3-8B. Their vision paths differ. Qwen3-VL uses a DeepStack ViT, InternVL3.5 uses InternViT-300M, and Molmo2 uses SigLIP 2. Molmo2-8B leads the field on marker localization and on laterality in \cref{sec:supp-c2}. The language model is the same in all three, so that lead comes from the vision path and not from the backbone.
 
\textbf{Two pairs share a vision encoder.} InternVL3-38B and InternVL3.5-38B both use InternViT-6B as vision encoder. InternVL3-38B runs Qwen2.5-32B as LLM backbone, whereas InternVL3.5-38B runs Qwen3-32B. InternVL3-8B and InternVL3.5-8B do the same with InternViT-300M.
 
\textbf{One model has no vision encoder.} Gemma-4-12B projects raw pixel patches into the language embedding space with one matrix multiply. The embedder holds 35M parameters. The model leads relative salience in \cref{sec:supp-c2}
and relative distance in \cref{sec:supp-c3}, and is a case against assuming that coarse visual judgement needs a large pretrained encoder.

\subsection{Protocol and Environment}
 
\paragraph{Decoding protocol.} Every model runs under one protocol on every component. We use greedy decoding with sampling disabled. We use a fixed system prompt per question family. We use no chain-of-thought and no few-shot examples. We allow one attempt per item. One parser reads all outputs. We choose this protocol so that the benchmark measures what a model produces under a single, cheap, reproducible setting. A model that improves with chain-of-thought or with a reasoning mode may well score higher, and we say so in \cref{sec:supp-limits}. What we measure is the zero-shot single-pass answer.

\paragraph{Environments.} 
All 18 models run in one shared Python 3.12 environment, except Molmo2-8B. AI2 ships Molmo2 as custom code pinned to \textit{transformers} $4.57.1$. 
 
\subsection{Response Parsing} 
We report two rates. The parse rate is the percentage of replies from which the parser recovers an answer. The letter-text mismatch rate is the percentage of replies where a model yields a letter and an option text that disagree, such as \emph{(B) brown} when B is \emph{black} in the original option set.
 
Mismatch only exists where a question offers lettered options. Most of \biovqa{} does not. \ctax{} asks eight of its nine types open-ended, so only the common-name multiple choice can mismatch. \cspat{} asks five of its eight tasks open-ended, so only relative counting, relative distance and relative position can mismatch. \creid{} takes a yes or a no and carries no letters at all, so mismatch is undefined for the whole component. \cattr{} is the one component where most items are multiple-choice and lettered. \cref{tab:supp-parse} reports both rates along with answer form.

\begin{table*}[t]
  \centering\footnotesize
  \setlength{\tabcolsep}{4pt}
  \renewcommand{\arraystretch}{1.12}
  \begin{tabularx}{\textwidth}{@{}llrrX@{}}
    \toprule
    Component and task & Answer form & Parse & Mismatch & Parsing Criteria \\
    \midrule
    \ctax{}, eight rank types   & open text     & $1.000$ & --- &
      We match answers with ground truth after normalization (\emph{The phylum is Chordata.} normalizes to \emph{chordata}) \\
    \ctax{}, common name MCQ    & bare options  & $1.000$ & --- &
      We match the answer against the four option strings \\
    \midrule
    \cattr{}, six MCQ subtasks  & lettered options & $.960$ & $.0117$ &
      A mismatch in letter and text. Original options: \textit{(A) Red}, \textit{(B) Blue}. Answer: \textit{(B) Red}. \\
    \cattr{}, four open subtasks & open text     & $1.000$ & --- & We accept any string and score it after normalization \\
    \midrule
    \cspat{}, counting          & open integer  & $.987$ & --- &
      We reject an answer that holds no integer, such as \emph{There are many beetles} \\
    \cspat{}, viewpoint         & open text     & $1.000$ & --- &
      10 candidates given in prompt. We reject an answer that holds none. \\
    \cspat{}, absolute distance & open number   & $.970$ & --- &
      We reject an answer that holds no number in centimeters \\
    \cspat{}, relative distance & non-lettered options  & $.642$ & --- &
      We reject an answer that names none of the four specimens \\
    \cspat{}, relative counting & non-lettered options  & $1.000$ & --- &
      4 candidates given in prompt. We reject an answer that names none (\emph{Image A/B/C/D}) \\
    \cspat{}, localization      & four integers & $.523$ & --- &
      We reject an answer that holds no box, such as \emph{The zebra is in the center} \\
    \cspat{}, relative position & non-lettered options  & $1.000$ & --- &
      8 candidates given in prompt. We reject an answer that holds none. \\
    \cspat{}, OCR               & code string   & $.996$ & --- &
      We reject an answer that holds no code-like string \\
    \midrule
    \creid{}                    & yes or no     & $1.000$ & --- &
      We reject an answer that does not contain a clear \emph{Yes} or \emph{No} answer.$^\dagger$ \\
    \bottomrule
  \end{tabularx}
  \caption{Parse quality: We report the model with lowest parsing score in each component and task. Mismatch needs a letter and an option text that can disagree, so only \cattr{} reports it. \emph{Parsing Criteria} column shows what the parser accepts or rejects for that answer form.~$^\dagger$ In \creid{}, \num{269903} answers are \emph{Yes} or \emph{No}. The other \num{97} contains \emph{Yes} or \emph{No} but in a sentence (\eg.~\emph{Yes, they are the same}).}
  \label{tab:supp-parse}
\end{table*}
\ctax{} parses at $1.000$ for a structural reason. An open-ended answer is a string, and any string compares against the gold string, so nothing fails to parse. On \creid{} the parser reads a bare \emph{Yes} or \emph{No} in \num{269903} of the \num{270000} answers. The other 97 need one loose rule, which accepts yes or no appearing as a word inside a longer reply.

\subsection{Specialist baselines}
We bound two components from above with specialists. BioCLIP-2 bounds \ctax{} and MegaDescriptor bounds \creid{}.
 
\paragraph{BioCLIP-2 for \ctax{}.}
BioCLIP-2 is a contrastive vision model. It trains on the TreeOfLife-200M dataset with hierarchical taxonomic supervision. We choose this as the specialist for two reasons. First, it is in-domain, so its accuracy approximates the best a current model reaches on this label space. Second, its supervision is the Linnaean hierarchy, so its predictions carry a full lineage and we score it at every rank with the machinery we use for the VLMs.

\paragraph{We get answers from one species prediction.}
BioCLIP-2 is a classifier and cannot read a prompt. We take top-1 species per image and read the eight coarser answers off that species lineage. For multiple choice, we restrict a custom-label classifier to the four options and take the \emph{arg-max}. This configuration implies a coarse rank is only correct when the predicted species shares true lineage at that rank.

\paragraph{We test two configurations.}
We evaluate the specialist under two candidate sets. \cref{tab:supp-labelspaces} lists the three. We have not evaluated the third one, but list as a potential ceiling. 

\begin{table}[t]
  \centering\footnotesize
  \setlength{\tabcolsep}{3pt}
  \begin{tabular}{llll}
    \toprule
    Space & Candidates & Size & Role \\
    \midrule
    \textit{kingdom} & species in the named kingdom & $10^4$--$10^5$ & fair \\
    \textit{tol}     & all of TreeOfLife-200M       & ${\sim}9{\times}10^5$ & ablation \\
    \textit{group}   & the group's gold species     & $10^2$ & ceiling \\
    \bottomrule
  \end{tabular}
  \caption{The three candidate sets for BioCLIP-2. We report \textit{tol} in the main paper, which gives BioCLIP-2 no kingdom cue; \textit{kingdom} differs from it by $+0.060$ at phylum and by at most $0.005$ at every finer rank.}
  \label{tab:supp-labelspaces}
\end{table}

\paragraph{Kingdom--TOL gap.}
In {\small{\texttt{BioCLIP-2/kingdom}}, we restrict the candidate set to only the kingdom of the target taxon. In {\small{\texttt{BioCLIP-2/tol}}, there is no restriction. So without the kingdom clue, a plant image can match an animal whose name aligns in embedding space. A \emph{Pinus strobus} image classifies as the \emph{white pine barkminer moth}, an insect named after the tree it feeds on. We test both configurations and find the gap is very little. {\small{\texttt{BioCLIP-2/kingdom}} and {\small{\texttt{BioCLIP-2/tol}} differ by $+0.060$ at phylum, which is $0.985$ against $0.925$, and by almost nothing at every finer rank. So the clue prevents gross cross-kingdom errors and gains nothing below phylum.
 
\paragraph{The specialist is far from perfect at species.}
BioCLIP-2 reaches $.545$ at species and not $1.0$. We identify four reasons for this. \textbf{Huge candidate space.} Species prediction is a top-1 decision over $10^5$ labels under kingdom restriction and $10^6$ over the full tree. The correct species competes against many near-identical neighbors, and we take top-1. So ranking it second scores zero. \textbf{Long-tailed dataset.} The training distribution is severely imbalanced. A species seen a handful of times cannot acquire a sharp decision boundary, even though its images are nominally in the training set. \textbf{Cryptic species.} Cryptic species can look identical in one photograph. The discriminating signal is not in the pixels. \textbf{Contrastive objective.} CLIP-style training aligns image and text embeddings. It does not optimize per-image label recall.
 
\paragraph{MegaDescriptor for \creid{}.}
MegaDescriptor-L-384 is a purpose-built wildlife re-identification embedding. We compare two images by cosine similarity and threshold it. We fit the threshold on the random condition and apply it unchanged to easy and hard, so those two rows are out of fit. AUC needs no threshold and converts to the units of the VLM table through $\dpr = \sqrt{2}\,z(\mathrm{AUC})$. We run MegaDescriptor on Grevy's zebra only, so we compare it against the 18 models on that dataset alone.

\section{Metric Definitions}
\label{sec:supp-metrics}
 
We give every equation we use, with a worked example for each. The main paper carries four of these. We report the rest here.
 
\subsection{\ctax{}: Exact match}
 
For a normalized prediction $\hat{y}$ and ground truth $y$,

\begin{equation}
  \EM \;=\; \mathbf{1}\!\left[\,\hat{y} = y\,\right]
  \label{eq:supp-em}
\end{equation}
 
We normalize both ground truth and predicted answers before we compare them. The normalizer case-folds, strips diacritics, removes punctuation, and removes leading boilerplate and rank prefixes. Under this map `\emph{The phylum is Chordata.}' and `\emph{chordata}' compare equal.

\subsection{\ctax{}: Soft exact match and the synonym relation}

Taxonomy changes. A species gets a new taxonomic name for the same animal. The \textbf{green treefrog} was \emph{Hyla cinerea} until 2016 and has been re-taxonomized to \emph{Dryophytes cinereus}. Both names appear in the literature, and a model trained on text written before the change learns the old one. Strict exact match calls the old name wrong, which is too strict. A model named the right animal but used a name the current backbone no longer contains. So we add a second column that credits it.
 
Let $\mathcal{S}(\hat{y})$ be the set of accepted taxa that $\hat{y}$ resolves to in the GBIF Backbone. Then
\begin{equation}
  \EMs \;=\; \mathbf{1}\!\left[\,\hat{y} = y \;\lor\; y \in \mathcal{S}(\hat{y})\,\right]
  \label{eq:supp-emsoft}
\end{equation}
 
The gap $\EMs - \EM$ measures how much of a model's apparent error is stale nomenclature and not absent knowledge. We report it~\cref{sec:supp-c1}. The gap widens exactly in the groups whose higher taxonomy was revised. The synonym mapping has two properties to address:

\textbf{It returns a set.} One taxon name can split into several modern taxa, which taxonomists call a pro-parte synonym. \emph{Reptilia} is the clearest case,
{\footnotesize
\begin{equation}
  \begin{aligned}
    \mathcal{S}(\textit{reptilia}) = \{\textit{squamata},\ \textit{testudines},\ \textit{crocodylia},\ \textit{sphenodontia}\}.
  \end{aligned}
  \label{eq:supp-proparte}
\end{equation}}
Any of these four (lizard, turtle, crocodile, tuatara) answered \emph{Reptilia} are therefore credited. A map that stored one target per name would keep one of the four and reject the other three. In the Backbone snapshot, \num{75315} names carry two or more accepted targets.

\textbf{It runs one way.} We look at the prediction and check whether it resolves onto the ground truth. We do not credit the prediction just to be in the same synonym cluster as the ground truth (\eg. \emph{Crocodylia} and \emph{Squamata} are in the same cluster). We do this to prevent over-crediting the models. As we see in~\cref{eq:supp-proparte}, \emph{Reptilia} resolves onto 4 classes. A cluster rule would link \emph{Squamata} and \emph{Crocodylia} to each other through \emph{Reptilia}, and it would credit a model that answers \emph{Crocodylia} for a lizard. The one-way rule credits \emph{Reptilia} for a \emph{Squamata}. It rejects that, because \emph{Crocodylia} is an accepted name and resolves onto itself.
 
\begin{table}[h]
  \centering\footnotesize
  \setlength{\tabcolsep}{4pt}
  \begin{tabular}{@{}lllc@{}}
    \toprule
    Truth $y$ & Answer $\hat{y}$ & $\mathcal{S}(\hat{y})$ & $\EMs$ \\
    \midrule
    \emph{Squamata} & \emph{Reptilia} & the four reptile classes & 1 \\
    \emph{Squamata} & \emph{Crocodylia} & \emph{Crocodylia} & 0 \\
    \emph{Puma concolor} & \emph{Felis concolor} & \emph{Puma concolor} & 1 \\
    \bottomrule
  \end{tabular}
\end{table}

\subsection{\ctax{}: Graded tree distance}
 
Exact match and soft match are binary. They cannot tell a near miss from a distant one. So we place the prediction and the truth in the taxonomic tree and measure how far apart they sit. Let $d(\cdot)$ be depth from the root and let $\ell = \mathrm{LCA}(\hat{y},y)$ be the lowest common ancestor.
 
\textbf{Mistake severity} counts how many ranks we climb from the truth before we meet the prediction,
\begin{equation}
  \mathrm{sev} \;=\; \mathbb{E}\!\left[\, d(y) - d(\ell) \;\middle|\; \hat{y} \neq y \,\right]
  \label{eq:supp-sev}
\end{equation}
A severity of 1 means the misses stay inside the correct parent. Higher values mean a model misses the broader clade too. We keep severity because it answers a question accuracy cannot. When a model is wrong, how wrong is it. A model that names the wrong frog family scores 1. A model that names a mouse scores 3.
 
\textbf{Wu--Palmer similarity} normalizes the depth of the shared ancestor:
\begin{equation}
  \mathrm{wup}(\hat{y},y) \;=\; \frac{2\,d(\ell)}{d(\hat{y}) + d(y)}
  \label{eq:supp-wup}
\end{equation}
Wu--Palmer asks how deep the agreement runs. Two answers that meet at the genus score higher than two that meet at the class. It turns severity into a bounded similarity, so we can average it across ranks that sit at different depths.

\textbf{Lin similarity} weights the agreement by how rare the shared ancestor is, so agreeing on an obscure clade counts for more than agreeing on a common one. 
\begin{equation}
  \mathrm{lin}(\hat{y},y) = \frac{2\,\mathrm{IC}(\ell)}{\mathrm{IC}(\hat{y}) + \mathrm{IC}(y)},
  \quad
  \mathrm{IC}(c) = -\log\frac{|\mathrm{subtree}(c)|}{N}
  \label{eq:supp-lin}
\end{equation}
Here $N$ is the total node count. Wu--Palmer treats every rank as equally informative. Lin does not. Half our images are chordates, so agreeing on \emph{Chordata} is easier than agreeing on a small genus. Lin computes that difference.
 
\paragraph{Worked example.}
We show a green treefrog and we ask for its species. The answer is \emph{Dryophytes cinereus}. Suppose five models answer.
\begin{table}[h]
  \centering\footnotesize
  \setlength{\tabcolsep}{3.5pt}
  \begin{tabular}{@{}llrrrrr@{}}
    \toprule
    Answer & What it is & $\EM$ & $\EMs$ & sev & wup & lin \\
    \midrule
    \emph{D.\ cinereus}   & the right species     & 1 & 1 & 0 & $1.00$ & $1.00$ \\
    \emph{Hyla cinerea}   & its pre-2016 name     & 0 & 1 & 0 & $1.00$ & $1.00$ \\
    \emph{D.\ gratiosus}  & barking treefrog      & 0 & 0 & 1 & $0.83$ & $0.75$ \\
    \emph{A.\ americanus} & a toad, other family  & 0 & 0 & 3 & $0.50$ & $0.35$ \\
    \emph{Mus musculus}   & a mouse, other class  & 0 & 0 & 5 & $0.17$ & $0.05$ \\
    \bottomrule
  \end{tabular}
\end{table}
We compute Lin-similarity with round subtree counts. We take $N = \num{900000}$ nodes, with \emph{Chordata} covering half the tree, \emph{Anura} covering \num{7500} nodes, \emph{Dryophytes} covering 30, and each species covering one. From the scores, we see that:
\begin{itemize}
    \item \textbf{EM} credits one answer out of five. It calls the pre-2016 name wrong, and it gives the barking treefrog, the toad and the mouse the same zero.
    \item \textbf{EM\textsubscript{soft}} fixes the first problem. \emph{Hyla cinerea} resolves onto \emph{Dryophytes cinereus} in the backbone, so the model gets its point. It is the same animal under an older name. $\EMs$ leaves the other three at zero, because they are different species.
    \item \textbf{Severity} fixes the second problem. The three wrong answers are not equally wrong. One is a treefrog in the same genus. One is a toad in another family. One is a mouse. Severity scores 1, 3, 5. 
    \item \textbf{Wu--Palmer} turns that into a similarity we can average, and it keeps the ordering. 
    \item \textbf{Lin} rescores them based on rarity of the taxon. Wu--Palmer awards the mouse $0.17$ for sharing \emph{Chordata}. But this it is over-credited, becasue half our images are chordates. Lin awards it $0.05$. Lin also pulls the toad down from $0.50$ to $0.35$, because \emph{Anura} is a large clade. It leaves the barking treefrog near the top, because \emph{Dryophytes} is a smaller clade.
\end{itemize}

So the columns answer four different questions: 1. \emph{Is the answer right?} 2. \emph{Is it right under an older name?} 3. \emph{How far up the tree does it miss?} 4. \emph{How much does the part it got right tell us?}

\subsection{\cattr{}: Chance-adjusted graded score}
 
A raw graded score fails \cattr{} in two ways. First, answer spaces have structure. On a \emph{brown wing}, \emph{red} is a better answer than \emph{green}. A binary score calls both wrong. Second, answer spaces differ in size. Bill length offers 3 values and color offers 15. A image-blind model that guesses on a 3-way item scores better than a model that guesses on a 15-way item. \cags{} fixes both. We score similarity with a family kernel, then we subtract what an image-blind guess earns on that same item. Let $s(\hat{y},y) \in [0,1]$ be the kernel and let $b$ be the mean similarity of the item's own distractors to the ground truth,
\begin{equation}
  b \;=\; \frac{1}{|D|}\sum_{d \in D} s(d, y)
  \label{eq:supp-chance}
\end{equation}
$D$ is the set of distractors for a multiple-choice item. For an open-ended item we draw a fixed-seed sample of 40 other ground truth answers from that family. Then
\begin{equation}
  \cags{}_i \;=\; \frac{s_i - b_i}{1 - b_i},
  \qquad
  \overline{\cags{}} \;=\; \frac{1}{N}\sum_{i=1}^{N}\cags{}_i
  \label{eq:supp-cags}
\end{equation}
 
A \cags{} of 1 is a perfect answer. A \cags{} of 0 is exactly the chance level for that item. A negative \cags{} means the answer sits farther from the truth than a random chance.
 
\paragraph{Worked example.}
We ask for a wing color. The truth is \emph{blue} and the distractors are \emph{purple}, \emph{green} and \emph{yellow}. The color kernel in \cref{eq:supp-color} gives $s(\text{purple},\text{blue}) = 0.498$, $s(\text{green},\text{blue}) = 0.134$ and
$s(\text{yellow},\text{blue}) = 0.053$, so $b = 0.228$. 
\begin{table}[h]
  \centering\footnotesize
  \setlength{\tabcolsep}{5pt}
  \begin{tabular}{@{}lrrl@{}}
    \toprule
    Answer & $s$ & \cags{} & Reading \\
    \midrule
    \emph{blue}   & $1.000$ & $1.000$  & correct \\
    \emph{purple} & $0.498$ & $0.350$  & better than a guess \\
    \emph{green}  & $0.134$ & $-0.122$ & worse than a guess \\
    \emph{yellow} & $0.053$ & $-0.227$ & much worse than a guess \\
    \bottomrule
  \end{tabular}
  \refstepcounter{table}
  \label{tab:supp-cags-scoring}
\end{table}

Based on ~\cref{eq:supp-cags} and the $b$ value, we calculate the \cags{}~scores and report them in~\cref{tab:supp-cags-scoring}. We keep the negative numbers and do not clip them at zero. Clipping would put \emph{green} and \emph{yellow} at the same score, and the severity metric in \cref{eq:supp-c2sev} depends on the difference in mistakes. A model that scores negative is confidently wrong, not uncertain. We want to test and report it.

Also, the difficulty of the question lies in the distractor set. When every distractor is close to the truth, $b$ is high and a decent answer gets partial score in \cags{}. When the distractors are far away, $b$ is low and only a good answer gets credited. Exact match cannot do this. We also report the mean raw score $\mathrm{GS}_{\mathrm{raw}}$. The gap $\mathrm{GS}_{\mathrm{raw}} - \overline{\cags{}}$ grows when a family of question is easy to look good on (See~\cref{subsec:supp-metrics-cags-kernel}).
 
\subsection{\cattr{}: The six similarity kernels}
\label{subsec:supp-metrics-cags-kernel}

Every question family has its own idea of a near miss, so each gets its own kernel.

\paragraph{Color.}
We anchor each of the 15 CUB color words to an sRGB triple. We push it through inverse gamma, linear RGB, XYZ under D65, and CIE-Lab. Then we compare two colors with CIEDE2000 $\Delta E_{00}$, which is a perceptual distance and not a distance between the words,
\begin{equation}
  \mathrm{color}(c_1,c_2) =
  \begin{cases}
    1 & c_1 = c_2 \\
    0 & \text{either is iridescent} \\
    \exp(-\Delta E_{00}/\tau) & \tau = 30 \text{ default} \\
    \max(0, 1 - \Delta E_{00}/\Delta E_{\max}) & \text{linear}
  \end{cases}
  \label{eq:supp-color}
\end{equation}
\emph{Iridescent} has no fixed hue, so we score it by exact match only.
 
\begin{table}[h]
  \centering\footnotesize
  \setlength{\tabcolsep}{5pt}
  \begin{tabular}{lrr}
    \toprule
    Pair & exponential & linear \\
    \midrule
    blue, purple  & 0.498 & 0.791 \\
    blue, green   & 0.134 & 0.396 \\
    blue, yellow  & 0.053 & 0.120 \\
    brown, rufous & 0.760 & 0.918 \\
    yellow, orange& 0.507 & 0.796 \\
    black, white  & 0.036 & 0.000 \\
    \bottomrule
  \end{tabular}
\end{table}
 
\paragraph{Ordinal.}
Bill length and size run on an ordered scale, so we use normalized rank distance,
\begin{equation}
  \mathrm{ord}(v_1,v_2) = 1 - \frac{|\mathrm{rank}(v_1) - \mathrm{rank}(v_2)|}{\mathrm{rank}_{\max}}.
  \label{eq:supp-ord}
\end{equation}
Bill length \emph{shorter} against \emph{about the same} scores $0.50$, and \emph{shorter} against \emph{longer} scores $0$. Size \emph{small} against \emph{medium} scores $0.75$, so an off-by-one size error keeps three quarters
of its credit.

\paragraph{Pattern.}
Patterns have no order, so rank distance does not apply. We set the values by hand from one fact. A bird that is spotted, striped or multicolored has a patterned surface. A bird that is \emph{solid} has none. So confusing two patterns is a smaller error than calling a patterned bird solid, so \emph{spotted} for \emph{striped} keeps half its credit and \emph{solid} for \emph{striped} keeps none.
\begin{equation}
  \mathrm{pat}(v_1,v_2) =
  \begin{cases}
    1.0 & v_1 = v_2 \\
    0.5 & v_1, v_2 \in \{\text{spotted, striped, multicolored}\} \\
    0.0 & \text{otherwise} 
  \end{cases}
  \label{eq:supp-pat}
\end{equation}
 
\paragraph{Anatomy.}
Body parts sit at coordinates in a three-dimensional layout. The bird faces right, $y$ runs down, and $z$ is a small lateral offset that separates left from right. Part similarity is normalized proximity in that layout, 
\begin{equation}
  \mathrm{part}(p_1,p_2) = \max\!\left(0,\ 1 - \frac{\lVert p_1 - p_2 \rVert_2}{d_{\max}}\right)
  \label{eq:supp-part}
\end{equation}
where $d_{\max}$ is the layout diameter. We build the layout from the data. For each part we take the median bounding-box-normalized coordinate over every image where CUB records that part as visible.
 
\begin{table}[h]
  \centering\footnotesize
  \setlength{\tabcolsep}{5pt}
  \begin{tabular}{lr}
    \toprule
    Pair & part similarity \\
    \midrule
    nape, crown       & 0.922 \\
    beak, forehead    & 0.857 \\
    left wing, right wing & 0.883 \\
    crown, belly      & 0.430 \\
    beak, tail        & 0.000 \\
    \bottomrule
  \end{tabular}
\end{table}

We use this kernel on anatomical marker localization, where it credits \emph{crown} for \emph{nape} at $0.92$ and \emph{tail} for \emph{nape} at $0$. Exact match gives both a flat zero, which erases the difference between a model that nearly found the marker and one that was lost. The \emph{left wing} and \emph{right wing} pair scores $0.883$ and not $1.000$ because of the $z$ offset. This is a limitation of the \cags{}~metrics which we report in \cref{sec:supp-c2}.
 
\paragraph{Paired attribute.}
A relative salience option names a region and a value together, so a near miss can be in the region, in the value, or in both. We score the two halves separately and combine them,
\begin{equation}
  \mathrm{pair} = \alpha\, g(r_1,r_2) + (1-\alpha)\, v(v_1,v_2)
  \qquad \alpha = 0.5
  \label{eq:supp-pair}
\end{equation}
where $g$ blends the anatomy similarity of the two regions with a property-match indicator, and $v$ dispatches to the color, ordinal or pattern kernel. \emph{Wing color: blue} scores $0.749$ against \emph{wing color: purple}, $0.952$ against \emph{breast color: blue}, and $0.395$ against \emph{tail color: yellow}. Getting the region right and the color close beats getting the color right and the region wrong.
 
\paragraph{Set overlap.}
A visibility answer is a set, so we use a soft Jaccard,
\begin{equation}
  \mathrm{sJac}(A,B) = \frac{\sum_{a \in A}\max_{b \in B}\kappa(a,b)}
                             {|A| + |B| - \sum_{a \in A}\max_{b \in B}\kappa(a,b)}
  \label{eq:supp-sjac}
\end{equation}
with element kernel $\kappa$ equal to exact match by default. \emph{Both} against \emph{Left} scores $0.50$, because the model named one of the two. \emph{Both} against \emph{None} scores $0.00$, and \emph{Left} against \emph{Right} scores $0.00$.

\paragraph{Naming.}
In \cattr{}, three question families ask about the bird's common English name or what type of bird it is, and two about the taxonomy at genus and species ranks (See ~\cref{tab:supp-c2families}). Although the task seems similar to \ctax{}, we do not reuse the \ctax{} tree metrics. The metrices in \ctax{}~use a taxonomic lineage that comes from the TreeofLife-200M dataset. Whereas ground truth answers in \cattr{}~come from CUB-200-2011 dataset and has no information on lineage, so there is no tree to evaluate the \textbf{Wu--Palmer} and \textbf{Lin}. A label in \cattr{}~is a bird name, not a node. So we score the names directly.
 
\paragraph{The CUB name mapping.}
CUB-200-2011 gives a class a common English name. There is no scientific name. So we build a map from each of the 200 CUB classes to its accepted scientific name, we verify every entry against both the iNaturalist taxonomy and Wikidata. We release the map with the benchmark. 

\paragraph{Metrics.} Let $T(\cdot)$ be the set of lowercased tokens in a name. Token $F_1$ is the harmonic mean of token precision and recall:
\begin{equation}
  F_1(p,g) \;=\; \frac{2\,\bigl|T(p) \cap T(g)\bigr|}{\bigl|T(p)\bigr| + \bigl|T(g)\bigr|}
  \label{eq:supp-tokenf1}
\end{equation}
\emph{Song sparrow} against \emph{fox sparrow} shares one token out of two on each side, so $F_1 = 2/4 = 0.5$. We then score the three name forms. Here $\mathrm{head}(\cdot)$ is the last token of a common name, which is the noun the name ends on,
{\footnotesize
\begin{align}
  \mathrm{common}(p,g) &= 0.6\,F_1(p,g) + 0.4\,\mathbf{1}\!\left[\mathrm{head}(p) = \mathrm{head}(g)\right] \notag \\
  \mathrm{sci}(p,g)    &= 0.5\,a_{\mathrm{gen}} + 0.5\,a_{\mathrm{epi}} \label{eq:supp-naming} \\
  \mathrm{genus}(p,g)  &= a_{\mathrm{gen}} \notag
\end{align}}
 
A component that agrees scores 1. A component that disagrees falls back to a down-weighted string similarity, so a wrong genus that looks like the right one earns a partial score:
\begin{equation}
  a_x =
  \begin{cases}
    1 & \text{the two agree on } x \\
    0.3\,\mathrm{ratio}(x_p, x_g) & \text{otherwise}
  \end{cases}
  \label{eq:supp-namefallback}
\end{equation}
where $\mathrm{ratio}$ is the sequence-matching ratio between the two strings. We lift the common-name score to at least $0.75$ when the two names share a genus, because two birds in one genus are close whatever their common names look like.

\begin{table}[h]
  \centering\footnotesize
  \setlength{\tabcolsep}{3.5pt}
  \begin{tabularx}{\columnwidth}{@{}llrl@{}}
    \toprule
    Prediction & Ground truth & Score & Reason \\
    \midrule
    song sparrow & fox sparrow & $0.700$ & $F_1 = 0.5$, same head \\
    american crow & fox sparrow & $0.000$ & no shared token, other head \\
    \emph{M.\ georgiana} & \emph{M.\ melodia} & $0.575$ & genus agrees, epithet fallback \\
    \emph{C.\ brachyrhynchos} & \emph{M.\ melodia} & $0.054$ & both components fallback \\
    \bottomrule
  \end{tabularx}
  \refstepcounter{table}
  \label{tab:supp-naming-example}
\end{table}
 
The last two rows in \cref{tab:supp-naming-example} show the fallback effect. \emph{Melospiza georgiana} keeps $0.5$ from the matching genus and adds $0.5 \times 0.3 \times 0.5 = 0.075$ from the epithet, which gives $0.575$. \emph{Corvus brachyrhynchos} agrees on nothing, so both halves fall back and it lands near zero.

\paragraph{Mistake severity.}
 We average over wrong items only,
\begin{equation}
  \mathrm{Severity} = \frac{1}{N_{\text{wrong}}}\sum_{i\,:\,\EM_i = 0}\left(1 - \cags{}_i\right).
  \label{eq:supp-c2sev}
\end{equation}
A severity at or below 1 means the average miss better than a random guess. A severity above 1 means the average miss lands farther from the truth than a random guess. We report the severity numbers in \cref{sec:supp-c2}.

\subsection{\cspat{}: Spatial reasoning metrics} 

\paragraph{Bounded count agreement.}
A model predicting 14 beetles when the truth is 15 shows better spatial reasoning than one predicting 10. We address this using relative error:
\begin{equation}
  \cas{}(\hat{n}, n) = \max\!\left(0,\ 1 - \frac{|\hat{n} - n|}{n}\right).
  \label{eq:supp-cas}
\end{equation}
We then macro-average the scores across datasets. We do this because dataset sizes vary widely. For instance, the GZCD dataset produces three questions per image. If we micro-average, GZCD dominates 29\% of the final score. Macro-averaging reduces this to a fair 10\%.

\begin{table}[h]
  \centering\footnotesize
  \setlength{\tabcolsep}{5pt}
  \begin{tabular}{rrrr}
    \toprule
    true & pred & \cas{} & $\EM$ \\
    \midrule
    10 & 10 & 1.000 & 1 \\
    10 & 12 & 0.800 & 0 \\
    10 & 15 & 0.500 & 0 \\
    10 & 20 & 0.000 & 0 \\
    10 & 90 & 0.000 & 0 \\
    100 & 95 & 0.950 & 0 \\
    \bottomrule
  \end{tabular}
\end{table}

\paragraph{The image-blind floor and lift.}
Every \cspat{} task has a performance floor. We define this floor using simple rules that do not look at the image, like guessing the most common count or drawing a box over the whole image. Let $\mathcal{G}_0$ be this set of image-blind rules. We define the floor $\phi$ and the resulting lift as:
\begin{equation}
  \phi = \max_{g_0 \in \mathcal{G}_0}\ \mathbb{E}\!\left[S(g_0, y)\right],
  \qquad
  \mathrm{lift} = \frac{S - \phi}{1 - \phi}.
  \label{eq:supp-floor}
\end{equation}
A lift of 0 means the model performs exactly like a blind guess. A negative lift means it performs worse. We calculate this baseline strictly on the rows the model successfully parses. If a model fails to format answers for some rows, we drop those rows and compute a fair baseline on the remainder.

\paragraph{Viewpoint agreement.}
There are 10 possible viewpoint orientation. In \cspat{}, the \emph{right} viewpoint dominates. Because of this skew, raw accuracy is misleading. To address this, we assign partial credit for near misses and average the scores across all ground truth labels. We write $v(y,\hat{y})$ for the graded credit of a predicted viewpoint against a gold one, and $\mathcal{L}'$ for the gold labels present:
\begin{align}
  \mathrm{VAS} &= \frac{1}{n}\sum_i v(y_i,\hat{y}_i), \notag \\
  \mathrm{macroVAS} &= \frac{1}{|\mathcal{L}'|}\sum_{\ell \in \mathcal{L}'}
                        \frac{1}{n_\ell}\sum_{i\,:\,y_i = \ell} v(y_i,\hat{y}_i).
  \label{eq:supp-vas}
\end{align}
This macro-averaging gives rare and common viewpoints equal weight. Next, we correct this score for chance against a model that always guesses the same answer, yielding $\kappa$VAS. This is our headline metric. For example, a rule that always guesses \emph{right} gets a 52\% exact match accuracy, but a 0 score in $\kappa$VAS. Finally, we also measure angular error. Both metrics rank the models identically, confirming our scoring method works.

\paragraph{Absolute distance.}
Models can cheat on distance tasks by memorizing priors. We use three checks to resolve this.
\begin{table}[t]
  \footnotesize
  \setlength{\tabcolsep}{4pt}
  \begin{tabularx}{\columnwidth}{@{} l X r @{}}
    \toprule
    Control & Strategy it catches & Score under it \\
    \midrule
    per-type constant floor
      & one value for length, another for width
      & $\accat{20} = 0.410$ \\
    $\rho_{\text{within}}$
      & any predictor constant within a type
      & $\rho_{\text{within}} = 0$ \\
    compression $\kappa$
      & one remembered size for both types
      & $\kappa \approx 1$ \\
    \bottomrule
  \end{tabularx}
\end{table}
We use $\rho_{\text{within}}$ (Spearman correlation) to rank predictions within each measurement type. If a model just guesses a constant average size for every beetle, it gets a correlation of exactly zero. We use $\kappa$ compression to catch models that use one single size for everything. Finally, we report $\accat{20}$, which is the percentage of answers within 20\% of the true length. Together, these checks ensure a model only gets a high score if it actually measures the image.

\paragraph{Localization.}
We score localization using mean Intersection over Union (IoU) against the ground truth box. We report the share of predictions passing various IoU thresholds ($\accat{.25}$, $\accat{.50}$, $\accat{.75}$, and $\accat{.90}$). Our main metric is \texttt{meanIoU\_all}. We calculate this by multiplying the mean IoU by the model's parse rate. We penalize unparsed answers with a zero score instead of ignoring them. Models output bounding boxes in different formats (like absolute pixels or normalized coordinates). We test four common formats for each model and pick the one that gives the highest score. Getting this format wrong drops a model's score by up to $0.68$ mean IoU, so automatic detection is crucial.

\subsection{Signal detection for \creid{}}

In \creid{}, we ask models to verify if two images show the exact same individual animal. Because VLMs output a simple \emph{Yes} or \emph{No} instead of a similarity score, raw accuracy hides a model's bias. For example, a model that says \emph{Yes} to everything gets 50\% accuracy without looking at the images. We resolve this using signal detection theory. We track the hit rate $H$ (saying yes when they are the same) and the false alarm rate $F$ (saying yes when they are different):
\begin{equation}
  \dpr = z(H) - z(F),
  \qquad
  c = -\tfrac{1}{2}\left[z(H) + z(F)\right],
  \label{eq:supp-dprime}
\end{equation}
where $z$ is the inverse normal cumulative distribution function. We use $\dpr$ to measure how well the model separates the two classes, completely ignoring its answering bias. We use criterion $c$ to measure that bias. A negative $c$ means a model prefers saying yes.

\begin{table}[ht]
  \centering\footnotesize
  \setlength{\tabcolsep}{6pt}
  \begin{tabular}{rrl}
    \toprule
    $\dpr$ & balanced acc. & reading \\
    \midrule
    0.0 & 50\% & no discrimination \\
    0.5 & 60\% & weak \\
    1.0 & 69\% & moderate \\
    2.0 & 84\% & strong \\
    3.0 & 93\% & near ceiling \\
    \bottomrule
  \end{tabular}
\end{table}

If a model always gives the same answer, the math breaks because $z$ diverges. We apply a standard log-linear correction, $(x + 0.5)/(n + 1)$, to the rates unconditionally to prevent this.  To get an overall score across species, we pool all the raw counts first and compute one global $\dpr$. Sometimes we see a negative $\dpr$. This means the model says \emph{Yes} to different individuals more often than to identical ones. We explain this further in \cref{sec:supp-c4}.

\begin{figure}[t]
  \centering
  \includegraphics[width=0.9\columnwidth]{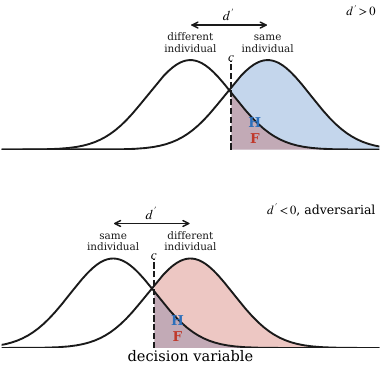}
  \caption{How $\dpr$ and criterion $c$ separate discrimination from bias. A negative $\dpr$ indicates the adversarial condition (lower panel), where the decision variable separates the two classes but points the wrong way.}
  \label{fig:supp-sdt}
\end{figure}

\section{\ctax{} Full Results}
\label{sec:supp-c1}

We evaluate 20 model configurations on an identical set of \num{890000} questions. We test 18 open-weight generalist VLMs and two configurations of the BioCLIP-2 specialist. The tables in this section combine all models unless stated otherwise.

\paragraph{The pooled leaderboard}

\cref{tab:supp-c1lb} ranks the field by $\EMs$.

\begin{table}[htbp]
  \centering\footnotesize
  \setlength{\tabcolsep}{3.5pt}
  \begin{tabular}{rlrrrrrr}
    \toprule
    \# & Model & Img & Ques & $\EM$ & $\EMs$ & wup & lin \\
    \midrule
    \rowcolor{rowtint}
    --- & BioCLIP-2 \texttt{kingdom} & 100K & 890K & .743 & .743 & .854 & .836 \\
    \rowcolor{rowtint}
    --- & BioCLIP-2 \texttt{tol}     & 100K & 890K & .737 & .737 & .836 & .824 \\
    \midrule
      1 & Qwen3-VL-30B-A3B           & 100K & 890K & .503 & .518 & .721 & .702 \\
      2 & Qwen3-VL-32B               & 100K & 890K & .488 & .505 & .702 & .682 \\
      3 & Qwen3-VL-8B                & 100K & 890K & .483 & .497 & .707 & .685 \\
      4 & InternVL3-38B              & 100K & 890K & .455 & .470 & .680 & .661 \\
      5 & Qwen2.5-VL-7B              & 100K & 890K & .445 & .459 & .658 & .638 \\
      6 & Gemma-3-12B                & 100K & 890K & .401 & .429 & .646 & .618 \\
      7 & InternVL3.5-38B            & 100K & 890K & .390 & .416 & .635 & .606 \\
      8 & Gemma-4-12B                & 100K & 890K & .384 & .394 & .609 & .580 \\
      9 & InternVL3-8B               & 100K & 890K & .355 & .373 & .575 & .551 \\
     10 & Qwen2.5-VL-3B              & 100K & 890K & .350 & .367 & .594 & .573 \\
     11 & Llama-3.2-Vision           & 100K & 890K & .338 & .353 & .556 & .529 \\
     12 & InternVL3.5-8B             & 100K & 890K & .325 & .349 & .584 & .552 \\
     13 & Molmo2-8B                  & 100K & 890K & .324 & .349 & .576 & .540 \\
     14 & Pixtral-12B                & 100K & 890K & .329 & .346 & .566 & .533 \\
     15 & InternVL3.5-30B-A3B        & 100K & 890K & .317 & .336 & .571 & .544 \\
     16 & Phi-3.5-Vision             & 100K & 890K & .235 & .245 & .451 & .422 \\
     17 & LLaVA-NeXT-Mistral         & 100K & 890K & .203 & .214 & .419 & .395 \\
     18 & LLaVA-NeXT-Vicuna          & 100K & 890K & .202 & .214 & .393 & .368 \\
    \bottomrule
  \end{tabular}
  \caption{\ctax{} pooled leaderboard, sorted by $\EMs$. Every configuration answers the same \num{890000} questions over the same \num{100000} images, so the rows are directly comparable. The strongest generalist recovers roughly 68\% of the specialist pooled accuracy.}
  \label{tab:supp-c1lb}
\end{table}

\subsection{Granularity collapse}

\cref{tab:supp-c1ranks} shows the mean exact match score at every taxonomic rank.
\begin{table}[htbp]
  \centering\footnotesize
  \setlength{\tabcolsep}{4pt}
  \begin{tabular}{@{}lrl|rl|r@{}}
    \toprule
    & \multicolumn{2}{c}{20 models} & \multicolumn{2}{c}{18 VLMs} & \\
    \cmidrule(lr){2-3}\cmidrule(lr){4-5}
    Rank & $\EM$ & step & $\EM$ & step & gap \\
    \midrule
    phylum          & .748 & ---        & .725 & ---        & .023 \\
    class           & .578 & \dwn{.170} & .545 & \dwn{.180} & .033 \\
    order           & .568 & \dwn{.010} & .541 & \dwn{.004} & .027 \\
    family          & .421 & \dwn{.147} & .382 & \dwn{.159} & .039 \\
    genus           & .292 & \dwn{.129} & .247 & \dwn{.135} & .045 \\
    scientific name & .165 & \dwn{.127} & .123 & \dwn{.124} & .042 \\
    species epithet & .086 & \dwn{.079} & .035 & \dwn{.088} & .051 \\
    \midrule
    common name MCQ & .677 & ---        & .645 & ---        & .032 \\
    common name     & .085 & \dwn{.592} & .039 & \dwn{.606} & .046 \\
    \bottomrule
  \end{tabular}
  \caption{Exact match falls at every step down the hierarchy. The step column gives the change from the row above, in red as the scores fall. The left block pools all 20 configurations, and the right block pools the 18 VLMs. The gap column is the difference between the two, and it widens from $.023$ at phylum to $.051$ at species, so the specialist contributes most exactly where the generalists collapse.}
  \label{tab:supp-c1ranks}
\end{table}
Model accuracy collapses as we move down the taxonomic tree. When we pool all 20 configurations, exact match falls by a factor of $8.7$ from phylum to species epithet. When we look at just the 18 VLMs, it falls by a factor of $20.7$. Every single model shows this exact pattern. Two things cause this. First, the number of possible answers grows rapidly from dozens of phyla to thousands of species. The choice simply becomes much harder. Second, specific species appear much less often in training data than entire phyla do. Models have less evidence to learn from. A species-level question demands the sharpest discrimination from the sparsest data, and these two penalties multiply together. This highlights why we include Lin similarity. Agreeing with the truth at a rare, deep clade contributes more information than agreeing on a broad phylum. Yet, models fail exactly on these high-information, fine-grained tasks.

\begin{figure}[t]
  \centering
  \includegraphics[width=.9\columnwidth]{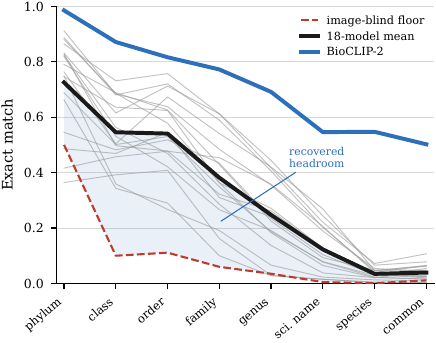}
  \caption{Every model's accuracy falls at every step down the tree. The dashed floor line shows that apparent competence at the phylum level largely comes from guessing common categories, not from actually reading the image.}
  \label{fig:supp-c1collapse}
\end{figure}

\subsection{Coarse accuracy is inflated}
Chordates make up half of the images in \ctax{}. Because of this, a model that never looks at the image and just guesses \emph{Chordata} every time gets a $.500$ score at the phylum rank. Three models actually score below this blind guess: InternVL3.5-30B-A3B ($.486$), Qwen2.5-VL-3B ($.416$), and LLaVA-NeXT-Mistral ($.364$). A blind rule beats them. When we subtract this blind baseline floor, models recover $.450$ of the possible headroom at the phylum level, and $.494$ at the class level. This means models actually do better at class than at phylum in headroom recover. The raw phylum accuracy is inflated partly because the answer pool is so narrow. In contrast, the most common family only covers 6\% of the images. The most common species covers just 0.5\%. At these specific ranks, a blind model scores near zero. Therefore, scoring high at these ranks require finer granularity.

\subsection{Recognition exceeds recall}

\cref{tab:supp-c1recog} compares how well VLMs recognize names versus recall them from scratch.

\begin{table}[t]
  \centering\footnotesize
  \setlength{\tabcolsep}{5pt}
  \begin{tabular}{lrrr}
    \toprule
    Model & open-ended & 4-way MCQ & gap \\
    \midrule
    Qwen3-VL-32B     & .078 & .809 & $+.732$ \\
    Qwen3-VL-8B      & .065 & .775 & $+.710$ \\
    InternVL3-38B    & .044 & .752 & $+.708$ \\
    Qwen2.5-VL-7B    & .062 & .753 & $+.692$ \\
    Qwen3-VL-30B-A3B & .107 & .772 & $+.665$ \\
    \midrule
    18-model mean    & .039 & .645 & $+.606$ \\
    20-config mean   & .085 & .677 & $+.592$ \\
    \bottomrule
  \end{tabular}
  \caption{We show models the identical image and ask for the identical organism. They recognize the correct name among four options far more often than they produce it unprompted. For the 18 VLMs, the ratio of mean accuracies is $16.4\times$.}
  \label{tab:supp-c1recog}
\end{table}
This is a significant effect we see in the benchmark. Past biological VQA works rely heavily on multiple-choice questions. If a benchmark only asks multiple-choice questions, it reports highly inflated scores. Models have the knowledge, but they cannot retrieve it without a hint. These high MCQ scores will immediately crash when users test the models in open-ended real-world use.

\begin{figure}[t]
  \centering
  \includegraphics[width=\columnwidth]{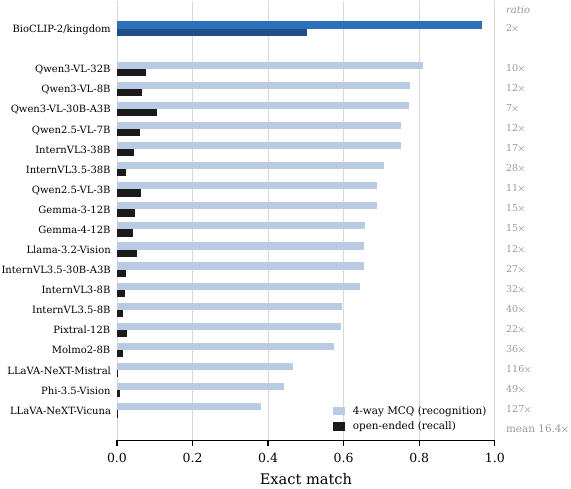}
  \caption{Every model recognizes the right answer from a list vastly better than it produces the name unprompted.}
  \label{fig:supp-c1recogfig}
\end{figure}

\subsection{Scale is not the bottleneck}

Models fail at finer ranks, and scaling it up does not solve this problem. Within the Qwen3-VL family, $\EM$ stays practically flat. It moves from $.483$ at 8B parameters to just $.488$ at 32B. We see this further when we break scores down by rank. When InternVL3 scales from 8B to 38B, it gains $18.7$ points at the class level. But it only gains $6.5$ points at the scientific name level. Extra parameters just help the model better map broad categories it already knows. They do not help it distinguish fine-grained species. \emph{The real bottleneck is the training data. Rare species just do not appear often enough on the web to train effectively, regardless of how large we make the model.}

\subsection{Difficulty stratification}

We group questions into Easy, Medium, and Hard tiers based purely on how often the animal or plant appears in the source catalog. We never define an item as hard just because models fail it. All twenty configurations follow the exact same pattern:
{\footnotesize
\begin{quote}
  $\overline{\EM}_{\text{Easy}}(0.497)
  > \overline{\EM}_{\text{Medium}}(0.342)
  > \overline{\EM}_{\text{Hard}}(0.190)$.
\end{quote}
}
Every single model follows this order. This proves that a species' prevalence in the training data genuinely controls how hard it is to classify, completely independent of the model architecture. \cref{sec:supp-tiers} details this setup.

\subsection{Severity dissociates from accuracy}
 
Severity measures how far a wrong answer lands from the truth, in ranks. We measure that distance through a node, so the model's answer has to name one. A reply of \emph{a small brown bird} names no node. We call that answer \emph{unresolved} and we leave it out of the severity average. Exact match still scores it zero, because a string always compares against a string. Writing $\mathcal{T}$ for the set of nodes in the tree,
\begin{equation}
  \mathrm{sev} = \mathbb{E}\!\left[\, d(y) - d(\ell) \;\middle|\;
                 \hat{y} \neq y,\; \hat{y} \in \mathcal{T} \,\right]
  \label{eq:supp-sevcond}
\end{equation}
\cref{tab:supp-c1sevrank} gives it for every model at every rank, and \cref{tab:supp-c1sevgroup} gives it for every organismal group. The \emph{unres} column in \cref{tab:supp-c1sevrank} shows the percentage of answers we leave out per model. It runs from $.025$ to $.233$ and correlates with exact match at $\rho = -0.854$, so the models that fail to resolve are the models that were going to be wrong anyway. We print it beside severity because a model with a high share is averaged over a smaller sample of its own errors. 

\begin{table*}[t]
  \centering
  \footnotesize
  \setlength{\tabcolsep}{6pt}
  \begin{tabular}{@{}lrrrrrrrrr|r@{}}
    \toprule
    Model & phy & cls & ord & fam & gen & spc & sci & com & mcq & pooled \\
    \midrule
    \rowcolor{rowtint}
    BioCLIP-2 \textit{kingdom}  & 1.00$^\dagger$ & 1.11 & 1.83 & 2.40 & 2.78 & 2.88 & 2.89 & 2.29 & 2.72 & 2.523 \\
    \rowcolor{rowtint}
    BioCLIP-2 \textit{tol}      & 1.00$^\dagger$ & 1.61 & 2.22 & 2.65 & 2.94 & 2.96 & 2.99 & 2.37 & 2.72 & 2.638 \\
    \midrule
    Qwen3-VL-30B-A3B            & 0.53 & 0.72 & 1.60 & 1.84 & \snd{2.06} & \bst{3.47} & \snd{2.48} & \bst{2.71} & \snd{3.04} & 2.472 \\
    Qwen3-VL-32B                & 0.38 & 0.77 & 1.41 & 1.78 & 2.09 & \snd{3.52} & \snd{2.48} & 2.89 & \bst{3.02} & 2.459 \\
    Qwen3-VL-8B                 & 0.61 & \snd{0.69} & 1.42 & 1.80 & \bst{2.04} & 3.65 & \bst{2.46} & \snd{2.85} & \snd{3.04} & 2.480 \\
    InternVL3-38B               & 0.65 & \bst{0.62} & 1.49 & 1.90 & 2.29 & 3.71 & 2.82 & 3.22 & 3.10 & 2.614 \\
    Qwen2.5-VL-7B               & 0.58 & 0.92 & \snd{1.37} & 1.76 & 2.15 & 3.77 & 2.58 & 3.22 & 3.09 & \snd{2.444} \\
    Gemma-3-12B                 & \snd{0.19} & 0.83 & 1.57 & 1.97 & 2.38 & 3.72 & 2.93 & 3.06 & 3.17 & 2.574 \\
    InternVL3.5-38B             & 0.21 & 0.75 & 1.60 & \snd{1.74} & 2.35 & 3.78 & 2.95 & 3.59 & 3.12 & 2.606 \\
    Gemma-4-12B                 & \wst{0.99} & 1.00 & 1.84 & 2.04 & 2.48 & 3.90 & 3.07 & 3.18 & 3.20 & 2.763 \\
    InternVL3-8B                & 0.46 & 0.93 & 1.62 & 1.91 & 2.49 & 3.90 & 3.05 & 3.50 & 3.17 & 2.692 \\
    Qwen2.5-VL-3B               & 0.29 & 0.70 & \bst{1.05} & \bst{1.73} & 2.20 & 3.86 & 2.61 & 2.93 & 3.14 & \bst{2.221} \\
    Llama-3.2-Vision            & \bst{0.16} & 0.70 & 1.43 & 1.77 & 2.32 & 4.05 & 2.75 & 3.26 & 3.18 & 2.493 \\
    InternVL3.5-8B              & 0.28 & 0.91 & 1.49 & 2.01 & 2.60 & 4.09 & 3.21 & 3.62 & 3.22 & 2.740 \\
    Molmo2-8B                   & \snd{0.19} & 0.97 & 1.82 & 2.00 & 2.53 & 4.00 & 3.15 & 3.53 & 3.20 & 2.746 \\
    Pixtral-12B                 & 0.43 & 1.00 & 1.71 & 2.10 & 2.62 & 3.96 & 3.23 & 3.69 & 3.23 & 2.775 \\
    InternVL3.5-30B-A3B         & 0.42 & 0.72 & 1.69 & 1.89 & 2.50 & 3.81 & 3.12 & 3.63 & 3.17 & 2.667 \\
    Phi-3.5-Vision              & 0.83 & \wst{1.31} & 1.65 & \wst{2.30} & \swst{3.12} & 4.02 & \swst{3.91} & 4.01 & \swst{3.34} & \swst{2.992} \\
    LLaVA-NeXT-Mistral          & 0.64 & 0.76 & \swst{1.86} & 2.08 & 3.04 & \swst{4.19} & 3.28 & \swst{4.09} & 3.32 & 2.739 \\
    LLaVA-NeXT-Vicuna           & \swst{0.93} & \swst{1.07} & \wst{2.02} & \swst{2.24} & \wst{3.66} & \wst{4.35} & \wst{4.02} & \wst{4.18} & \wst{3.40} & \wst{3.135} \\
    \bottomrule
  \end{tabular}
  \caption{Mistake severity by rank, with the unresolved share beside it (lower is better). Colors mark the \bst{best}, \snd{second best}, \swst{second worst} and \wst{worst} over the 18 generalists only. Severity is conditional on being wrong and on resolving, so it is not comparable across models with very different accuracy. $^\dagger$BioCLIP-2 carries the worst phylum severity in the table because it is wrong at phylum only $1.5\%$ of the time, and that residual is the hard part.}
  \label{tab:supp-c1sevrank}
\end{table*}
 
\begin{table}[htbp]
  \centering\footnotesize
  \setlength{\tabcolsep}{4pt}
  \begin{tabular}{@{}lrrr@{}}
    \toprule
    Group & $\EM$ & severity & unresolved \\
    \midrule
    Bryophytes  & .246 & \wst{3.396}  & \wst{.154} \\
    Fish        & .274 & \swst{2.966} & .109 \\
    Gymnosperms & .321 & \bst{2.424}  & \swst{.176} \\
    Reptiles    & .323 & 2.463        & .167 \\
    Angiosperms & .401 & \snd{2.423}  & .101 \\
    Amphibians  & .404 & 2.493        & .122 \\
    Arachnids   & .452 & 2.479        & .108 \\
    Mammals     & .496 & 2.483        & .123 \\
    Insects     & .528 & 2.573        & \bst{.095} \\
    Birds       & .549 & 2.531        & \snd{.097} \\
    \bottomrule
  \end{tabular}
  \caption{Exact match, severity and the unresolved share by organismal group, pooled over all 20 configurations and sorted by exact match. Unresolved is the percentage of answers that name no node in the tree, so they carry no severity. The three columns do not track each other. Gymnosperms carry the lowest severity and the second highest unresolved share, so their few measurable misses are close ones. Bryophytes carry the worst severity and the field's lowest accuracy together.}
  \label{tab:supp-c1sevgroup}
\end{table}
 
\paragraph{Severity rises with depth for every model.}
A wrong phylum answer lands within one rank of the truth for all 18 generalists, from $0.16$ to $0.99$. A wrong species answer lands three to four ranks up, from $3.47$ to $4.35$. So a model that misses the species does not miss it by a sibling. It misses it by a family or an order. The species column is also the narrowest, with a spread of $0.88$ across the field against $2.05$ at genus.
 
\paragraph{Severity is conditional on being wrong.}
BioCLIP-2 has the worst phylum severity in the table at $1.00$, worse than every generalist except Gemma-4-12B. That is not a defect. BioCLIP-2/kingdom answers phylum correctly $98.5\%$ of the time, so its severity is computed on a residual of $1.5\%$ of items, and those are the genuinely hard ones. Llama-3.2-Vision is wrong at phylum $45\%$ of the time and averages $0.16$, because most of its errors are near misses inside a large easy set. We therefore read severity within a model across ranks, or between models of similar accuracy, and not as a standalone ranking. \cref{tab:supp-c1sevrank} prints the pooled column beside the per-rank columns for that reason.
 
\paragraph{Accuracy and severity rank models differently.}
The rank correlation between $\EM$ and severity is $\rho\!\left(\EM,\ \mathrm{sev}\right) = -0.662$, which means they order the models' performance differently. Qwen2.5-VL-3B is the clearest case of this. It has the best pooled severity in the field at $2.221$ and the best severity at order and at family, while ranking 12th of 18 on $\EM$. It is bad at being right and good at being wrong in the right neighborhood, and it moves 11 places between the two orderings. The opposite case is Phi-3.5-Vision, which is second worst pooled at $2.992$ and worst at class, family and scientific name. The same signature appears in \cref{sec:supp-c2}, where Phi is the only model whose \cattr{} severity exceeds 1. It is a model that is wrong confidently in both components. Wu--Palmer and Lin correlate with $\EM$ at $\rho = 0.959$ and $\rho = 0.965$ respectively, so they reorder nothing from the $\EM$ rank.
\begin{figure}[htbp]
  \centering
  \includegraphics[width=0.9\columnwidth]{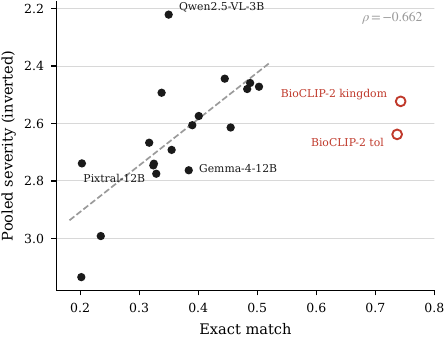}
  \caption{Accuracy and severity measure different things. A model can be wrong
  often and still miss by a close relative.}
  \label{fig:supp-c1sev}
\end{figure}

\subsection{Taxonomic revision lag}
Biologists often rewrite the taxonomy tree and rename groups. When they do, older textbooks and training data become outdated. We use the gap between soft exact match ($\EMs$) and strict exact match ($\EM$) to measure this. \cref{tab:supp-c1lag} gives the gap by rank on the left and by organismal group on the right.
\begin{table*}[tbp]
  \centering\footnotesize
  \setlength{\tabcolsep}{8pt}
  \begin{tabular}{@{}lrrl@{\hspace{3em}}|@{\hspace{3em}}lrrl@{}}
    \toprule
    Rank & $\EM$ & $\EMs$ & gap & Group & $\EM$ & $\EMs$ & gap \\
    \midrule
    class           & .578 & .658 & \upp{.080} & Reptiles    & .323 & .422 & \upp{.100} \\
    phylum          & .748 & .789 & \upp{.041} & Angiosperms & .401 & .432 & \upp{.031} \\
    scientific name & .165 & .172 & \upp{.007} & Gymnosperms & .321 & .339 & \upp{.018} \\
    family          & .421 & .425 & \upp{.004} & Bryophytes  & .246 & .250 & \upp{.004} \\
    genus           & .292 & .295 & \upp{.003} & Fish        & .274 & .277 & \upp{.003} \\
    order           & .568 & .570 & \upp{.002} & Amphibians  & .404 & .407 & \upp{.002} \\
    species epithet & .086 & .086 & ---        & Arachnids   & .452 & .453 & \upp{.001} \\
    common name     & .085 & .085 & ---        & Birds       & .549 & .550 & \upp{.001} \\
    common name MCQ & .677 & .677 & ---        & Insects     & .528 & .528 & \upp{.001} \\
    \emph{all ranks} & .400 & .416 & \upp{.015} & Mammals     & .496 & .498 & \upp{.001} \\
    \bottomrule
  \end{tabular}
  \caption{Taxonomic revision lag, pooled over all 20 models. The gap is $\EMs - \EM$, so it is always a gain and we mark it with a teal arrow. Both panels are sorted by gap. The two panels line up. Reptiles carry the largest group gap and \emph{Reptilia} dissolved at class rank, which is where the largest rank gap sits. Angiosperms and Gymnosperms carry the next two and their revision moved the phylum from \emph{Magnoliophyta} to \emph{Tracheophyta}, which is the second largest rank gap. A dash means the gap rounds to zero.}
  \label{tab:supp-c1lag}
\end{table*}
Our observations are: 
\begin{itemize}
    \item \textbf{Class carries the largest gap at $.080$.} \emph{Reptilia} is a class-rank name and the backbone dissolved it into four modern classes. Reptiles carry the largest group gap at $.100$, three times the next group. A model answering \emph{Reptilia} for a snake is repeating textbook Linnaean taxonomy, and our strict backbone marks it wrong.
    \item \textbf{Phylum carries the second largest at $.041$.} The plant revision moved the phylum from \emph{Magnoliophyta} to \emph{Tracheophyta}. Angiosperms and Gymnosperms carry the second and third largest group gaps at $.031$ and $.018$.
    \item \textbf{Everything below order is flat.} From family down the gap never exceeds $.004$, and at the species epithet, the common name and the MCQ it rounds to zero. Nothing was renamed there, so there is nothing to credit. The gap is not a general leniency in the metric. It appears only where the taxonomy moved.
\end{itemize}
We verify every soft credit against GBIF with a separate audit script. All of roughly \num{130000} soft credits resolve to a valid old-to-new substitution and none is flagged for review. So this $\EMs$--$\EM$ gap measures how old a model's training data is, not an actual failure to identify the animal.

\subsection{Cross-group generality}

We check if models that classify birds well also classify fish well. They do. We rank the models for each of the 10 organism groups. We then compare these rankings using Spearman rank correlation. We summarize the resulting $10 \times 10$ matrix by its mean off-diagonal value. \cref{tab:supp-c1groupcorr} gives the full matrix.
 
\begin{table*}[tbp]
  \centering\footnotesize
  \setlength{\tabcolsep}{3.5pt}
  \begin{tabular}{@{}lrrrrrrrrrr|r@{}}
    \toprule
    & Amph & Angio & Arach & Birds & Bryo & Fish & Gymno & Insect & Mamm & Rept & mean \\
    \midrule
    Amphibians  & 1.00 & .94 & .82 & .90 & .92 & .88 & \wst{.39} & .89 & .95 & .98 & .852 \\
    Angiosperms & .94 & 1.00 & .87 & .88 & .96 & .89 & \wst{.49} & .91 & .92 & .94 & .867 \\
    Arachnids   & .82 & .87 & 1.00 & .90 & .87 & .81 & \wst{.41} & .94 & .83 & .83 & \swst{.809} \\
    Birds       & .90 & .88 & .90 & 1.00 & .91 & .87 & \wst{.42} & .97 & .91 & .89 & .850 \\
    Bryophytes  & .92 & .96 & .87 & .91 & 1.00 & .94 & \wst{.46} & .95 & .89 & .91 & .868 \\
    Fish        & .88 & .89 & .81 & .87 & .94 & 1.00 & \wst{.37} & .91 & .83 & .88 & .820 \\
    Gymnosperms & \wst{.39} & \wst{.49} & \wst{.41} & \wst{.42} & \wst{.46} & \wst{.37} & 1.00 & \wst{.45} & \wst{.46} & \wst{.45} & \wst{.433} \\
    Insects     & .89 & .91 & .94 & .97 & .95 & .91 & \wst{.45} & 1.00 & .88 & .90 & .867 \\
    Mammals     & .95 & .92 & .83 & .91 & .89 & .83 & \wst{.46} & .88 & 1.00 & .95 & .847 \\
    Reptiles    & .98 & .94 & .83 & .89 & .91 & .88 & \wst{.45} & .90 & .95 & 1.00 & .859 \\
    \bottomrule
  \end{tabular}
  \caption{Spearman rank correlation between organismal groups. The mean column averages the nine off-diagonal cells in each row. Red marks the \wst{Gymnosperms} row and column, which hold the lowest value in every row of the table without exception, and \swst{Arachnids} carries
  the next lowest mean, but almost $2\times$ Gymnosperms. The whole matrix averages $0.807$. If we drop Gymnosperms, it jumps to $0.901$, so one group costs the benchmark almost a tenth of its cross-group agreement.}
  \label{tab:supp-c1groupcorr}
\end{table*}
Our observations are:
\begin{itemize}
    \item \textbf{Taxonomic ability is mostly one-dimensional.} If we ignore Gymnosperms, every pairwise correlation sits between $0.81$ and $0.98$. Models that rank highly on amphibians also rank highly on mammals, fish, and birds. A single underlying taxonomic ability drives the leaderboard. The specific clade barely matters. We use this strong correlation as our baseline when we analyze cross-task abilities in \cref{sec:supp-cross}.
    \item \textbf{Gymnosperms break the pattern.} Every correlation involving Gymnosperms collapses to roughly $0.4$. This is not noise. The Gymnosperm label space mostly consists of very similar-looking conifer trees. The specialist model handles this highly repetitive visual space very differently than generalists do. This difference scrambles the leaderboard for this group alone. We report it instead of smoothing it away.
\end{itemize}

\begin{figure}[t]
  \centering
  \includegraphics[width=.9\columnwidth]{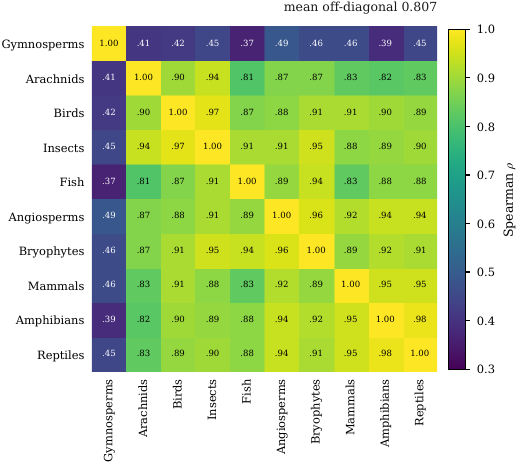}
  \caption{Model rankings remain nearly identical across nine of the ten organism groups. Gymnosperms are the single exception.}
  \label{fig:supp-c1heat}
\end{figure}

\subsection{Difficulty tiers}

We rank question difficulty by how common the target taxon is. We count how many times it appears in the TreeOfLife-200M database. Let $n(t)$ be its image count, and $\mathcal{T}_r$ be all taxa at rank $r$. We calculate its prevalence percentile as:
\begin{equation}
  P(t) = \frac{\left|\{\, u \in \mathcal{T}_r : n(u) \le n(t) \,\}\right|}{|\mathcal{T}_r|}.
  \label{eq:supp-prev}
\end{equation}
If a taxon sits in the top third ($P(t) > 0.67$), we label it Easy. If it sits in the bottom third ($P(t) \le 0.33$), we label it Hard. We label the remainder Medium.

\subsection{Non-English common names.}
The GBIF Backbone \texttt{vernacularName} field is multilingual by specification. TreeOfLife-200M inherits the field unchanged, and the download
surfaces no per-record language indicator. So a non-trivial share of our common-name questions arrive with a Swedish, Danish or Norwegian ground truth. It breaks the questions two ways. A model that recognizes the organism and answers \emph{Moor frog} is marked wrong against a ground truth of \emph{{\aa}kergroda}, so the item stops measuring recognition. And a foreign string sitting in an otherwise English option set is an odd one out, so a model can pick it without looking at the image. The second failure is the worse one, because it turns a visual item into a textual one.
\begin{table}[htbp]
  \centering\footnotesize
  \setlength{\tabcolsep}{2pt}
  \begin{tabular}{@{}llll@{}}
    \toprule
    Scientific name & As shipped & Lang. & Resolved \\
    \midrule
    \emph{Juniperus communis}  & \textit{en}                & sv/da & Common juniper \\
    \emph{Picea abies}         & \textit{gran}              & sv & Norway spruce \\
    \emph{Pinus sylvestris}    & \textit{tall}              & sv & Scots pine \\
    \emph{Rana arvalis}        & \textit{{\aa}kergroda}     & sv & Moor frog \\
    \emph{Rana arvalis}        & \textit{Spidssnudet fr{\o}}& da & Moor frog \\
    \emph{Bufotes viridis}     & \textit{gr{\o}nfl{\"a}ckig padda} & sv & European green toad \\
    \emph{Anguis fragilis}     & \textit{kopper{\"o}dla}    & sv & Slow worm \\
    \bottomrule
  \end{tabular}
  \caption{Contaminated ground truth labels and their resolutions.}
  \label{tab:supp-c1lang}
\end{table}
 
We resolve each affected taxon through three stages and each taxon leaves the pipeline as soon as one stage settles it. First we triage offline. A GBIF language tag of \texttt{en} accepts a name, a name carrying non-ASCII letters is rejected before any other test, and otherwise we score the fraction of tokens that appear in an English wordlist. We reject before we accept here on purpose. A foreign name wrongly accepted poisons the answer key, while a false rejection only costs one lookup.
 
Second, we query authorities in order, taking the first clean English candidate. iNaturalist gives a preferred common name, GBIF gives English vernaculars after following synonyms to the accepted name, and Wikipedia's lead sentence is the last resort. Every candidate passes a gate that rejects rank words, taxonomic qualifiers and echoes of the scientific name, so a doubtful candidate is deferred rather than accepted.
 
Third, a curator resolves the residual by hand with a cited source. These are obscure invertebrates and newly described species with no English name in any authority. Because the stages are sequential and disjoint, no taxon appears in two tiers, and the merge produces zero conflicts.

\section{Difficulty Stratification}
\label{sec:supp-tiers}

Every \ctax{} question carries a difficulty tier derived strictly from the taxonomy itself, rather than from empirical model performance. We compute this difficulty once over the union of all answer keys and propagate it identically to every model's prediction file, ensuring all models are evaluated against an identical stratification.

\subsection{Design principles}

\noindent\textbf{Difficulty is an intrinsic property.}
We compute tiers solely from the taxonomy and the answer key. No prediction file influences any tier, which prevents the circularity of defining an item as hard simply because models struggle with it.

\noindent\textbf{Unified computation.}
We assemble a corpus of unique group, image, and question triples across all matched files prior to scoring. We then compute one tier per item and write it to all files containing that item. A model with missing predictions cannot inadvertently shift a boundary and receive a different stratification from its peers.

\noindent\textbf{Taxon-deterministic tiers.}
Two items sharing a target taxon always receive the same tier. Because a taxon never straddles a tier boundary, per-tier accuracy reliably reflects performance over a set of taxa rather than an arbitrary partition of individual items.

\noindent\textbf{Exhaustive, audited resolution.}
Every item resolves to a specific node in the taxonomy. We record the resolution mechanism per item, allowing the proportion of non-direct resolutions to be explicitly reported.

\subsection{Two axes for two formats}

Because the two question formats measure distinct competencies, we evaluate them along two different axes. \Cref{tab:supp-tieraxes} gives the axis used for each format.

\begin{table}[t]
  \centering
  \footnotesize
  \setlength{\tabcolsep}{4pt}
  \begin{tabular}{@{}llp{0.42\linewidth}@{}}
    \toprule
    Format      & Axis          & Question it answers \\
    \midrule
    Open recall & Prevalence    & How much of the visual world does this taxon occupy? \\
    4-way MCQ   & Confusability & How close is the nearest wrong option in the tree? \\
    \bottomrule
  \end{tabular}
  \caption{The difficulty axis used for each question format. Prevalence is unsuitable for a choice item: recognizing a common organism among three close relatives is harder than recognizing a rare organism among three distant relatives.}
  \label{tab:supp-tieraxes}
\end{table}

\subsection{The prevalence tier}

For a given target taxon $t$, let $c(t)$ denote the number of images whose lineage passes through taxon $t$. Exposure is the number of catalog images beneath it:
\begin{equation}
  c(t) = \left|\{i : t \in \operatorname{lineage}(i)\}\right|,
  \qquad
  x(t) = \log_{10}\!\left(c(t)+1\right).
  \label{eq:supp-exposure}
\end{equation}
We report exposure on a logarithmic axis because catalog abundance is heavily right-tailed. The most-photographed species appear in $10^{5}$ to $10^{6}$ images, whereas the rarest appear in only a handful --a span of five to six orders of magnitude. Difficulty does not scale linearly with this count: the jump from 10 to 100 images represents the same multiplicative change in rarity as the jump from \num{10000} to \num{100000}, and this multiplicative scale better reflects the volume of visual evidence a model has plausibly encountered. The base-10 logarithm maps each order of magnitude to one unit on the axis, while the unit pseudocount ensures the transformation remains defined for taxa with empty subtrees. The tertile cut below is rank-based and therefore invariant to this monotone transform; $x(t)$ is the quantity we report and use wherever a continuous prevalence covariate is required, and on a linear axis that display would collapse almost all taxa into an indistinguishable low-count mass.

We then divide taxa into tertiles within a given reference cell. Distinct taxa are sorted by exposure (with taxon names breaking ties), and each taxon is assigned an item-weighted mid-rank percentile:
\begin{equation}
  p(t) = \frac{\mathrm{cum}(t) + \tfrac{1}{2}\, n(t)}{N_{\text{cell}}}
  \label{eq:supp-percentile}
\end{equation}
where $\mathrm{cum}(t)$ is the sum of item counts for rarer taxa, $n(t)$ is the number of items targeting $t$, and $N_{\text{cell}}$ is the total item count in the cell. A taxon with $p(t) < 1/3$ is Hard, one with $1/3 \le p(t) < 2/3$ is Medium, and the remainder are Easy; every item inherits the tier of its target taxon.

Two structural requirements dictate this formulation. First, a tier must represent a set of taxa rather than a mere partition of items; thus, each taxon must receive exactly one percentile. Second, the three strata must contain comparable item mass to preserve statistical power for per-tier accuracy estimates. Assigning each taxon the midpoint of its item-mass block satisfies both criteria. Using the midpoint rather than the leading edge prevents the boundary-splitting failures typical of naive cuts on sorted lists, and the item-count weighting guarantees balanced strata.

\subsection{Reference cells and thin cells}
We define reference cells by organismal group and rank class. The rank class consolidates three species-level wordings (epithet, scientific name, and common name) to ensure difficulty reflects the organism itself, not the phrasing of the question.

Stratifying within a rank prevents the confounding effect where higher ranks inherently possess larger subtree counts (e.g., a class always contains more images than any of its constituent families). Stratifying within a group ensures that \emph{rare} is defined relative to the specific peers a model is tasked with distinguishing.

A tertile split is only statistically meaningful when a cell contains enough distinct taxa. However, some cells are structurally thin. For instance, within a single group, every phylum answer might be identical, and class answers might span only two or three values.

If a cell contains fewer than six distinct taxa, we escalate the stratification, re-cutting against that rank class pooled across all groups. Because exposure is already a global metric, this pooled cut provides a coherent reference population. A cell is only labeled as degenerate if the pooled cell also proves too thin. \Cref{tab:supp-tierescalate} details the escalation protocol for a thin cell.

\begin{table}[t]
  \centering
  \footnotesize
  \setlength{\tabcolsep}{4pt}
  \begin{tabular}{@{}lll@{}}
    \toprule
    Cell                 & Distinct taxa      & Outcome \\
    \midrule
    (Birds, phylum)      & 1, \emph{Chordata} & escalate \\
    (phylum, all groups) & a handful          & \textbf{degenerate} \\
    \addlinespace
    (Birds, class)       & 1, \emph{Aves}     & escalate \\
    (class, all groups)  & many               & tertiles, global \\
    \bottomrule
  \end{tabular}
  \caption{The escalation protocol for resolving thin cells prevents the fabrication of non-existent contrasts. Because every bird belongs to the phylum \emph{Chordata}, and the entire benchmark spans only a few phyla, there is genuinely no rare-versus-common phylum contrast to measure within birds.}
  \label{tab:supp-tierescalate}
\end{table}

The degenerate label marks a structural reality, not a pipeline failure. It indicates items for which no meaningful prevalence contrast exists. These items still possess resolved targets and exposure counts, but are excluded from prevalence-stratified analyses because sorting two or three taxa into three tiers yields no actionable insight. All escalations and degenerate cell designations are recorded in the run log.

\subsection{The confusability tier}
Let $d(\cdot)$ denote node depth, with the root at depth $0$, and let $D$ be the set of distractors for an item ($\lvert D \rvert = 3$). For a ground-truth node $g$ and each distractor $j \in D$, the nearest-distractor distance (NDD) is
\begin{equation}
  d_j = d(g) - d\!\bigl(\mathrm{LCA}(g,j)\bigr),
  \qquad
  \mathrm{NDD} = \min_{j \in D} d_j 
  \label{eq:supp-ndd}
\end{equation}
NDD is measured in edges rather than named ranks. An NDD of at most $1$, in practice exactly $1$, since two distinct nodes differ by at least one edge is Hard, an NDD of $2$ is Medium, and an NDD of at least $3$ is Easy. For the species-level targets that dominate the benchmark these correspond to a congeneric, a con-familial, and a more distant nearest distractor respectively; because vernacular targets may sit at coarser ranks
(\cref{sec:supp-vernacular}), we additionally emit the rank at which the nearest distractor meets the answer, so that an edge distance can always be read back as a rank.

The depth of the lowest common ancestor (LCA) serves as the standard metric for confusion severity. Two taxa meeting only at the root are maximally dissimilar, whereas two meeting at the genus are closely related. Taking the minimum distance across all three distractors enforces a weakest-link model for multiple-choice items: a single genus-level distractor renders the entire question Hard, even if the other two distractors belong to a different phylum, because the model must still execute fine-grained discrimination to exclude that near option. Using the maximum or mean distance would allow easy distractors to artificially mask a hard one.

To complement this tier, we introduce a continuous confusability score where each additional rank of separation halves the credit:

\begin{equation}
  \mathrm{confusability}
  = \frac{1}{\lvert D \rvert} \sum_{j \in D} \exp\!\bigl(-\ln 2 \cdot d_j\bigr)
  = \frac{1}{\lvert D \rvert} \sum_{j \in D} 2^{-d_j}.
  \label{eq:supp-confusability}
\end{equation}

This geometric decay matches the intuition that taxonomic proximity scales multiplicatively. Averaging the contributions provides a comprehensive profile summary that complements the NDD floor summary. Analyses requiring a continuous covariate should use this confusability score, while categorical analyses use the three-level NDD tier.

\paragraph{Worked example.}
Consider an item whose answer is \emph{Zootoca vivipara} at depth $6$, with three distractors: one that shares its family, one that shares only its class, and one from a different phylum (\cref{tab:supp-worked}). The NDD is $2$ (Medium), yielding a confusability score of $0.110$. If the first distractor is replaced by a congener, the LCA moves to the genus level (depth $5$), the distance $d_j$ drops to $1$, and its contribution rises to $0.500$. The NDD drops to $1$, pushing the item into the Hard tier, and confusability increases to $0.193$. This demonstrates how a single close negative appropriately flips the tier under the weakest-link minimum constraint.

\begin{table}[t]
  \centering
  \footnotesize
  \setlength{\tabcolsep}{5pt}
  \begin{tabular}{@{}llrrr@{}}
    \toprule
    Distractor   & LCA    & depth & $d_j$ & $2^{-d_j}$ \\
    \midrule
    same family  & family & 4 & 2 & 0.250 \\
    same class   & class  & 2 & 4 & 0.063 \\
    other phylum & root   & 0 & 6 & 0.016 \\
    \bottomrule
  \end{tabular}
  \caption{Worked example for a ground truth node \emph{Zootoca vivipara} at depth $6$. The nearest distractor sits two edges away, so the item is Medium under the weakest-link NDD rule and scores $0.110$ on the continuous axis.}
  \label{tab:supp-worked}
\end{table}

\subsection{Target resolution}

Every item resolves to exactly one node. Because each image's answer key provides its complete lineage, resolution does not rely on a single answer string being independently discoverable. We apply a four-step protocol and record the step that successfully resolves the target. \Cref{tab:supp-resolve} outlines the four steps alongside examples.

\begin{table}[t]
  \centering
  \footnotesize
  \setlength{\tabcolsep}{4pt}
  \begin{tabular}{@{}lp{0.34\linewidth}p{0.28\linewidth}@{}}
    \toprule
    Step                 & Rule & Example \\
    \midrule
    \texttt{exact}       & the answer names a node at the asked rank
                         & \emph{Lacertidae} as family \\
    \texttt{vernacular}  & a common name through the vernacular index
                         & \emph{Orb weavers} to Araneidae \\
    \texttt{lineage}     & the asked rank read off this image's lineage
                         & \emph{vivipara} through the image \\
    \texttt{anchor}      & lowest resolvable node in the lineage
                         & an order question with a null order \\
    \bottomrule
  \end{tabular}
  \caption{The four-step resolution protocol. The lineage step provides essential robustness; for example, a question asking for an organism's class resolves through the organism's full metadata rather than just the isolated string.}
  \label{tab:supp-resolve}
\end{table}

\subsection{Vernacular names and rank heterogeneity}
\label{sec:supp-vernacular}

Common names do not function as a strict parallel nomenclature for species. In biodiversity corpora, they denote taxa at whatever rank a naturalist finds useful, and this rank varies heavily by clade. While birds and mammals almost universally possess species-level vernaculars, arachnids and bryophytes frequently do not; their familiar names often refer to families or orders (\eg, \emph{Orb weavers} maps to Araneidae, \emph{Long-jawed orb weavers} to Tetragnathidae, and \emph{Daddy longlegs} to an entire order). A resolution framework that naively maps one common name to one species would systematically misclassify items across entire phyla.

To address this, we derive the vernacular index dynamically from the corpus. Each image contributes a pair comprising its common-name answer and its resolved species. A name observed alongside a single species resolves directly to that species. A name observed across multiple species resolves to their lowest common ancestor, accurately reflecting the broader taxon the name denotes in practice. 

This data-driven approach yields three key properties. First, rank is inferred rather than strictly asserted: the index discovers that a name is family-level because the corpus uses it as such. Second, exposure accurately matches the question; a family-level name resolves to a family node and inherits the family's broader prevalence, correctly reflecting that identifying a spider family is easier than identifying a specific species. Third, distractors resolve successfully, as MCQ distractors are drawn from the same pool of common names as the correct answers.

Consequently, the species-level reference cell inherently includes a small number of coarser-rank targets that carry larger exposures. This heterogeneity is deliberate, capturing a genuine difference in task difficulty while remaining strictly within the phrasing constraints of the original question.

\paragraph{Output columns.}
We append 13 metadata columns to each processed file: the resolved target taxon, its rank, the reference cell, the resolution mechanism, the exposure count, the $\log_{10}$ exposure, the prevalence percentile, the prevalence tier, the nearest-distractor distance (NDD), the taxonomic rank at which the NDD meets the answer, the continuous confusability score, the MCQ tier, and a unified difficulty column (which applies the MCQ tier for choice items and the prevalence tier for open-recall items).

\section{\cattr{} Full Results}
\label{sec:supp-c2}

We report \cags{} and $\EM$ metrics alongside one another throughout this section. $\EM$ serves as a conservative anchor, while \cags{} acts as a finely-grained diagnostic instrument. Their agreement validates the overall model ranking, and their divergences help localize specific model behaviors.

\subsection{\cags{} leaderboard}

\cref{tab:supp-c2lb} ranks the field by \cags{}. Performance across the evaluated models ranges from $.378$ to $.593$, demonstrating an absence of ceiling saturation, an essential characteristic of a robust benchmark. The graded metric immediately surfaces three behavioral anomalies, which we expand upon below.

\begin{table}[htbp]
  \centering\footnotesize
  \setlength{\tabcolsep}{3pt}
  \begin{tabular}{@{}rlrrrcr@{}}
    \toprule
    \# & Model & $\EM$ & GS$_{\text{raw}}$ & \cags{} & 95\% CI & sev \\
    \midrule
     1 & Qwen3-VL-32B       & .570 & .687 & \textbf{.593} & .592--.594 & 0.946 \\
     2 & Qwen3-VL-30B-A3B   & .538 & .665 & .569 & .568--.570 & 0.933 \\
     3 & InternVL3-38B      & .535 & .665 & .568 & .567--.569 & 0.929 \\
     4 & Gemma-4-12B        & .526 & .651 & .554 & .553--.555 & 0.941 \\
     5 & Qwen3-VL-8B        & .528 & .654 & .553 & .551--.554 & 0.947 \\
     6 & InternVL3.5-38B    & .528 & .648 & .548 & .546--.549 & 0.959 \\
     7 & Molmo2-8B          & .520 & .634 & .533 & .532--.534 & 0.974 \\
     8 & Qwen2.5-VL-7B      & .497 & .635 & .529 & .528--.530 & 0.936 \\
     9 & InternVL3.5-30B    & .501 & .631 & .524 & .523--.525 & 0.954 \\
    10 & InternVL3-8B       & .486 & .619 & .507 & .506--.508 & 0.959 \\
    11 & InternVL3.5-8B     & .461 & .612 & .505 & .504--.506 & 0.919 \\
    12 & Gemma-3-12B        & .468 & .605 & .487 & .486--.488 & 0.965 \\
    13 & Qwen2.5-VL-3B      & .441 & .586 & .461 & .460--.462 & 0.964 \\
    14 & Pixtral-12B        & .414 & .565 & .439 & .438--.440 & 0.956 \\
    15 & LLaVA-NeXT-Vicuna  & .367 & .541 & .419 & .418--.420 & 0.918 \\
    16 & Phi-3.5-Vision     & .415 & .535 & .405 & .403--.406 & \textbf{1.018} \\
    17 & Llama-3.2-Vision   & .376 & .527 & .382 & .380--.383 & 0.990 \\
    18 & LLaVA-NeXT-Mistral & .376 & .517 & .378 & .377--.379 & 0.997 \\
    \bottomrule
  \end{tabular}
  \caption{\cattr{} leaderboard over \num{824104} questions per model. Intervals
  are 95\% bootstrap over $B=1000$ resamples of the per-item \cags{}. Adjacent
  ranks separated by less than $0.001$ are statistical ties. Phi-3.5-Vision is
  the only model with severity above 1 and the only one whose \cags{} falls below
  its $\EM$.}
  \label{tab:supp-c2lb}
\end{table}

\subsection{\text{EM} leaderboard and format diagnostics}

\cref{tab:supp-c2em} provides the $\EM$ scores alongside three format-specific diagnostics.

\begin{table}[htbp]
  \centering\footnotesize
  \setlength{\tabcolsep}{2.5pt}
  \begin{tabular}{@{}lrrrrrr@{}}
    \toprule
    Model & Acc & MCQ & Open & parse & mism. & choose A \\
    \midrule
    Qwen3-VL-32B       & 57.31 & 57.63 & 51.96 & 100.0 & 0.00 & 19.15 \\
    Qwen3-VL-30B-A3B   & 54.08 & 54.82 & 41.53 & 100.0 & 0.03 & 24.30 \\
    InternVL3-38B      & 53.70 & 54.42 & 41.72 & 100.0 & 0.02 & 17.08 \\
    Qwen3-VL-8B        & 52.98 & 53.92 & 37.11 & 100.0 & 0.04 & 14.56 \\
    InternVL3.5-38B    & 52.88 & 54.39 & 27.57 & 100.0 & 0.02 & 22.85 \\
    Gemma-4-12B        & 52.71 & 55.06 & 13.33 &  98.90 & 0.01 & 17.00 \\
    Molmo2-8B          & 52.09 & 54.54 & 10.94 & 100.0 & 0.02 & 21.41 \\
    InternVL3.5-30B    & 50.22 & 51.46 & 29.49 & 100.0 & 0.02 & 27.41 \\
    Qwen2.5-VL-7B      & 49.99 & 50.04 & 49.05 &  99.93 & \textbf{1.17} & 13.98 \\
    InternVL3-8B       & 48.69 & 49.68 & 32.14 &  99.85 & 0.16 & 17.87 \\
    Gemma-3-12B        & 46.96 & 48.46 & 21.79 & 100.0 & 0.02 & 14.63 \\
    InternVL3.5-8B     & 46.15 & 47.67 & 20.68 & 100.0 & 0.01 & 15.35 \\
    Qwen2.5-VL-3B      & 44.44 & 44.43 & 44.48 & 100.0 & 0.05 & 10.77 \\
    Pixtral-12B        & 41.51 & 42.72 & 21.21 & 100.0 & 0.10 & 20.83 \\
    Phi-3.5-Vision     & 41.48 & 43.73 &  3.81 & 100.0 & 0.06 & 17.50 \\
    LLaVA-NeXT-Mistral & 37.65 & 39.69 &  3.36 & 100.0 & 0.21 & 33.96 \\
    Llama-3.2-Vision   & 37.55 & 38.04 & 29.33 &  99.96 & 0.02 & \textbf{63.26} \\
    LLaVA-NeXT-Vicuna  & 36.71 & 38.73 &  2.83 & 100.0 & 0.19 & 17.98 \\
    \bottomrule
  \end{tabular}
  \caption{\cattr{} $\EM$ in percent, alongside three format diagnostics. The \textit{choose A} metric tracks the frequency of multiple-choice answers equal to option A (against a uniform expectation near 20\%). Every model's MCQ score exceeds its open-recall score, often by tens of percentage points.}
  \label{tab:supp-c2em}
\end{table}

Parse rates remain uniformly at or above 98.9\%, ensuring no model is unfairly penalized for malformed formatting. However, two \textit{choose A} values stand out as strong indicators of position bias. Llama-3.2-Vision selects option A nearly two-thirds of the time, revealing a prior over answer position rather than content. LLaVA-NeXT-Mistral selects A approximately one-third of the time. Qwen2.5-VL-7B shows the highest letter--text mismatch rate in the field at 1.17\%; while small in absolute terms, it is the only model where the benchmark's letter-against-text arbitration policy produces a measurable effect.

\subsection{Per-category \cags{}}

\cref{tab:supp-c2cat} breaks down \cags{} by question category. The naming category shows the widest performance spread (ranging from $.198$ to $.731$), indicating that taxonomic nomenclature remains a primary differentiator among models. Conversely, the morphological attributes category is the most compressed ($.454$ to $.645$), suggesting that basic attribute extraction is a broadly shared competence. Phi-3.5-Vision's score of $.154$ on anatomical localization represents the lowest single category score in the evaluation.

\begin{table}[htbp]
  \centering\footnotesize
  \setlength{\tabcolsep}{3pt}
  \begin{tabular}{@{}lrrrr@{}}
    \toprule
    Model & Anat.\ loc. & Morph. & Salience & Naming \\
    \midrule
    Qwen3-VL-32B       & \bst{.4754} & \snd{.6441} & \snd{.5839} & \bst{.7310} \\
    Qwen3-VL-30B-A3B   & .4525 & .6313 & .5582 & .6584 \\
    InternVL3-38B      & .4132 & \bst{.6451} & .5621 & .6691 \\
    Gemma-4-12B        & .4306 & .6205 & \bst{.5915} & .3742 \\
    Qwen3-VL-8B        & .4411 & .6244 & .5309 & .6429 \\
    InternVL3.5-38B    & .3747 & .6379 & .5595 & .5363 \\
    Molmo2-8B          & \snd{.4741} & .6429 & .5027 & .3301 \\
    Qwen2.5-VL-7B      & .3556 & .5800 & .5409 & \snd{.7127} \\
    InternVL3.5-30B    & .3504 & .6093 & .5367 & .5391 \\
    InternVL3-8B       & .3462 & .5844 & .5081 & .5813 \\
    InternVL3.5-8B     & .3722 & .6028 & .4953 & .4572 \\
    Gemma-3-12B        & .4103 & .5506 & .4657 & .5113 \\
    Qwen2.5-VL-3B      & .3890 & .5005 & .4272 & .6615 \\
    Pixtral-12B        & .3199 & .4950 & .4471 & .4654 \\
    LLaVA-NeXT-Vicuna  & .3640 & \wst{.4535} & .4571 & \wst{.1977} \\
    Phi-3.5-Vision     & \wst{.1541} & .5335 & .4540 & \swst{.2175} \\
    Llama-3.2-Vision   & .2862 & \swst{.4831} & \wst{.3168} & .5203 \\
    LLaVA-NeXT-Mistral & \swst{.2587} & .4843 & \swst{.3722} & .2383 \\
    \bottomrule
  \end{tabular}
  \caption{\cags{} by category. Colors mark the \bst{best}, \snd{second best}, \swst{second worst} and \wst{worst} model in each column. The category leaders vary across domains: InternVL3-38B leads morphological attributes, Gemma-4-12B leads relative salience, and Qwen3-VL-32B leads anatomical localization and naming. Molmo2-8B is second on anatomical localization at $.4741$, which is $.0013$ behind a model four times its size. A single aggregated scalar masks these localized strengths.}
  \label{tab:supp-c2cat}
\end{table}

\subsection{Per-family \cags{}}

\cref{tab:supp-c2fam} details \cags{} across all models and families, while \cref{tab:supp-c2famem} provides the $\EM$ scores on the identical item partition.

\begin{table*}[h]
  \centering\footnotesize
  \setlength{\tabcolsep}{3.5pt}
  \begin{tabular}{@{}rlrrrrrrrrrr@{}}
    \toprule
    \# & Model & attr\_bare & attr\_group & cat\_mcq & marker & common & genus & kind\_of & sci & vis\_set & vis\_side \\
    \midrule
     1 & Qwen3-VL-32B       & \snd{.644} & \snd{.584} & \snd{.959} & \snd{.727} & .619 & \bst{.696} & \bst{.763} & \bst{.629} & \bst{.435} & \snd{.526} \\
     2 & Qwen3-VL-30B-A3B   & .631 & .558 & .951 & .660 & .608 & .423 & .733 & .594 & \snd{.434} & .445 \\
     3 & InternVL3-38B      & \bst{.645} & .562 & .948 & .603 & .495 & \snd{.636} & .665 & .614 & .373 & .484 \\
     4 & Gemma-4-12B        & .621 & \bst{.592} & .772 & .720 & .286 & .121 & .615 & .105 & .385 & .488 \\
     5 & Qwen3-VL-8B        & .624 & .531 & .958 & .608 & .543 & .486 & .691 & .553 & .426 & .435 \\
     6 & InternVL3.5-38B    & .638 & .560 & .920 & .656 & .389 & .417 & .518 & .454 & .310 & .498 \\
     7 & Molmo2-8B          & .643 & .503 & \wst{.532} & \bst{.743} & .284 & .147 & .579 & .129 & .366 & \bst{.746} \\
     8 & Qwen2.5-VL-7B      & .580 & .541 & \bst{.962} & .563 & \bst{.627} & \snd{.636} & \snd{.736} & \snd{.616} & .329 & .377 \\
     9 & InternVL3.5-30B    & .609 & .537 & .869 & .574 & .301 & .505 & .606 & .433 & .328 & .350 \\
    10 & InternVL3-8B       & .584 & .508 & .884 & .468 & .374 & .523 & .644 & .499 & .332 & .353 \\
    11 & InternVL3.5-8B     & .603 & .495 & .783 & .499 & .277 & .347 & .557 & .342 & .348 & .409 \\
    12 & Gemma-3-12B        & .551 & .466 & .915 & .572 & .438 & .299 & .688 & .243 & .416 & .339 \\
    13 & Qwen2.5-VL-3B      & .501 & .427 & .926 & \swst{.334} & .566 & .565 & .665 & .597 & .399 & .373 \\
    14 & Pixtral-12B        & .495 & .447 & .847 & .529 & .396 & .259 & .613 & .235 & .328 & \wst{.225} \\
    15 & LLaVA-NeXT-Vicuna  & \wst{.454} & .457 & \swst{.550} & \wst{.254} & \wst{.142} & \swst{.027} & \swst{.231} & \swst{.056} & .358 & .420 \\
    16 & Phi-3.5-Vision     & .534 & .454 & .716 & .337 & .177 & .066 & \wst{.096} & \wst{.049} & \wst{.108} & \swst{.245} \\
    17 & Llama-3.2-Vision   & \swst{.483} & \wst{.317} & .859 & .488 & \snd{.624} & .186 & .704 & .252 & .260 & .307 \\
    18 & LLaVA-NeXT-Mistral & .484 & \swst{.372} & .636 & .392 & \swst{.154} & \wst{.023} & .338 & .063 & \swst{.210} & .377 \\
    \midrule
       & \textbf{mean}      & .574 & .495 & .833 & .540 & .406 & .353 & .580 & .359 & .341 & .411 \\
    \bottomrule
  \end{tabular}
  \caption{Colors mark the \bst{best}, \snd{second best}, \swst{second worst} and \wst{worst} model in each column. \cags{} by question family for all 18 models. Hardest questions are concentrated in two areas: open-ended taxonomic naming and anatomical set/side reasoning. In contrast, coarse recognition via \textit{category\_mcq} is nearing saturation with a field mean of $.833$.}
  \label{tab:supp-c2fam}
\end{table*}
 
\begin{table*}[h]
  \centering\footnotesize
  \setlength{\tabcolsep}{3.5pt}
  \begin{tabular}{@{}rlrrrrrrrrrr@{}}
    \toprule
    \# & Model & attr\_bare & attr\_group & cat\_mcq & marker & common & genus & kind\_of & sci & vis\_set & vis\_side \\
    \midrule
     1 & Qwen3-VL-32B       & \snd{64.51} & \snd{59.01} & \snd{95.90} & \snd{61.15} & \bst{35.49} & \bst{69.46} & \bst{51.15} & \snd{51.80} & 34.12 & \snd{55.65} \\
     2 & Qwen3-VL-30B-A3B   & 63.16 & 55.53 & 95.11 & 51.92 & 33.57 & 40.96 & 44.17 & 47.61 & 32.07 & 46.34 \\
     3 & InternVL3-38B      & 64.44 & 56.54 & 94.81 & 47.25 & 18.94 & 63.22 & 34.14 & 50.38 & 23.02 & 49.94 \\
     4 & Gemma-4-12B        & 63.17 & \bst{60.40} & 77.20 & 59.93 & 12.73 & 12.53 & 23.78 &  4.73 & 19.30 & 52.84 \\
     5 & Qwen3-VL-8B        & 62.81 & 53.35 & 95.77 & 47.12 & 25.13 & 47.83 & 33.85 & 41.56 & 33.40 & 46.10 \\
     6 & InternVL3.5-38B    & 64.46 & 56.46 & 92.00 & 54.01 & 13.18 & 40.89 & 21.68 & 34.38 & 21.63 & 53.08 \\
     7 & Molmo2-8B          & \bst{64.56} & 51.03 & \wst{53.18} & \bst{65.41} &  7.86 & 14.97 & 15.61 &  5.50 & 34.37 & \bst{71.11} \\
     8 & Qwen2.5-VL-7B      & 58.24 & 54.86 & \bst{96.16} & 43.18 & 34.00 & \snd{63.28} & 46.75 & \bst{52.17} & 17.05 & 38.96 \\
     9 & InternVL3.5-30B    & 60.50 & 54.45 & 86.86 & 43.16 & 11.88 & 50.15 & 22.43 & 33.28 & 23.50 & 37.24 \\
    10 & InternVL3-8B       & 58.19 & 52.84 & 88.42 & 41.35 & 13.30 & 51.65 & 25.16 & 38.23 & 22.14 & 35.82 \\
    11 & InternVL3.5-8B     & 59.29 & 49.64 & 78.31 & 37.57 &  8.75 & 34.98 & 14.76 & 24.03 & \wst{13.85} & 43.99 \\
    12 & Gemma-3-12B        & 54.73 & 46.37 & 91.45 & 42.54 & 17.27 & 29.92 & 26.80 & 13.40 & \bst{41.48} & 29.41 \\
    13 & Qwen2.5-VL-3B      & 48.37 & 41.86 & 92.64 & 28.61 & 33.19 & 56.52 & 39.67 & 48.40 & \snd{41.22} & 37.29 \\
    14 & Pixtral-12B        & 49.20 & 46.01 & 84.68 & 37.22 &  8.92 & 25.32 & 36.10 & 15.23 & 17.93 & 32.60 \\
    15 & LLaVA-NeXT-Vicuna  & \wst{45.30} & 43.56 & \swst{55.02} & \wst{25.96} & \wst{1.14} & \swst{3.01} & \swst{4.72} & \swst{2.56} & \swst{15.46} & \wst{19.44} \\
    16 & Phi-3.5-Vision     & 53.54 & 46.11 & 71.60 & \swst{28.30} &  3.16 &  7.26 & \wst{1.94} &  2.80 & 18.04 & \swst{26.28} \\
    17 & Llama-3.2-Vision   & 46.75 & \wst{32.74} & 85.86 & 37.72 & \snd{34.54} & 18.87 & \snd{50.79} & 14.00 & 29.40 & 30.87 \\
    18 & LLaVA-NeXT-Mistral & \swst{46.20} & \swst{38.84} & 63.64 & 35.34 & \swst{1.53} & \wst{2.96} & 7.03 & \wst{2.08} & 24.88 & 39.08 \\
    \midrule
       & \textbf{mean}      & 57.08 & 49.98 & 83.26 & 43.76 & 17.48 & 35.21 & 27.81 & 26.79 & 25.71 & 41.45 \\
    \bottomrule
  \end{tabular}
  \caption{$\EM$ accuracy in percent by question family, reported on the identical item partition as \cref{tab:supp-c2fam}. Colors mark the \bst{best}, \snd{second best}, \swst{second worst} and \wst{worst} model in each column. Coarse recognition peaks at 96.16\%, whereas the highest open common-name score reaches only 35.49\%. Reading the \emph{mean} row against the one in \cref{tab:supp-c2fam} isolates where partial credit does work. The two agree to within half a point on \textit{attr\_bare}, \textit{attr\_group}, \textit{cat\_mcq}, \textit{genus} and \textit{vis\_side}, and they diverge most on \textit{kind\_of} and \textit{common}, where near misses are frequent and \cags{} credits them.}
  \label{tab:supp-c2famem}
\end{table*}

\subsection{Family difficulty}

\cref{tab:supp-c2famdiff} orders 10 question families by mean \cags{} across the field.

\begin{table}[h]
  \centering\footnotesize
  \setlength{\tabcolsep}{4pt}
  \begin{tabular}{@{}lrrrl@{}}
    \toprule
    Family & mean & min & max & reading \\
    \midrule
    \textit{visibility\_partset}   & \textbf{.341} & .108 & .435 & hardest \\
    \textit{open\_genus}           & .353 & .023 & .696 & widest spread \\
    \textit{open\_scientific\_name}& .359 & .049 & .629 & broadly hard \\
    \textit{open\_common\_name}    & .406 & .141 & .627 & free naming \\
    \textit{visibility\_side}      & .411 & .225 & .746 & laterality \\
    \textit{attr\_group\_value}    & .495 & .317 & .592 & moderate \\
    \textit{marker\_localization}  & .540 & .254 & .743 & marker binding \\
    \textit{attr\_bare\_value}     & .574 & .454 & .645 & most compressed \\
    \textit{open\_kind\_of}        & .580 & .096 & .763 & folk category \\
    \textit{category\_mcq}         & \textbf{.833} & .532 & .962 & near solved \\
    \bottomrule
  \end{tabular}
  \caption{Mean \cags{} per family across all 18 models. The most tightly compressed family is \textit{attr\_bare\_value} (minimum $.454$), underscoring that basic attribute extraction is a foundational competence for modern VLMs. Consequently, the harder families remain the primary discriminators of model capability.}
  \label{tab:supp-c2famdiff}
\end{table}

\subsection{The recognition--recall gap}

\cref{tab:supp-c2recog} contrasts the two naming formats for the models exhibiting the widest and narrowest performance gaps.
\begin{table}[h]
  \centering
  \footnotesize
  \setlength{\tabcolsep}{4pt}
  \begin{tabular}{@{}lrrr@{}}
    \toprule
    & \multicolumn{2}{c}{Taxonomic questions} & \\
    \cmidrule(lr){2-3}
    Model & MCQ & OE & gap \\
    \midrule
    InternVL3.5-30B  & .869 & .301 & \textbf{.568} \\
    Phi-3.5-Vision   & .716 & .177 & .539 \\
    InternVL3.5-38B  & .920 & .389 & .531 \\
    InternVL3-8B     & .884 & .374 & .510 \\
    InternVL3.5-8B   & .783 & .277 & .506 \\
    Gemma-4-12B      & .772 & .285 & .487 \\
    InternVL3-38B    & .948 & .495 & .453 \\
    Qwen3-VL-32B     & .959 & .619 & .340 \\
    Qwen2.5-VL-7B    & .962 & .627 & .335 \\
    Llama-3.2-Vision & .859 & \textbf{.624} & \textbf{.235} \\
    \bottomrule
  \end{tabular}
  \caption{Recognition--recall gap for taxonomic questions.}
  \label{tab:supp-c2recog}
\end{table}
InternVL3.5 models show the largest gaps (from $.51$ to $.57$). They recognize items well, but they struggle to generate names. On the other hand, the Qwen models show the smallest gaps. They generate names as well as they recognize them. Llama-3.2-Vision scores $.235$. It ranks 17th overall. However, its open common-name \cags{} of $.624$ is the second highest. The model names birds very well. Its bad multiple-choice score comes from position bias, not poor knowledge. We only see this by looking at \cags{} and format metrics together.

\subsection{Three anomalies the graded metric surfaces}

\paragraph{Position bias.}Llama-3.2-Vision picks option A $63.26\%$ of the time. This is about $3.2\times$ the normal rate. LLaVA-NeXT-Mistral picks option A $33.96\%$ of the time. Llama gets many correct answers just by guessing A. This shows we need a score that fixes position bias. We also must report how often models choose A. 

\paragraph{Confidently wrong.} Phi-3.5-Vision is the only model with a severity over 1 (at $1.018$). It is also the only model where \cags{} ($.405$) is lower than $\EM$ ($.415$). For all other models, \cags{} is equal or higher. This happens because we give partial credit for near misses. Phi is different. Its wrong answers are worse than random guesses. This is very clear on \textit{visibility\_partset}. Here, its \cags{} drops to .108 and severity rises to 1.088. A severity over 1 shows why we need \cags{}. $\EM$ cannot show this problem.

\paragraph{The grounding outlier.} Molmo2-8B gets the highest score on \textit{visibility\_side} ($.746$) and \textit{marker\_localization} ($.743$). It beats much larger 30B and 38B models here. But, it does poorly on naming ($.330$). It ranks 7th overall. This low rank hides a major strength. Molmo2-8B is the only model in our test that uses explicit pointing training.

\subsection{\cags{} re-ranks where it matters}

\cref{tab:supp-c2rerank} details every model whose ranking shifts between the two metrics.

\begin{table}[htbp]
  \centering\footnotesize
  \setlength{\tabcolsep}{3pt}
  \begin{tabular}{@{}lrrrlp{2.8cm}@{}}
    \toprule
    Model & \cags{} rank & $\EM$ rank & $\Delta$ & Why \\
    \midrule
    Gemma-4-12B        & 4  & 6  & $+2$ & near misses earn \\
    LLaVA-NeXT-Vicuna & 15 & 18 & $+3$ & misses land close \\
    Qwen2.5-VL-7B     & 8  & 9  & $+1$ & \\
    InternVL3.5-8B    & 11 & 12 & $+1$ & \\
    Phi-3.5-Vision    & 16 & 15 & $-1$ & confidently wrong \\
    LLaVA-NeXT-Mistral& 18 & 16 & $-2$ & misses land far \\
    \midrule
    \multicolumn{2}{@{}l}{11 models unchanged}
      & -- & 0 & two metrics concur \\
    \bottomrule
  \end{tabular}
  \caption{Instances where \cags{} and $\EM$ disagree. The signature shift is LLaVA-NeXT-Vicuna surpassing Phi-3.5-Vision. While Vicuna achieves fewer $\EM$s, its misses consistently land closer to the truth; thus, a closeness-aware, chance-corrected metric appropriately ranks it higher.}
  \label{tab:supp-c2rerank}
\end{table}

Because no swap shifts a model by more than three positions, and the top-tier rankings remain undisturbed, \cags{} refines the exact-match ranking rather than upending it. This property ensures the graded metric is both trustworthy and diagnostically rich, rather than merely divergent.

\subsection{Laterality}

We isolate the items where a model successfully identifies the correct anatomical part. Then we evaluate its ability to determine the correct laterality (left versus right). 17 of the 18 models perform at or below chance, with 9 falling significantly below. Mirror confusion ranges from 21.8\% to 54.0\% (against a 20\% uniform baseline), and this error rate correlates heavily with overall model strength ($\rho=.75$). This combination of factors rules out two potential explanations. First, the failure is not due to an inability to locate the part, as we condition on successful part identification. Second, the models are not simply guessing, as random guessing would drive the rate toward the uniform baseline, not significantly below it. Instead, stronger models systematically isolate the correct part and then actively choose the mirrored side. The notable exception is Molmo2-8B at $.689$, reflecting its unique training with explicit pointing supervision. Laterality exhibits the lowest loading coefficient on the shared cross-component factor ($.154$), consistent with the premise that it measures a capability largely disjoint from the rest of the benchmark tasks. 

We also acknowledge a corresponding limitation in the metric. The \cags{} anatomy kernel assigns a high similarity score ($.883$) between the left wing and right wing because they are spatially proximal in the structural layout. Consequently, the kernel over-credits the exact laterality confusion detailed in this analysis. For this reason, we compute the laterality figures above using $\EM$ rather than \cags{}. The human validation study (see \cref{sec:supp-validation}) confirms that annotators correctly reject these mirrored items, highlighting the limitation in the kernel's spatial credit assignment.

\begin{figure}[htbp]
  \centering
  \includegraphics[width=0.9\columnwidth]{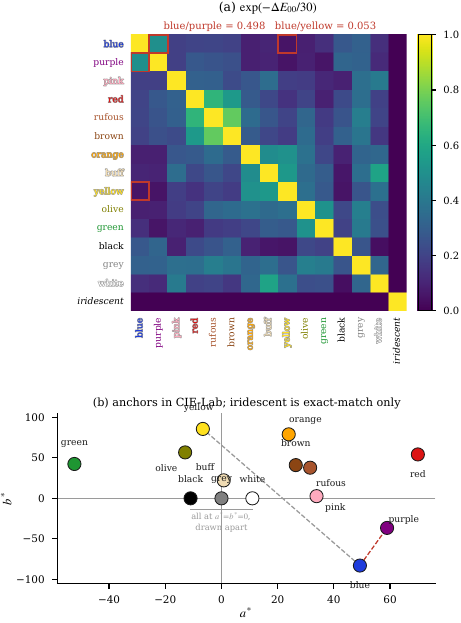}
  \caption{Color terms are compared perceptually rather than lexically; thus,
  predicting purple instead of blue receives significantly more partial credit
  than predicting yellow instead of blue.}
  \label{fig:supp-c2color}
\end{figure}

\begin{figure}[htbp]
  \centering
  \includegraphics[width=0.9\columnwidth]{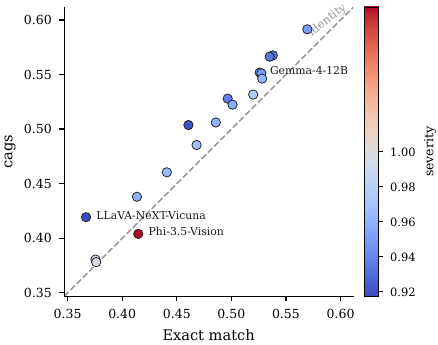}
  \caption{Partial credit elevates every model above the identity line, with one
  exception. Phi-3.5-Vision falls below the line because its prediction errors
  land further from the ground truth than a random distractor.}
  \label{fig:supp-c2scatter}
\end{figure}

\section{\cspat{} Full Results}
\label{sec:supp-c3}

Every task carries an image-blind floor, which we report in the same table as the score. Multiple tasks fall below their floor, which no accuracy-only report would reveal.

\subsection{Per-task summary}

\cref{tab:supp-c3summary} gives the best model and the floor for each of the eight tasks.

\begin{table}[h]
  \centering\footnotesize
  \setlength{\tabcolsep}{2.5pt}
  \begin{tabular}{llrl}
    \toprule
    Task & Best model & Score & Floor \\
    \midrule
    Counting          & InternVL3-38B & .799 & modal count .501 \\
    Viewpoint         & Qwen3-VL-32B  & .358 & majority .520, $\kappa$VAS 0 \\
    Absolute distance & Qwen3-VL-32B  & .858 & per-type constant .410 \\
    Relative distance & Gemma-4-12B   & .443 & chance .250 \\
    Relative counting & Qwen3-VL-32B  & .403 & chance .250 \\
    Localization      & Qwen3-VL-32B  & .866 & mean box .415 \\
    Relative position & Qwen3-VL-32B  & .550 & majority .125 \\
    OCR (control)     & Qwen3-VL-30B  & .987 & constant string .349 \\
    \bottomrule
  \end{tabular}
  \caption{Per-task headline. Counting is macro-\cas{}, viewpoint is $\kappa$VAS, absolute distance is within-plate Spearman $\rho$, relative distance and relative counting are balanced accuracy, localization is mean IoU, relative position is accuracy, and OCR is suffix character accuracy.}
  \label{tab:supp-c3summary}
\end{table}

\subsection{Counting}
 
Counting runs on all ten 500-image subsets. The floor is $0.501$ on the graded metric and $0.089$ on $\EM$, which makes the entire measurable band on the graded metric $0.30$ wide and places one of eighteen models within $0.013$ of its own floor. Models average $.695$ macro-\cas{} against that floor of $.501$. An average of $.695$ means the typical answer is off by about 30\% of the true count. InternVL3-38B leads at $.799$. $\EM$, however, does not exceed $.242$ for any model. So three quarters of the answers miss by at least one animal. $\EM$ falls between $.066$ and $.239$ against a constant-predictor floor of $.089$, so several models fall below the floor on $\EM$ while sitting well above it on the graded score. We keep both columns for that reason. No model counts 180 bees exactly, so on dense images, $\EM$ collapses toward the modal-answer rate and stops discriminating.
 
\paragraph{The two-panel penalty.}
We build counting items on three canvas tiers. The Easy tier shows one image filling the canvas. The Medium tier splits the canvas into two panels drawn from different datasets. The Hard tier splits it into two panels drawn from the same dataset. The encoder budget is fixed across tiers, so a target occupies about half the area in Medium and Hard that it occupies in Easy. We score each model on each tier and take the difference between consecutive tiers. \cref{tab:supp-c3panel} reports three numbers per contrast. The \emph{drop} is the mean of those 18 per-model differences, written so a positive value means the harder tier costs accuracy. $\sigma$ is the standard deviation of the same 18 differences, so it says how much the effect varies from model to model. The last column counts the models whose own drop is positive.
\begin{table}[h]
  \centering\footnotesize
  \setlength{\tabcolsep}{5pt}
  \begin{tabular}{lrrr}
    \toprule
    Contrast & Effect & $\sigma$ & Holds for \\
    \midrule
    easy to medium   & $+.064$ & $.037$ & 18 of 18 \\
    medium to hard   & $-.005$ & $.027$ & 6 of 18 \\
    \bottomrule
  \end{tabular}
  \caption{Cutting 50\% the area a target occupies costs every model accuracy. Drawing both panels from the same corpus makes no further difference.}
  \label{tab:supp-c3panel}
\end{table}
The first row holds for every model without exception, so the panel split is a real and universal effect. The second row holds for a third of models and its mean is within noise, so dataset homogeneity across panels is not what makes the task hard.

\paragraph{Rank correlation as a shortcut detector.}
We report the Spearman correlation between predicted and true counts per corpus. A model that answers a constant scores exactly zero, whatever its \cas{}. A model whose \cas{} is high and whose $\rho$ is near zero is not counting. It is answering near the modal count and collecting graded credit for proximity. This column is what separates counting from guessing, and neither \cas{} nor $\EM$ can do it alone.

\subsection{Viewpoint}
 
Viewpoint has ten orientations and a skewed prior, so the three floors differ by a factor of three. A constant \emph{right} predictor scores $\EM = 0.520$, $\mathrm{VAS} = 0.534$, $\mathrm{macroVAS} = 0.170$ and $\kappa\mathrm{VAS}=0.00$. We report $\kappa$VAS as the headline because it is the only column a constant answer cannot inflate. 17 of 18 models fall below the majority-class floor on $\EM$. The best $\kappa$VAS in the field is $.358$, shared by Qwen3-VL-32B and Molmo2-8B. \cref{tab:supp-c3view} gives the six models whose rank moves most between the two columns.
 
\begin{table}[htbp]
  \centering\footnotesize
  \setlength{\tabcolsep}{3pt}
  \begin{tabular}{lrrrrr}
    \toprule
    Model & $\EM$ & rank & $\kappa$VAS & rank & shift \\
    \midrule
    Molmo2-8B        & .228 & 10  & .358 & 1.5 & $+8.5$ \\
    Qwen2.5-VL-7B    & .050 & 15  & .182 & 7.5 & $+7.5$ \\
    Qwen3-VL-32B     & .417 & 7   & .358 & 1.5 & $+5.5$ \\
    InternVL3.5-38B  & .483 & 2   & .208 & 6   & $-4.0$ \\
    InternVL3.5-8B   & .456 & 3.5 & .117 & 10  & $-6.5$ \\
    Phi-3.5-Vision   & .456 & 3.5 & .083 & 11  & $-7.5$ \\
    \bottomrule
  \end{tabular}
  \caption{The six largest rank shifts between $\EM$ and $\kappa$VAS on viewpoint. Tied models share the average of the positions they span. InternVL3.5-8B and Phi-3.5-Vision both score $\EM = .456$ and fill positions 3 and 4, so both read 3.5. Molmo2-8B and Qwen3-VL-32B both score $\kappa\mathrm{VAS} = .358$ and fill positions 1 and 2, so both read 1.5. A shift is the difference between two midranks and is fractional for the same reason. Rising models place a large share of answers in the \emph{over} class, which means they name a compound direction where the annotation records a cardinal one. $\EM$ scores every one of those as zero.}
  \label{tab:supp-c3view}
\end{table}
 
Molmo2-8B and Qwen2.5-VL-7B each place 62.9\% of answers in the \emph{over} class. Phi-3.5-Vision places 0.7\%. So the reordering is not noise. It is an error-composition difference that $\EM$ cannot see.
 
\paragraph{Validating the rubric.}
The rubric grades a viewpoint answer by how many 45-degree steps separate it from the ground truth. Full credit at zero steps, partial credit at one, less at two, none at the opposite. We check these step weights two ways. First, we re-score the same predictions with a symmetric angular error on the ring. This construction shares none of the rubric's design decisions. It maps each viewpoint to a bearing and takes the absolute circular difference, so there is nothing to tune. \cref{tab:supp-c3ring} works one ground truth through both.
 
\begin{table}[htbp]
  \centering\footnotesize
  \setlength{\tabcolsep}{5pt}
  \begin{tabular}{@{}llrrr@{}}
    \toprule
    GT & Answer & steps & angle & $\EM$ \\
    \midrule
    left & left       & 0 & $0^{\circ}$   & 1 \\
    left & front-left & 1 & $45^{\circ}$  & 0 \\
    left & front      & 2 & $90^{\circ}$  & 0 \\
    left & right      & 4 & $180^{\circ}$ & 0 \\
    \bottomrule
  \end{tabular}
  \caption{One ground truth (GT) viewpoint under both scorers. The rubric reads the steps column and the angular metric reads the angle column. $\EM$ reads neither and gives the last three rows the same zero. Labels that sit off the ring carry no bearing, so angular check covers the ring classes only, which is why we use it as a check and not as the headline.}
  \label{tab:supp-c3ring}
\end{table}
 
The \emph{over} class is the second row of that table. A model answers \emph{front-left} where the annotation records \emph{left}. It has named the
correct side and added a component the annotation does not carry. $\EM$ scores that zero, the same as \emph{right}. The rubric and the angular metric both place it one step away. That single row is what separates Molmo2-8B, which lands there on $62.9\%$ of its answers, from Phi-3.5-Vision, which lands there on $0.7\%$, and it is why the two models move eight and a half ranks apart between the columns of \cref{tab:supp-c3view}. The two constructions agree on the model ordering. Second, we sweep the rubric weights and re-rank under each setting. The ordering is nearly invariant to them, so the viewpoint result does not rest on the particular credit we assign to one step.

\subsection{Absolute distance}

This is the sharpest \emph{below-floor} result in the benchmark. Answering one fixed size per measurement type would score $\accat{20} = 0.410$. 15 of 18 models fall below that. Only Qwen3-VL-32B at $.622$, Qwen3-VL-30B-A3B at $.528$ and InternVL3.5-38B at $.503$ beat it. The predictions are well-formed numbers in the right units with far too little spread across specimens. Mean relative error across the field is $0.374$.

\paragraph{The compression index.}
We define $\kappa$ as the ratio of predicted spread to gold spread within a measurement type. A model answering from one remembered size has $\kappa \approx 1$. Three models sit below $\kappa = 0.20$, which are Qwen3-VL-32B at $0.085$, Qwen3-VL-30B-A3B at $0.168$ and InternVL3.5-38B at $0.189$. These are the same three models that beat the floor.

\paragraph{Ranking and calibration dissociate.}
We compute $\accat{20}$ a second time after removing each model's per-type bias. \cref{tab:supp-c3dist} separates \emph{models that measure} from \emph{models that calibrate}.

\begin{table}[h]
  \centering\footnotesize
  \setlength{\tabcolsep}{3pt}
  \begin{tabular}{lrrrl}
    \toprule
    Model & $\rho_{\text{within}}$ & $\accat{20}$ & calib & diagnosis \\
    \midrule
    InternVL3.5-8B    & .703 & .263 & .509 & measures, no calibration \\
    Gemma-4-12B       & .753 & .385 & .512 & measures, no calibration \\
    Qwen3-VL-8B       & .682 & .303 & .453 & measures, no calibration \\
    Qwen3-VL-32B      & .858 & .622 & .614 & measures and calibrates \\
    Qwen2.5-VL-3B     & .081 & .194 & .306 & neither \\
    LLaVA-NeXT-Vicuna & $-.023$ & .217 & .326 & neither \\
    \bottomrule
  \end{tabular}
  \caption{Measurement and calibration come apart. InternVL3.5-8B orders the
  specimens well at $\rho_{\text{within}} = .703$ but sits below the floor on
  precision at $.263$. Removing its per-type bias lifts it to $.509$, from below
  the floor to above it.}
  \label{tab:supp-c3dist}
\end{table}

This is the reason we report $\rho_{\text{within}}$ as the headline rather than $\accat{20}$. A model can read the image and still answer on the wrong scale. Accuracy alone calls that failure and ranking alone calls it success. 

\subsection{Localization}

\cref{tab:supp-c3loc} gives the full leaderboard with the mean gold box inserted at its rank.
\begin{table}[h]
  \centering\footnotesize
  \setlength{\tabcolsep}{2.2pt}
  \begin{tabular}{rlrrrrrl}
    \toprule
    \# & Model & IoU$_{\text{all}}$ & a.25 & a.50 & a.75 & parse & convention \\
    \midrule
     1 & Qwen3-VL-32B      & \textbf{.866} & .995 & .984 & .877 & 1.000 & n1000 xyxy \\
     2 & Qwen3-VL-8B       & .862 & .995 & .985 & .868 & 1.000 & n1000 xyxy \\
     3 & Qwen3-VL-30B-A3B  & .861 & .995 & .984 & .867 & 1.000 & n1000 xyxy \\
     4 & InternVL3-38B     & .744 & .941 & .832 & .675 & 0.999 & n1000 xyxy \\
     5 & LLaVA-NeXT-Vicuna & .701 & .977 & .885 & .444 & 1.000 & n01 xyxy \\
     6 & Gemma-4-12B       & .589 & .749 & .631 & .475 & 0.999 & n1000 xywh \\
    -- & \emph{mean gold box} & \emph{.415} & \emph{.784} & \emph{.369} & \emph{.025} & & \\
     7 & InternVL3.5-38B   & .412 & .605 & .329 & .102 & 0.997 & abs xywh \\
     8 & Gemma-3-12B       & .329 & .525 & .263 & .054 & 1.000 & abs xywh \\
     9 & Qwen2.5-VL-7B     & .308 & .529 & .210 & .003 & 0.998 & abs xywh \\
    10 & Qwen2.5-VL-3B     & .307 & .527 & .210 & .003 & 1.000 & abs xywh \\
    11 & InternVL3.5-30B   & .289 & .446 & .174 & .053 & 0.995 & abs xywh \\
    12 & InternVL3-8B      & .246 & .399 & .200 & .086 & 0.940 & abs xywh \\
    13 & InternVL3.5-8B    & .236 & .376 & .156 & .032 & 0.965 & abs xywh \\
    14 & Pixtral-12B       & .229 & .331 & .202 & .077 & 0.999 & abs xywh \\
    15 & LLaVA-NeXT-Mistral& .089 & .239 & .171 & .141 & \textbf{0.523} & abs xywh \\
    16 & Llama-3.2-Vision  & .061 & .017 & .001 & .001 & 0.996 & abs xywh \\
    17 & Molmo2-8B         & .055 & .070 & .027 & .018 & 0.998 & abs xywh \\
    18 & Phi-3.5-Vision    & .030 & .026 & .008 & .001 & 1.000 & abs xywh \\
    \bottomrule
  \end{tabular}
  \caption{Localization. The headline column charges unparsed replies as zero. Twelve of eighteen models fall below the mean gold box, which is a rule that draws the same box on every image. The convention column records which of eight box formats the model emitted, detected by scoring under all eight. Molmo2-8B and Phi-3.5-Vision mostly repeat the example box given in the prompt.}
  \label{tab:supp-c3loc}
\end{table}
The convention detection is crucial. InternVL3-38B scores $0.147$ under absolute $xywh$ and $0.829$ under normalized $xyxy$ in $[0,1000]$. Scoring it under the wrong convention would understate it by $0.68$ mean IoU and would read as a capability claim and not a parsing error. LLaVA-NeXT-Mistral parses only 52.3\% of the time. Its mean IoU over parsed replies is $.170$ and its headline is $.089$. We report both so the failure is attributable to output format as well.

\subsection{Relative counting}

\cref{tab:supp-c3relcnt} gives the full leaderboard.

\begin{table}[h]
  \centering\footnotesize
  \setlength{\tabcolsep}{3pt}
  \begin{tabular}{rlrrrr}
    \toprule
    \# & Model & lift & balAcc & acc & posBias \\
    \midrule
     1 & Qwen3-VL-32B      & $+.403$ & .653 & .672 & .445 \\
     2 & Gemma-4-12B       & $+.388$ & .638 & .658 & .479 \\
     3 & InternVL3-38B     & $+.366$ & .616 & .634 & \textbf{.513} \\
     4 & InternVL3.5-38B   & $+.338$ & .588 & .599 & .409 \\
     5 & Molmo2-8B         & $+.329$ & .579 & .594 & .235 \\
     6 & Qwen3-VL-8B       & $+.320$ & .570 & .587 & .239 \\
     7 & Qwen3-VL-30B-A3B  & $+.312$ & .562 & .576 & .394 \\
     8 & InternVL3.5-30B   & $+.305$ & .555 & .569 & .416 \\
     9 & InternVL3.5-8B    & $+.277$ & .527 & .540 & .275 \\
    10 & Qwen2.5-VL-7B     & $+.273$ & .523 & .536 & .283 \\
    11 & Pixtral-12B       & $+.233$ & .483 & .494 & .205 \\
    12 & Gemma-3-12B       & $+.228$ & .478 & .486 & .345 \\
    13 & Qwen2.5-VL-3B     & $+.184$ & .434 & .441 & .193 \\
    14 & Phi-3.5-Vision    & $+.140$ & .390 & .397 & .232 \\
    15 & InternVL3-8B      & $+.098$ & .348 & .357 & .289 \\
    16 & Llama-3.2-Vision  & $+.056$ & .306 & .311 & .262 \\
    17 & LLaVA-NeXT-Mistral& $+.005$ & .255 & .256 & .257 \\
    18 & LLaVA-NeXT-Vicuna & \textbf{$-.005$} & .245 & .245 & .249 \\
    \bottomrule
  \end{tabular}
  \caption{Relative counting. Lift is accuracy above the .250 chance floor,
  normalized. The bottom model sits marginally below chance and answers one
  option 100\% of the time.}
  \label{tab:supp-c3relcnt}
\end{table}

\subsection{Relative position}

\cref{tab:supp-c3relpos} gives the full leaderboard with the angular error beside the accuracy.

\begin{table}[h]
  \centering\footnotesize
  \setlength{\tabcolsep}{3pt}
  \begin{tabular}{rlrrrr}
    \toprule
    \# & Model & acc & lift & angErr & within45 \\
    \midrule
     1 & Qwen3-VL-32B      & \textbf{.550} & $+.425$ & \textbf{27.0} & \textbf{.895} \\
     2 & Gemma-4-12B       & .495 & $+.370$ & 35.4 & .815 \\
     3 & Molmo2-8B         & .457 & $+.332$ & 32.9 & .864 \\
     4 & Qwen3-VL-8B       & .421 & $+.296$ & 40.1 & .803 \\
     5 & Qwen3-VL-30B-A3B  & .355 & $+.230$ & 55.0 & .673 \\
     6 & Pixtral-12B       & .266 & $+.141$ & 64.4 & .576 \\
     7 & InternVL3.5-30B   & .252 & $+.127$ & 69.8 & .561 \\
     8 & InternVL3.5-8B    & .224 & $+.099$ & 71.5 & .541 \\
     9 & InternVL3.5-38B   & .219 & $+.094$ & 73.1 & .520 \\
    10 & Qwen2.5-VL-7B     & .213 & $+.088$ & 69.2 & .545 \\
    11 & InternVL3-38B     & .205 & $+.080$ & 80.5 & .471 \\
    12 & InternVL3-8B      & .203 & $+.078$ & 72.9 & .515 \\
    13 & Phi-3.5-Vision    & .160 & $+.035$ & 80.7 & .440 \\
    14 & Gemma-3-12B       & .145 & $+.020$ & 86.9 & .395 \\
    15 & LLaVA-NeXT-Mistral& .131 & $+.006$ & 89.7 & .395 \\
    16 & Llama-3.2-Vision  & .125 & $0.000$ & 89.7 & .382 \\
    17 & LLaVA-NeXT-Vicuna & .125 & $0.000$ & 90.0 & .375 \\
    \bottomrule
  \end{tabular}
  \caption{Relative position over eight directions on the beetle trays. The
  majority-direction floor is $.125$. Angular error is the mean error on the
  45-degree ring and \texttt{within45} is the share within one step. Two models
  sit exactly at the floor and answer one direction every time.}
  \label{tab:supp-c3relpos}
\end{table}

\subsection{The OCR negative control}

Locating a specimen label is a spatial act, but the answer depends on transcribing characters. There is no position, extent, direction or magnitude in the answer. We therefore run OCR on the same imagery, the same models and the same pipeline, and we hold it as a negative control of the \cspat{} composite. The control answers the obvious objection to any coherence result. If the seven spatial tasks agree merely because larger models are better at everything, then any eighth vision task on the same imagery should agree too. \cref{tab:supp-c3ocr} gives the four statistics with and without the control.

\begin{table}[h]
  \centering\footnotesize
  \setlength{\tabcolsep}{5pt}
  \begin{tabular}{lrr}
    \toprule
    Statistic & Seven spatial & $+$~OCR \\
    \midrule
    Mean inter-task $\rho$      & $+0.703$ & $+0.619$ \\
    Krippendorff $\alpha$       & $0.940$  & $0.921$ \\
    Variance on PC1             & $75\%$   & $68\%$ \\
    Mean $\rho$ against $\log$ params & $\mathbf{+0.588}$ & $\mathbf{+0.028}$ \\
    \bottomrule
  \end{tabular}
  \caption{OCR acts as the negative control. It correlates $+0.366$ with the seven spatial tasks, against $+0.703$ among themselves. Every spatial task individually beats its correlation with the control.}
  \label{tab:supp-c3ocr}
\end{table}

The last row is the argument. The seven spatial tasks track parameter count at $+0.588$ and the control tracks it at $+0.028$. The coherence is therefore not a general capability gradient. 

\paragraph{The constant-string exploit.}
OCR also surfaces a failure mode that only a floor makes visible. Specimen codes share a fixed prefix, so a model that answers the prefix and guesses the suffix scores $0.675$ on whole-code character accuracy while reading nothing. We therefore report suffix character accuracy against a constant-string floor of $0.349$. \cref{tab:supp-c3ocrinfl} gives the two columns for the four most inflated and the two least inflated models.

\begin{table}[h]
  \centering\footnotesize
  \setlength{\tabcolsep}{5pt}
  \begin{tabular}{lrrr}
    \toprule
    Model & charAcc & suffixCharAcc & inflation \\
    \midrule
    Phi-3.5-Vision    & .568 & .161 & \textbf{.407} \\
    Pixtral-12B       & .593 & .199 & .394 \\
    LLaVA-NeXT-Vicuna & .607 & .214 & .393 \\
    InternVL3-38B     & .620 & .246 & .374 \\
    Qwen2.5-VL-3B     & .969 & .940 & .029 \\
    Qwen3-VL-30B-A3B  & .993 & .987 & \textbf{.007} \\
    \bottomrule
  \end{tabular}
  \caption{Whole-code character accuracy against suffix character accuracy. The inflation column is what the \emph{common prefix} boosts a model. Phi-3.5-Vision reads $.161$ of the informative characters against a floor of $.349$.}
  \label{tab:supp-c3ocrinfl}
\end{table}

LLaVA-NeXT-Vicuna is a clear case. It gives a well-formatted answer 100\% of the time and scores $0.00$ $\EM$, $0.214$ suffix character accuracy, and between $0.990$--$1.000$ per-position accuracy on the first seven characters. The format compliance, per-position profile and floor together establish that nothing was read.

\subsection{Internal coherence}

\cref{tab:supp-c3coh} gives the four coherence statistics.

\begin{table}[h]
  \centering\footnotesize
  \setlength{\tabcolsep}{6pt}
  \begin{tabular}{lr}
    \toprule
    Statistic & Value \\
    \midrule
    Mean inter-task $\rho$ & $+0.703$ \\
    Range & $+0.351$ to $+0.909$ \\
    Krippendorff $\alpha$ & $0.940$ \\
    Variance on PC1 & $75\%$ \\
    \bottomrule
  \end{tabular}
  \caption{The seven spatial tasks cohere. The tightest pairs are viewpoint with relative counting at $+0.909$ and viewpoint with relative position at $+0.887$, which are both spatial-relation judgments.}
  \label{tab:supp-c3coh}
\end{table}

Within the cross-component analysis, \emph{relative position} at $.168$ and \emph{localization} at $.176$ load lowest on the shared factor, which is consistent with them measuring something the rest of the benchmark does not.

\begin{figure}[h]
  \centering
  \includegraphics[width=.9\columnwidth]{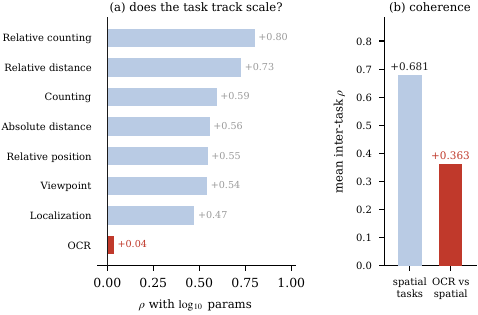}
  \caption{The seven spatial tasks track parameter count and the control does
  not, on the same imagery with the same models. The coherence is spatial rather
  than a general capability gradient.}
  \label{fig:supp-c3ocrfig}
\end{figure}

\begin{figure}[h]
  \centering
  \includegraphics[width=.9\columnwidth]{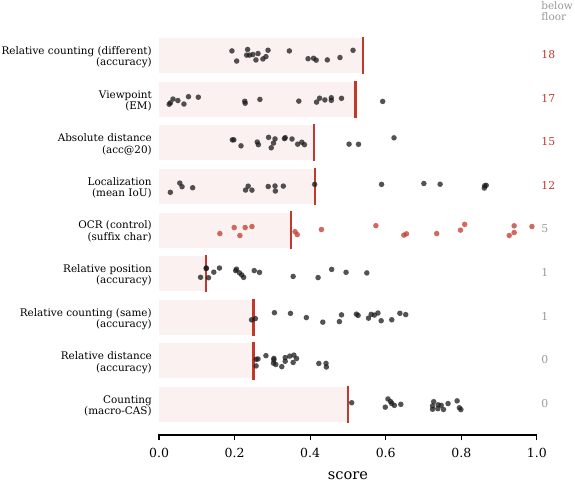}
  \caption{Several tasks sit below what a rule that never looks at the image achieves. A benchmark reporting only accuracy would record these as partial
  competence.}
  \label{fig:supp-c3floors}
\end{figure}

\section{\creid{} Full Results}
\label{sec:supp-c4}

The evaluation encompasses 18 models across 5 species and 3 conditions, resulting in 270 distinct subgroups and \num{270000} individual answers. We establish verdicts based on the random condition; since easy-condition pairs are separable by appearance alone, benchmarking against them would introduce circularity.

\subsection{Parsing integrity}

All \num{270000} answers resolve successfully. Of these, \num{269903} resolve at the strictest tier (a bare \emph{Yes} or \emph{No} after stripping markdown, quotes, and trailing punctuation). The remaining 97 answers resolve one tier down, containing exactly one of the two target words as an isolated token. No answers require the prefix, negated, or unparsed fallback tiers. We implement two critical parser safeguards. First, the string \emph{Yes} or \emph{No} occurs when a model echoes the prompt. Without an explicit rule, this would incorrectly resolve to \emph{Yes} via prefix matching. Second, the string \emph{I am not sure} contains the word \emph{not} and would resolve to \emph{No} under a contains-unique rule, mistakenly converting an abstention into a confident negative answer. To prevent this, we explicitly match refusal phrasings first and route them to the unparsed tier. Consequently, no result in this evaluation depends on parser leniency.

\subsection{Discriminability}

\cref{tab:supp-c4dprime} presents $\dpr$ for every model across all three conditions.
\begin{table}[t]
  \centering\footnotesize
  \setlength{\tabcolsep}{4pt}
  \begin{tabular}{lrrrl}
    \toprule
    Model & easy & random & hard & verdict \\
    \midrule
    Qwen3-VL-32B       & 1.455 & \textbf{0.696} & $-0.604$ & weak \\
    InternVL3-8B       & 2.062 & 0.684 & $-0.738$ & weak \\
    Qwen2.5-VL-3B      & 2.290 & 0.655 & $-0.764$ & weak \\
    InternVL3.5-30B    & 1.563 & 0.568 & $-0.529$ & weak \\
    Phi-3.5-Vision     & 1.919 & 0.562 & $-0.932$ & weak \\
    Qwen3-VL-30B-A3B   & 1.281 & 0.518 & $-0.248$ & weak \\
    Qwen3-VL-8B        & 1.635 & 0.492 & $-0.589$ & weak \\
    Molmo2-8B          & 1.820 & 0.491 & $-0.411$ & weak \\
    LLaVA-NeXT-Mistral & 1.738 & 0.476 & $-0.550$ & weak \\
    InternVL3-38B      & 1.597 & 0.463 & $-0.725$ & weak \\
    Qwen2.5-VL-7B      & 1.335 & 0.442 & $-0.395$ & weak \\
    Gemma-4-12B        & 1.304 & 0.351 & $-0.414$ & weak \\
    InternVL3.5-38B    & 1.278 & 0.258 & $-0.460$ & weak \\
    InternVL3.5-8B     & 0.905 & 0.236 & $-0.286$ & weak \\
    Gemma-3-12B        & 0.946 & 0.120 & $-0.091$ & degenerate \\
    Pixtral-12B        & 0.651 & 0.119 & $-0.240$ & at chance \\
    Llama-3.2-Vision   & 0.864 & 0.000 & $-0.208$ & degenerate \\
    LLaVA-NeXT-Vicuna  & 0.000 & 0.000 & $\;\;0.000$ & degenerate \\
    \bottomrule
  \end{tabular}
  \caption{Discriminability $\dpr$ pooled across the five species. The highest random-condition $\dpr$ in the field is $0.696$, which is approximately 64\% balanced accuracy. No model reaches $\dpr = 1.0$. Seventeen of eighteen models exhibit negative $\dpr$ under adversarial pairing.}
  \label{tab:supp-c4dprime}
\end{table}
We highlight two principal findings.

\textbf{A performance ceiling.} Individual re-identification remains largely unsolved for this generation of vision-language models, regardless of dataset manipulation. The finest granularity of visual discrimination in \biovqa{} represents the domain where current models struggle most.

\textbf{Below-chance performance under adversarial pairing.} Seventeen of the eighteen models exhibit negative $\dpr$ in the hard condition, ranging from $-0.932$ to $-0.091$. The single exception produces a constant output and is immune to dataset manipulation.

Model scale does not reliably predict performance; for instance, Qwen2.5-VL-3B ranks third, outperforming the significantly larger InternVL3-38B.

\subsection{Response bias}

\cref{tab:supp-c4crit} presents the criterion $c$ for the seven most biased models.

\begin{table}[t]
  \centering\footnotesize
  \setlength{\tabcolsep}{6pt}
  \begin{tabular}{lrrr}
    \toprule
    Model & $c$ easy & $c$ random & $c$ hard \\
    \midrule
    LLaVA-NeXT-Vicuna  & $-3.09$ & $-3.09$ & $-3.09$ \\
    LLaVA-NeXT-Mistral & $-2.41$ & $-2.46$ & $-2.54$ \\
    Gemma-3-12B        & $-1.60$ & $-1.99$ & $-1.99$ \\
    Llama-3.2-Vision   & $-1.13$ & $-1.59$ & $-1.59$ \\
    Qwen2.5-VL-7B      & $-1.18$ & $-1.34$ & $-1.51$ \\
    Qwen2.5-VL-3B      & $-0.88$ & $-1.32$ & $-1.53$ \\
    Qwen3-VL-30B-A3B   & $+0.13$ & $+0.15$ & $-0.06$ \\
    \bottomrule
  \end{tabular}
  \caption{Criterion $c$, pooled. Negative values indicate a standing preference for \emph{Yes}. 17 of the 18 models lean toward \emph{Yes}, and this bias deepens from easy to hard in nearly every case. Six models exceed $|c| = 1.0$ in the neutral condition, for which we report $\dpr$ and balanced accuracy rather than raw accuracy.}
  \label{tab:supp-c4crit}
\end{table}

As image pairs become increasingly confusable, models consistently become more affirmative rather than exhibiting appropriate caution. This behavior contradicts expectations for a model actively tracking its own uncertainty. Because this trend holds across different model families and scales, it reflects a general vulnerability rather than an architecture-specific artifact.

\subsection{The difficulty manipulation}

To validate our difficulty manipulation, we quantify its impact empirically. \cref{tab:supp-c4manip} details the easy-minus-hard accuracy difference for every model, alongside 95\% confidence intervals.

\begin{table}[t]
  \centering\footnotesize
  \setlength{\tabcolsep}{4pt}
  \begin{tabular}{lrl}
    \toprule
    Model & easy $-$ hard & 95\% CI \\
    \midrule
    InternVL3-8B       & $+38.8$ & $[+37.0, +40.6]$ \\
    InternVL3.5-30B    & $+38.4$ & $[+36.6, +40.1]$ \\
    InternVL3-38B      & $+38.4$ & $[+36.6, +40.2]$ \\
    Phi-3.5-Vision     & $+37.0$ & $[+35.2, +38.8]$ \\
    Qwen3-VL-32B       & $+36.9$ & $[+35.1, +38.7]$ \\
    Molmo2-8B          & $+36.0$ & $[+34.2, +37.7]$ \\
    Qwen2.5-VL-3B      & $+35.3$ & $[+33.5, +37.1]$ \\
    LLaVA-NeXT-Mistral & $+5.2$  & $[+3.2, +7.1]$ \\
    LLaVA-NeXT-Vicuna  & $+0.0$  & $[-2.0, +2.0]$ \\
    \bottomrule
  \end{tabular}
  \caption{Accuracy difference in percentage points between the easy and hard conditions, pooled, with two-proportion difference intervals. Seventeen models confirm the effect, one remains flat, and none reverse.}
  \label{tab:supp-c4manip}
\end{table}

Analyzed by species, 77 out of 90 model-species combinations exhibit a weakly monotonic easy $\geq$ random $\geq$ hard ordering in balanced accuracy. Under a null hypothesis where the condition is irrelevant, a random ordering of three conditions would be monotonic with a probability of $1/6$. The exact binomial probability of observing 77 or more monotonic orderings out of 90 is $1.9 \times 10^{-46}$. The single flat case corresponds to the model generating a constant answer, which by construction cannot react to visual manipulations.

\subsection{By species}

\cref{tab:supp-c4species} provides the mean $\dpr$ separated by species.
\begin{table}[t]
  \centering\footnotesize
  \setlength{\tabcolsep}{5pt}
  \begin{tabular}{lrrrr}
    \toprule
    Species & easy & random & hard & easy $-$ hard \\
    \midrule
    Whale shark    & 2.812 & 0.276 & $-0.906$ & 3.718 \\
    Spotted hyena  & 1.743 & 0.672 & $-0.126$ & 1.869 \\
    Grevy's zebra  & 1.346 & 0.391 & $-0.596$ & 1.942 \\
    Leopard        & 1.280 & 0.517 & $-0.188$ & 1.468 \\
    Beluga whale   & 0.508 & 0.383 & $-0.758$ & 1.267 \\
    \bottomrule
  \end{tabular}
  \caption{Mean $\dpr$ across models, partitioned by species. Beluga is diagnostically
  useful, as its easy-condition separation is notably the weakest.}
  \label{tab:supp-c4species}
\end{table}
We anticipated the anomalous beluga result based on dataset statistics prior to model evaluation. Beluga is the only species for which the random condition yields a negative similarity margin ($-0.0613$). Consequently, under uniform sampling, different-individual pairs are on average more visually similar than same-individual pairs. Because the DINOv2 embedding geometry for belugas is poorly organized around individual identity, manipulations defined within this space are inherently less effective.

\subsection{The specialist baseline}

\cref{tab:supp-c4mega} demonstrates the performance of the specialist model across all three conditions.

\begin{table}[t]
  \centering\footnotesize
  \setlength{\tabcolsep}{6pt}
  \begin{tabular}{lrrr}
    \toprule
    Condition & AUC & $\dpr$ & accuracy \\
    \midrule
    easy   & 0.927 & 2.055 & 0.902 \\
    random & 0.698 & \textbf{0.733} & 0.683 \\
    hard   & 0.247 & \textbf{$-0.969$} & 0.317 \\
    \bottomrule
  \end{tabular}
  \caption{MegaDescriptor evaluation on Grevy's zebra. The threshold is fit on the random condition and applied unchanged to both easy and hard conditions, making the easy and hard rows out-of-fit evaluations.}
  \label{tab:supp-c4mega}
\end{table}

The specialist model performs well on the easy and random conditions, but falls below chance on the hard condition. It demonstrates an effect magnitude that exceeds generalist VLMs. Because this system relies strictly on thresholded cosine similarity and lacks any language model or reasoning component, its performance drop is driven exclusively by the appearance anti-correlation induced by the adversarial condition. This provides compelling evidence that the adversarial condition successfully isolates and targets appearance-based matching rather than inadvertently penalizing linguistic reasoning. We do not claim a fundamental capability gap between VLMs and the specialist. On neutral pairs, the specialist achieves a $\dpr$ of $0.733$ compared to the best VLM's $0.696$, effectively a statistical tie. These pairs represent a formidable challenge for appearance-based methods in general.

\subsection{Construction check}

Using the same evaluation script, we score the raw DINOv2 similarities stored in the pair metadata. While these scores do not constitute a baseline, we report them to transparently quantify the degree of construction circularity. \cref{tab:supp-c4construct} shows the AUC of the embedding itself in each condition.

\begin{table}[t]
  \centering\footnotesize
  \setlength{\tabcolsep}{6pt}
  \begin{tabular}{lrrr}
    \toprule
    Species & easy AUC & random AUC & hard AUC \\
    \midrule
    Leopard       & 1.000 & 0.835 & 0.114 \\
    Whale shark   & 1.000 & 0.718 & 0.053 \\
    Spotted hyena & 1.000 & 0.784 & 0.095 \\
    Grevy's zebra & 1.000 & 0.658 & 0.068 \\
    Beluga whale  & 0.990 & 0.386 & 0.010 \\
    \bottomrule
  \end{tabular}
  \caption{The DINOv2 similarity itself, scored on the pairs it defines. An easy-condition AUC of $1.000$ confirms the construction operates exactly as intended. We report this to illustrate the empirical upper bound of the circularity concern.}
  \label{tab:supp-c4construct}
\end{table}

\subsection{Why below chance}

A natural objection is that $0.500$ should serve as a performance floor for a binary task, suggesting that below-chance accuracy implies the model \emph{knows} the correct answer but deliberately reports the opposite. We address this directly. An accuracy of $0.500$ is the expected value for a responder whose output is entirely independent of the stimulus (\ie random guessing). For any system whose output is conditioned on the input image, performance can be driven below $0.500$ through adversarial sample selection, which requires no latent knowledge of the true label.

\paragraph{The mechanism.}
Models implicitly apply a decision rule that approximates \emph{answer yes if the two animals look alike}. Because this rule generally correlates with ground truth, these models score above chance in the random condition. However, the hard condition explicitly selects pairs where visual similarity is \emph{anti-correlated with true identity}. The models apply their appearance-based rule unchanged, but we have adversarially arranged the sample such that this otherwise sound rule consistently points to the incorrect answer.

\paragraph{What negative $\dpr$ does and does not establish.}
A negative $\dpr$ establishes that the model's decision variable is anti-correlated with the ground truth label within this specific sample. It does \emph{not} establish that the model's internal decision variable represents actual identity. We posit that this variable is simply appearance similarity, a proxy that normally correlates with identity, but is intentionally decorrelated from it in this condition. Under this interpretation, the model never successfully forms an abstract identity judgment, meaning \emph{there is no true identity judgment being inverted}. Two evidences support this mechanism. First, a specialist system incapable of deliberation (\emph{MegaDescriptor}) exhibits the same vulnerability, reaching $\dpr = -0.969$ in the hard condition. A thresholded dot product cannot implicitly know an answer and willfully choose the opposite. Second, the magnitude of the model's sensitivity is largely preserved while only the sign flips. \cref{tab:supp-c4sign} pairs the random and hard columns for four models to illustrate this.
\begin{table}[t]
  \centering\footnotesize
  \setlength{\tabcolsep}{6pt}
  \begin{tabular}{lrrl}
    \toprule
    Model & $\dpr$ random & $\dpr$ hard & $|\dpr|$ \\
    \midrule
    InternVL3-8B  & 0.684 & $-0.738$ & 0.68 to 0.74 \\
    Qwen3-VL-32B  & 0.696 & $-0.604$ & 0.70 to 0.60 \\
    Qwen2.5-VL-3B & 0.655 & $-0.764$ & 0.66 to 0.76 \\
    Molmo2-8B     & 0.491 & $-0.411$ & 0.49 to 0.41 \\
    \bottomrule
  \end{tabular}
  \caption{The magnitude of the sensitivity survives the manipulation; only the sign changes. This rules out the hypothesis that the model lacks signal and is merely guessing, which would drive $\dpr$ toward 0 rather than to $-0.7$.}
  \label{tab:supp-c4sign}
\end{table}
This behavior is the hallmark signature of a constant heuristic mechanism being evaluated against inverted labels.

\subsection{Why retrieval is viewpoint-constrained}

We constrain our retrieval process to within specific viewpoint groups rather than searching across the entire catalog. Because viewpoint dominates the DINOv2 embedding space, not having this constraint would mean an anchor's most dissimilar neighbors would almost universally feature the opposite viewing angle. Consequently, hard same-individual pairs would merely consist of a left-flank photograph paired with a right-flank photograph, degenerating the evaluation from true identity discrimination into trivial viewpoint matching. Constraining retrieval strictly to a viewpoint group forces the embedding to discriminate based on features other than pose, ensuring the hard condition effectively evaluates fine-grained identity matching as intended.

\begin{figure}[t]
  \centering
  \includegraphics[width=0.9\columnwidth]{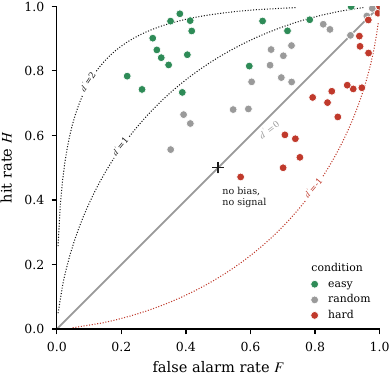}
  \caption{Difficulty is set by DINOv2 embedding similarity. In the hard condition appearance and identity disagree, so a model reading resemblance
  rather than identity scores below chance.}   \label{fig:supp-c4cond}
\end{figure}

\begin{figure}[t]
  \centering
  \includegraphics[width=.9\columnwidth]{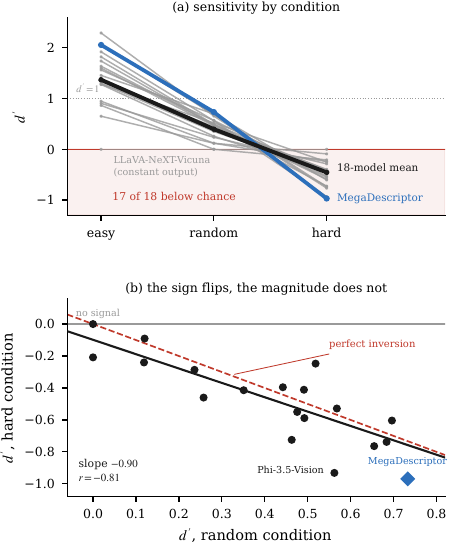}
  \caption{Seventeen of eighteen models and the specialist all fall below chance
  under adversarial pairing, while preserving the magnitude of their sensitivity.}
  \label{fig:supp-c4slope}
\end{figure}

\section{Metric Validation}
\label{sec:supp-validation}

We check \cags{} two ways, and we audit the other three components.

\subsection{The \cags{} sensitivity sweep}

We re-score every model under a grid of configurations on a fixed, family-stratified \num{60000}-row subsample per model, with identical rows across configurations. Let $r^{(0)}_i$ and $r^{(c)}_i$ be model $i$'s rank under the default and under configuration $c$. Then
\begin{equation}
  \rho = 1 - \frac{6\sum_i d_i^2}{n(n^2-1)},
  \qquad d_i = r^{(0)}_i - r^{(c)}_i
  \label{eq:supp-spearman}
\end{equation}
and displacement is $\max_i |d_i|$. \cref{tab:supp-sweep} gives $\rho$, Kendall $\tau$ and displacement for every configuration.

\begin{table}[t]
  \centering\footnotesize
  \setlength{\tabcolsep}{3pt}
  \begin{tabular}{lrrrcr}
    \toprule
    Config & $\rho$ & $\tau$ & disp. & top-3 & mean $|\Delta|$ \\
    \midrule
    default        & 1.0000 & 1.0000 & 0 & \cmark & 0.0000 \\
    $\tau = 20$    & 0.9979 & 0.9869 & 1 & \cmark & 0.0008 \\
    $\tau = 40$    & 0.9979 & 0.9869 & 1 & \cmark & 0.0008 \\
    $\tau = 50$    & 0.9979 & 0.9869 & 1 & \cmark & 0.0015 \\
    $\alpha = 0.3$ & 1.0000 & 1.0000 & 0 & \cmark & 0.0011 \\
    $\alpha = 0.4$ & 1.0000 & 1.0000 & 0 & \cmark & 0.0006 \\
    $\alpha = 0.6$ & 1.0000 & 1.0000 & 0 & \cmark & 0.0007 \\
    $\alpha = 0.7$ & 0.9979 & 0.9869 & 1 & \cmark & 0.0015 \\
    linear kernel  & 0.9959 & 0.9739 & 1 & \cmark & 0.0027 \\
    \midrule
    prefer text    & 0.9979 & 0.9869 & 1 & \cmark & 0.0004 \\
    anat in vis    & \textbf{0.9587} & \textbf{0.8562} & \textbf{3} & \cmark & \textbf{0.0200} \\
    \bottomrule
  \end{tabular}
  \caption{The \cags{} sensitivity sweep across eleven configurations. The base grid varies the color-kernel temperature $\tau$, the region-value weight $\alpha$, and the color kernel shape. The policy grid adds the letter against text resolution preference and one substantive modeling toggle.}
  \label{tab:supp-sweep}
\end{table}

Across the base grid, the minimum $\rho$ is $0.9959$, the minimum $\tau$ is $0.9739$, the maximum displacement is one rank, and the top three set is stable in every configuration. The only rank exchanges are single-position swaps between models whose bootstrap intervals already overlap, which are statistical ties rather than instability. The one configuration that moves mid-table ranks is folding anatomy part similarity into the visibility set kernel, so that naming the left wing when the right wing was visible earns partial credit. This is a substantive modeling
decision and not an arbitrary constant, and it changes the visibility-family scores. \cref{tab:supp-sweepmove} names the models that move under it.
\begin{table}[t]
  \centering\footnotesize
  \setlength{\tabcolsep}{5pt}
  \begin{tabular}{lrrr}
    \toprule
    Model & default & anat in vis & $\Delta$ \\
    \midrule
    Gemma-3-12B        & 12 & 9  & $+3$ \\
    Llama-3.2-Vision   & 17 & 15 & $+2$ \\
    Qwen2.5-VL-7B      & 8  & 10 & $-2$ \\
    InternVL3.5-8B     & 11 & 13 & $-2$ \\
    LLaVA-NeXT-Vicuna  & 15 & 18 & $-3$ \\
    Top three          & 1, 2, 3 & 1, 2, 3 & $0$ \\
    \bottomrule
  \end{tabular}
  \caption{Models movement under one substantive toggle. Top three set is unchanged and the movement is confined to families where crediting near-miss parts is a defensible design choice.}
  \label{tab:supp-sweepmove}
\end{table}
We adopt the $\EM$ default and we disclose that the alternative reshuffles the middle of the table by up to three positions while leaving the podium intact. It shows that the ranking's one sensitivity is a known semantic knob and not a fragile numeric one.

\subsection{The human study}

\cags{} could in principle produce the model ordering by itself, so we compare it against human judgment on a blind stratified sample.

\paragraph{Sample design.}
We draw \num{306} items across a category by \cags{}-band grid. The three bands are low with \cags{} in $[0, 0.2)$, mid with \cags{} in $[0.4, 0.6)$, and high with \cags{} in $[0.8, 1.0]$. Each band contributes 34 items per category and 102 items overall. The five families represented are \textit{attr\_bare\_value} with 102 items, \textit{attr\_group\_value} with 102, \textit{visibility\_side} with 41, \textit{visibility\_partset} with 40 and \textit{marker\_localization} with 21.

\paragraph{Protocol.}
Each item renders as a card showing the image, the question, the reference answer and the model answer. Model identity, the \cags{} value and the band are hidden. Three annotators rate each card on a three-level ordinal scale. A rating of 2 is correct or equivalent to the reference. A rating of 1 is incorrect but close, which covers the right body region with the wrong property, an adjacent color, an off-by-one size, one of two correct parts, or a near-miss location. A rating of 0 is unrelated or far off. An empty cell is an abstention. We treat it as a missing value, not a zero. We aggregate the ratings into $h_i \in [0,1]$, which is the weight-renormalized mean over the annotators who rated item $i$, rescaled to the unit interval. \cref{tab:supp-humanrel} gives the reliability statistics.

\begin{table}[t]
  \centering\footnotesize
  \setlength{\tabcolsep}{4pt}
  \begin{tabular}{lrr}
    \toprule
    Statistic & Value & $n$ \\
    \midrule
    Krippendorff $\alpha$, ordinal & \textbf{0.7937} & 904 \\
    Mean pairwise quadratic $\kappa$ & 0.7914 & \\
    $\kappa$, annotator 1 against 2 & 0.7579 & 292 \\
    $\kappa$, annotator 1 against 3 & 0.8312 & 306 \\
    $\kappa$, annotator 2 against 3 & 0.7851 & 292 \\
    \midrule
    Abstention, any annotator & 0.0458 & 306 \\
    Abstention, all annotators & 0.0000 & 306 \\
    \bottomrule
  \end{tabular}
  \caption{Inter-annotator reliability. Every item receives at least one rating.}
  \label{tab:supp-humanrel}
\end{table}

\cref{tab:supp-humanagree} gives agreement between the metric and the annotators.
\begin{table}[t]
  \centering
  \footnotesize
  \setlength{\tabcolsep}{3pt}
  \begin{tabularx}{\columnwidth}{@{}Xrr@{}}
    \toprule
    Statistic & All, $n=306$ & Complete, $n=292$ \\
    \midrule
    $\tau_b$, \cags{} against human & \textbf{0.5883} & 0.5879 \\
    \quad 95\% interval & $[0.515, 0.657]$ & $[0.514, 0.659]$ \\
    $\tau_b$, $\EM$ against human & 0.7178 & 0.7245 \\
    Difference, \cags{} minus $\EM$ & $-0.1295$ & $-0.1366$ \\
    $\rho$, \cags{} against human & 0.7328 & 0.7330 \\
    $\rho$, $\EM$ against human & 0.8044 & 0.8102 \\
    \midrule
    Quadratic $\kappa$, metric against human & 0.7187 & \\
    Mean pairwise annotator $\kappa$ & 0.7914 & \\
    \textbf{Ceiling ratio} & \textbf{0.908} & \\
    \bottomrule
  \end{tabularx}
  \caption{Agreement between the metric and the annotators. The ceiling ratio is
  the metric against human $\kappa$ divided by the mean pairwise annotator
  $\kappa$. \cags{} reaches $0.908$ of the rate at which the annotators agree
  with each other.}
  \label{tab:supp-humanagree}
\end{table}

\paragraph{Where \cags{} correlates with human.}
$\EM$ agrees with the annotators more closely than \cags{} does on the full sample, which is expected. The band construction makes $\EM$ a deterministic function of the band, since every low and mid item has $\EM = 0$ and every high item has $\EM = 1$. So the interesting question is what happens inside the $\EM$-zero subset, where $\EM$ carries no information at all.

\cref{tab:supp-humanauc} gives the separation \cags{} achieves there.

\begin{table}[t]
  \centering\footnotesize
  \setlength{\tabcolsep}{4pt}
  \begin{tabular}{lrl}
    \toprule
    Statistic & Value & Counts \\
    \midrule
    Items with $\EM = 0$ & 204 & 102 low, 102 mid \\
    AUC, credited against rejected & \textbf{0.6755} & $[0.591, 0.757]$ \\
    AUC, partial against rejected & 0.6823 & strict variant \\
    \bottomrule
  \end{tabular}
  \caption{Within the 204 items $\EM$ scores zero, \cags{} separates the
  answers annotators credited from the ones they rejected. $\EM$ cannot,
  because it assigns all 204 the same score.}
  \label{tab:supp-humanauc}
\end{table}

\cref{tab:supp-humanband} gives the mean metric and human score in each band.

\begin{table}[t]
  \centering\footnotesize
  \setlength{\tabcolsep}{5pt}
  \begin{tabular}{lrrr}
    \toprule
    Band & $n$ & mean \cags{} & mean $h$ \\
    \midrule
    low  & 102 & 0.0357 & 0.2484 \\
    mid  & 102 & 0.4647 & 0.3595 \\
    high & 102 & 1.0000 & 0.9861 \\
    \bottomrule
  \end{tabular}
  \caption{The human label tracks the band monotonically. The mid band is where
  the two diverge most, which is where partial credit is being tested.}
  \label{tab:supp-humanband}
\end{table}

\paragraph{Where \cags{} disagrees with humans.}

We report disagreements in \cref{tab:supp-humanfam}, family by family.

\begin{table}[t]
  \centering\footnotesize
  \setlength{\tabcolsep}{3pt}
  \begin{tabular}{lrrrr}
    \toprule
    Family & $n$ & $\tau_b$ \cags{} & mean \cags{} & mean $h$ \\
    \midrule
    \texttt{attr\_bare\_value}    & 102 & 0.6692 & 0.4999 & 0.5596 \\
    \texttt{attr\_group\_value}   & 102 & 0.5544 & 0.5130 & 0.5441 \\
    \texttt{visibility\_side}     & 41  & 0.5638 & 0.5415 & 0.6301 \\
    \texttt{visibility\_partset}  & 40  & \textbf{0.8063} & 0.3611 & 0.4542 \\
    \texttt{marker\_localization} & 21  & \textbf{0.4431} & \textbf{0.6231} & \textbf{0.2857} \\
    \bottomrule
  \end{tabular}
  \caption{Agreement by family. \texttt{marker\_localization} is the weakest
  agreement and the largest gap between metric and human, with mean \cags{} at
  $0.6231$ against mean human at $0.2857$.}
  \label{tab:supp-humanfam}
\end{table}

The mechanism is the laterality confusion we describe in \cref{sec:supp-c2}. The three worst-disagreeing items are all left wing answered as right wing or the reverse, where \cags{} scores between $0.559$ and $0.588$ and all three annotators score zero.

\cref{tab:supp-humandis} gives the three items.

\begin{table}[t]
  \centering\footnotesize
  \setlength{\tabcolsep}{4pt}
  \begin{tabular}{llrr}
    \toprule
    Reference & Response & \cags{} & $h$ \\
    \midrule
    (C) left wing & (D) right wing & 0.5884 & 0.0 \\
    (A) left wing & (C) right wing & 0.5693 & 0.0 \\
    (D) left wing & (C) right wing & 0.5594 & 0.0 \\
    \bottomrule
  \end{tabular}
  \caption{The three items where \cags{} and the annotators disagree most. The anatomy kernel places the left and right members of a paired part close together, so it over-credits exactly the confusion that matters. We state this as a limitation in the main paper and we compute the laterality analysis on $\EM$ for that reason.}
  \label{tab:supp-humandis}
\end{table}

The disagreement runs the other way as well. On four items the annotators credit a response that \cags{} scores at exactly zero, all of them cases where the reference is \emph{None} and the model names a part. Annotators judged the part visible in the image and the reference wrong. These are annotation errors in the source, not metric errors.

\paragraph{Leave-one-annotator-out.}

We check that no single annotator drives the result. \cref{tab:supp-humanloo} repeats the three headline statistics with each annotator dropped in turn.

\begin{table}[t]
  \centering\footnotesize
  \setlength{\tabcolsep}{5pt}
  \begin{tabular}{lrrr}
    \toprule
    Dropped & $\tau_b$ \cags{} & AUC & $\kappa$ metric-human \\
    \midrule
    none        & 0.5883 & 0.6755 & 0.7187 \\
    annotator 1 & 0.5306 & 0.6071 & 0.6464 \\
    annotator 2 & 0.6294 & 0.7063 & 0.7189 \\
    annotator 3 & 0.6039 & 0.6805 & 0.6782 \\
    \bottomrule
  \end{tabular}
  \caption{Leave-one-annotator-out. The conclusion holds in every subset. The
  unique-dissent rates are $0.103$, $0.182$ and $0.055$ respectively.}
  \label{tab:supp-humanloo}
\end{table}

\begin{figure}[t]
  \centering
  \includegraphics[width=0.9\columnwidth]{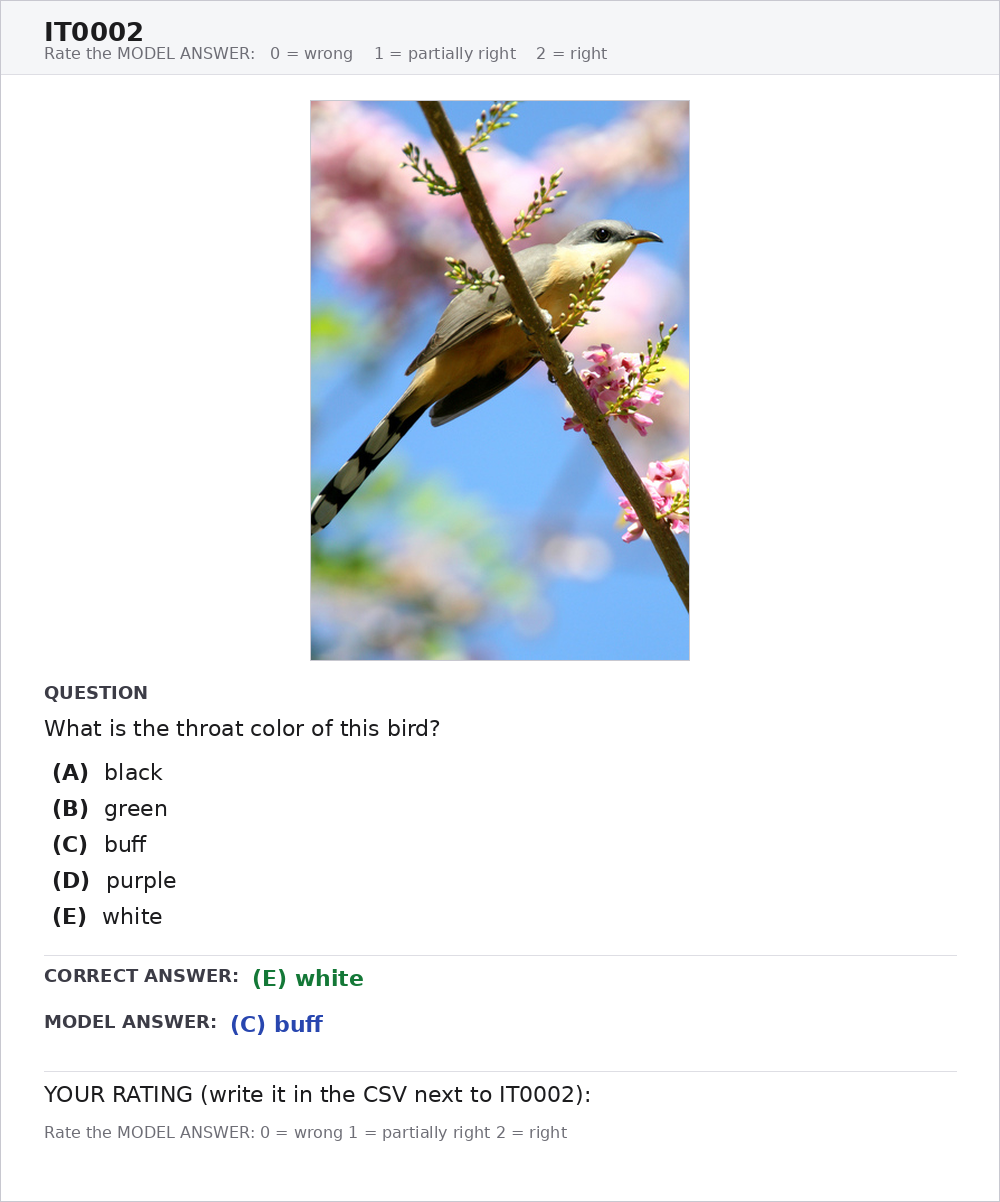}
  \caption{The annotators see the image, the question, the reference and the
  response. They do not see the model identity, the metric value or the stratum.}
  \label{fig:supp-card}
\end{figure}

\begin{figure}[t]
  \centering
  \includegraphics[width=.9\columnwidth]{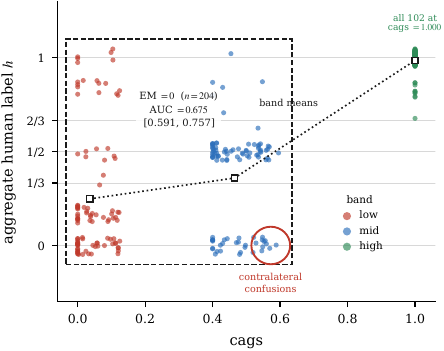}
  \caption{Inside the $\EM$-zero region, \cags{} separates credited from
  rejected answers. The three circled points are the laterality items, which are
  the metric's one systematic disagreement with the annotators.}
  \label{fig:supp-humanscatter}
\end{figure}

\paragraph{The \ctax{} synonym audit.}
Every soft credit is re-verified against GBIF by a separate script that reads the soft column the scorer wrote and cannot conjure credits the scorer did not compute. In the audited run, 100\% of the roughly \num{130000} soft credits resolve to a valid old-to-new substitution and zero are flagged for review. The order build, run, check is a data dependency. The build stage writes the synonym map. The run stage is the only stage that computes the soft column. The check stage reads what the run stage wrote.

\paragraph{\cspat{} floors and shortcut controls.}
Each \cspat{} task carries at least one control that closes a specific route to a high score without reading the image. Absolute distance carries three, described in \cref{sec:supp-metrics}. Counting carries the rank correlation, which is exactly zero for any constant predictor. Viewpoint carries three floors that differ by a factor of three, so a score quoted without its floor is not interpretable. Localization carries the mean gold box. OCR carries the constant string. We fit every floor on the same row subset as the model under test, so a model with a lower parse rate is compared against a baseline recomputed on its own rows.

\paragraph{Reading \cags{} and $\mathrm{GS}_{\mathrm{raw}}$ together.}
We report the mean raw score alongside \cags{}. The gap $\mathrm{GS}_{\mathrm{raw}} - \overline{\cags{}}$ is large exactly when a family's questions are easy to look good on, which happens when the distractors sit close to the ground truth. That gap is a diagnostic of question design and not of the model, and we report it so a reader can tell the two apart.

\section{Cross-Component Analysis}
\label{sec:supp-cross}

\subsection{Construction}

We build an $18 \times 24$ matrix of model by task scores, using each task's headline metric. We correlate every pair of tasks with Spearman $\rho$ over the eighteen models. We then summarize within-component and across-component agreement, and we run a principal component analysis on the standardized matrix.

Correlating tasks over 18 models is a small-$n$ analysis. The intervals are wide. They are wide enough to separate correlated pairs from uncorrelated ones and not wide enough to rank one pair above another. We treat the block structure as the finding and individual pairwise values as
descriptive.

\subsection{Tasks agree more inside a component than across one}

Task pairs within a component correlate at mean $\rho = .749$. Pairs drawn from different components correlate at $.534$. A label permutation test puts $p = .0005$ on that gap.

\cref{tab:supp-crossinternal} gives internal agreement per component.

\begin{table}[t]
  \centering\footnotesize
  \setlength{\tabcolsep}{5pt}
  \begin{tabular}{lrr}
    \toprule
    Component & Internal $\rho$ & Krippendorff $\alpha$ \\
    \midrule
    \ctax{}  & .841 & .981 \\
    \cspat{} & .662 & .925 \\
    \cattr{} & .645 & .901 \\
    \creid{} & .617 & .889 \\
    \bottomrule
  \end{tabular}
  \caption{Within-component agreement. Tasks inside a component agree enough for the component score to summarize them, and differ enough to be worth reporting on their own.}
  \label{tab:supp-crossinternal}
\end{table}

\ctax{} is the most internally coherent component, which is consistent with the $10 \times 10$ group matrix in \cref{tab:supp-c1groupcorr}. \creid{} is the least, which is consistent with it being the component that dissociates.

\subsection{The component correlation matrix}

\cref{tab:supp-crossmatrix} gives the four-by-four matrix.

\begin{table}[t]
  \centering\footnotesize
  \setlength{\tabcolsep}{6pt}
  \begin{tabular}{lrrrr}
    \toprule
    Spearman $\rho$ & \ctax{} & \cattr{} & \cspat{} & \creid{} \\
    \midrule
    \ctax{}  & --- & .689 & .787 & $.433^{\dagger}$ \\
    \cattr{} & & --- & \textbf{.927} & $.514^{\dagger}$ \\
    \cspat{} & & & --- & .610 \\
    \creid{} & & & & --- \\
    \bottomrule
  \end{tabular}
  \caption{Component-level correlation over the 18 models. $^{\dagger}$ not significant after Holm correction. The mean across components is $.660$
  against $.691$ within them.}
  \label{tab:supp-crossmatrix}
\end{table}

\subsection{What the components buy is everything below the top}

\cref{tab:supp-crossmove} shows the movement statistics.
\begin{table}[t]
  \centering\footnotesize
  \setlength{\tabcolsep}{4pt}
  \begin{tabular}{lrr}
    \toprule
    Statistic & Value & Note \\
    \midrule
    Median model rank movement & 5.5 of 18 & across components \\
    Largest movement & 11 of 18 & Molmo2-8B \\
    Molmo2-8B \ctax{} percentile & 3rd & \\
    Molmo2-8B \cattr{} percentile & 79th & \\
    Phi-3.5 \ctax{} percentile & 8th & \\
    Phi-3.5 \creid{} percentile & 64th & \\
    \bottomrule
  \end{tabular}
  \caption{Rank movement between components. Three models appear in every component's top five. Below them the ranking reorders substantially, so anyone choosing a mid-field model cannot do it from one number.}
  \label{tab:supp-crossmove}
\end{table}
Molmo2-8B moves eleven of eighteen places, which is the largest movement in the
benchmark. It is the only model trained with explicit pointing supervision, and
that supervision buys grounding without buying taxonomy.

\subsection{Where the dissociations sit}
\cref{tab:supp-crossload} reports the lowest-loading task in each component.
\begin{table}[t]
  \centering\footnotesize
  \setlength{\tabcolsep}{4pt}
  \begin{tabular}{lll}
    \toprule
    Component & Lowest-loading task & Loading \\
    \midrule
    \creid{} & all five species & .091 to .185 \\
    \cattr{} & laterality & .154 \\
    \cspat{} & relative position & .168 \\
    \cspat{} & localization & .176 \\
    \ctax{} & gymnosperms & .183 \\
    \bottomrule
  \end{tabular}
  \caption{The tasks that load lowest on the shared factor. Each is a task we designed to isolate something the rest of the benchmark does not measure, and each sits where we designed it to sit.}
  \label{tab:supp-crossload}
\end{table}
We report this as a prediction that held and not as a post-hoc reading. The
adversarial \creid{} condition, the laterality gate, the relative-position
isolation and the OCR control were all specified before we saw any correlation.

\subsection{Imputation sensitivity}

Not every model has a score on every task. Two InternVL runs were superseded and we exclude them from \creid{} rather than deleting their files. Where a cell is missing we impute the component median and we re-run the analysis with the affected model dropped entirely. The block structure is unchanged under both treatments and the mean within-against-across gap moves by less than $.01$.

\begin{figure*}[t]
  \centering
  \includegraphics[width=\textwidth]{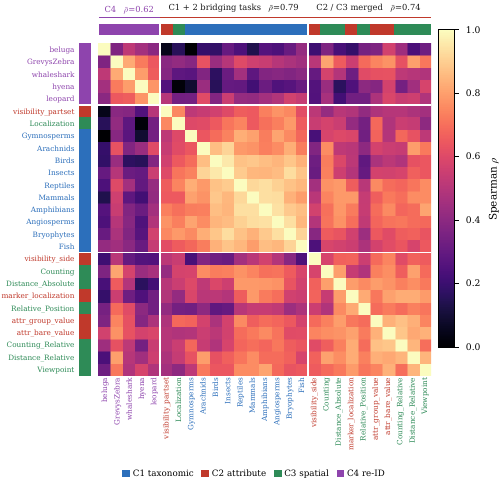}
  \caption{The effective structure is three blocks rather than four. \cattr{} and
  \cspat{} merge at the reliability we can measure. \creid{} separates.}
  \label{fig:supp-crossheat}
\end{figure*}

\begin{figure}[t]
  \centering
    \includegraphics[width=0.9\columnwidth]{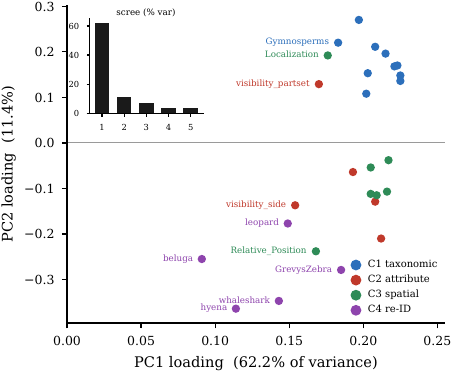}
  \caption{All tasks load positively on the first component, so a single
  ability drives most of the variance. The tasks with the lowest loadings are
  exactly the ones we designed to dissociate.}
  \label{fig:supp-crosspca}
\end{figure}

\begin{figure}[t]
  \centering
  \includegraphics[width=0.9\columnwidth]{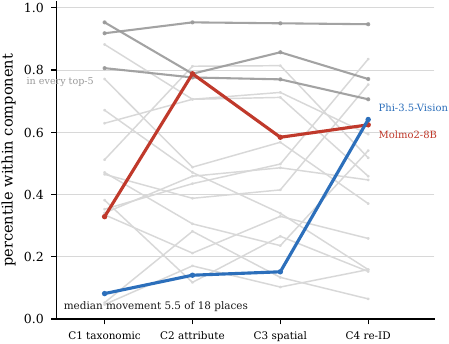}
  \caption{Three models stay in the top five everywhere. The median model moves
  5.5 of 18 places between components, so one leaderboard answers which model is
  best and not which model is best for a given task.}
  \label{fig:supp-crossslope}
\end{figure}

\section{Prompts and Parsing}
\label{sec:supp-prompts}

Every model receives the same question and the same answer-format instruction within a question family. We list both here exactly as they appear in the released question files and inference scripts. The instruction is appended to the question in the user turn. Three model-specific wrappers apply on top of these strings: Gemma-3 and Gemma-4 add the system message \texttt{You are a precise taxonomy expert. Reply with only the requested term.} on \ctax{} and \cattr{}; InternVL3-8B, InternVL3-38B and InternVL3.5-38B prefix \texttt{Please answer in English.}; InternVL uses its checkpoint's default system message, and the other models use none. \creid{} uses no system message for any model. All runs use greedy decoding.

\subsection{\ctax{} prompts}

Each item draws one question template at random (seed 42), with \{rank\} set to the asked rank (\emph{species epithet} for the species rank) and \{label\} set to \emph{animal}, \emph{plant} or \emph{organism} by kingdom.

\noindent\textit{Question templates (phylum to species epithet).}
\begin{quote}
  \footnotesize\ttfamily
  What is the \{rank\} of the \{label\} shown in this image? \\
  Which \{rank\} does the \{label\} in this image belong to? \\
  Identify the \{rank\} of the \{label\} shown in this image. \\
  Name the \{rank\} of the \{label\} in this image. \\
  What is the correct \{rank\} label for the \{label\} in this image?
\end{quote}

\noindent\textit{Question templates (scientific name and common name).}
\begin{quote}
  \footnotesize\ttfamily
  What is the \{rank\} of the \{label\} shown in this image? \\
  Identify the \{rank\} of the \{label\} shown in this image. \\
  Name the \{rank\} of the \{label\} shown in this image.
\end{quote}

\noindent\textit{Multiple-choice common name.} The four options are shuffled and listed without letters.
\begin{quote}
  \footnotesize\ttfamily
  Among the four options, what is the common name of the \{label\} shown in this image? \{o1\}, \{o2\}, \{o3\}, \{o4\}
\end{quote}

\noindent\textit{Answer-format instruction, appended by rank.}
\begin{quote}
  \footnotesize\ttfamily
  \textrm{phylum:} Answer with ONLY the Phylum name (e.g., Arthropoda). Do not include class, order, or any other taxonomic ranks. Do not write full sentences. \\
  \textrm{class:} Answer with ONLY the Class name (e.g., Arachnida, Insecta). Do not include order, family, or any other taxonomic ranks. Do not write full sentences. \\
  \textrm{order:} Answer with ONLY the Order name (e.g., Araneae). Do not include family, genus, or any other taxonomic ranks. Do not write full sentences. \\
  \textrm{family:} Answer with ONLY the Family name (e.g., Araneidae). Do not include genus, species, or any other taxonomic ranks. Do not write full sentences. \\
  \textrm{genus:} Answer with ONLY the Genus name (capitalized, e.g., Argiope). Do not include the species epithet or family name. Do not write full sentences. \\
  \textrm{species epithet:} Answer with ONLY the species epithet (the single, lowercase second word of the scientific name, e.g., aurantia). Absolutely DO NOT include the capitalized Genus name. Do not write full sentences. \\
  \textrm{scientific name:} Answer with ONLY the full binomial scientific name (Genus and species, e.g., Argiope aurantia). Do not include the common name or higher taxonomic ranks. Do not write full sentences. \\
  \textrm{multiple choice:} Answer with ONLY the exact correct option from the four choices provided in the question. Absolutely DO NOT write full sentences, and do not provide any extra context or scientific names. \\
  \textrm{common name:} Answer with ONLY the everyday English common name (e.g., Orb weavers). Absolutely DO NOT provide the Latin scientific name. Do not write full sentences.
\end{quote}

Generation is capped at 15 new tokens. The inference scripts keep only the last word of a multi-word species-epithet reply, which understates that rank; we analyze species-level naming through the scientific name.

\subsection{\cattr{} prompts}

\Cref{tab:supp-c2prompts} lists the ten \cattr{} question templates. Every multiple-choice item then receives the instruction below; open items receive the instruction for their family.

\begin{table}[t]
  \centering
  \footnotesize
  \setlength{\tabcolsep}{3pt}
  \begin{tabular}{@{}>{\raggedright\arraybackslash}p{3.1cm}>{\raggedright\arraybackslash}p{4.7cm}@{}}
    \toprule
    Family & Template \\
    \midrule
    \texttt{attr\_bare\_value}
      & What is the \{part\} \{property\} of this bird? Options: (A) \dots \\
    \texttt{attr\_group\_value}
      & Which of the following is the most prominently visible physical trait of
        the bird in this image? Options: \dots \\
    \texttt{marker\_localization}
      & What is the body part of this bird marked with the \{color\} \{shape\}?
        Options: \dots \\
    \texttt{visibility\_partset}
      & Which of the following parts of the bird are visible in this image?
        Options: (A) \{x\}, (B) \{y\}, (C) \{z\}, (D) None, (E) All three \\
    \texttt{visibility\_side}
      & Which of the \{eye|leg|wing\}s of the bird are visible in this image?
        Options: (A) Left, (B) Right, (C) Both, (D) None \\
    \texttt{category\_mcq}
      & What type of bird is this in the image? Options: \dots \\
    \texttt{open\_common\_name}
      & What is the common name of the bird? \\
    \texttt{open\_scientific\_name}
      & What is the scientific name of the bird? \\
    \texttt{open\_genus}
      & What is the genus of the bird? \\
    \texttt{open\_kind\_of}
      & What kind of \{category\} is the bird in this image? \\
    \bottomrule
  \end{tabular}
  \caption{The ten \cattr{} question templates. The marker family specifies the marker using both color and shape, as neither cue alone is sufficient for identification.}
  \label{tab:supp-c2prompts}
\end{table}

\noindent\textit{Multiple-choice instruction.}
\begin{quote}
  \footnotesize\ttfamily
  [CRITICAL INSTRUCTIONS - NO VIOLATIONS ALLOWED]: \\
  1. Locate the 'Options:' section in the question above. \\
  2. Identify the single correct choice from that list. \\
  3. Output ONLY the matching option letter in parentheses and its exact trailing text string. For example, if the correct option is '(X) alpha bravo charlie', your entire response must be exactly: (X) alpha bravo charlie \\
  4. DO NOT truncate the response to just the option letter (e.g., do not output just '(X)'). \\
  5. DO NOT invent, hallucinate, or insert any placeholder words from external examples. Absolutely DO NOT output any word unless it is explicitly written inside the chosen option string above. \\
  6. Output exactly zero extra words, zero markdown formatting, and zero punctuation outside the option string.
\end{quote}

\noindent\textit{Open-ended instructions.}
\begin{quote}
  \footnotesize\ttfamily
  \textrm{scientific name:} Answer with ONLY the scientific name of the bird. Do not write full sentences. \\
  \textrm{genus:} Answer with ONLY the capitalized Genus name of the bird. Do not write full sentences. \\
  \textrm{common name, kind of:} Answer with ONLY the everyday English common name of the bird. Do not write full sentences. \\
  \textrm{other:} Answer directly and concisely with only the exact terms requested. Do not write full sentences.
\end{quote}

\subsection{\cspat{} prompts}

Each \cspat{} prompt is a frame describing the image, the question, and an answer-format instruction, in that order. \Cref{tab:supp-c3prompts} gives the question for each task; the frames and instructions follow.

\begin{table}[t]
  \centering
  \footnotesize
  \setlength{\tabcolsep}{3pt}
  \begin{tabular}{@{}>{\raggedright\arraybackslash}p{2.0cm}>{\raggedright\arraybackslash}p{5.7cm}@{}}
    \toprule
    Task & Question \\
    \midrule
    Counting
      & How many \{animals\} are there in this image? \textrm{(two-panel tiers:} \dots\ in the left/right photograph?\textrm{)} \\
    Relative counting
      & Which of these 4 images contains \{most|least\} number of \{animals\}: Image A (top-left), Image B (top-right), Image C (bottom-left), Image D (bottom-right)? \\
    Absolute distance
      & What is the measurement of the elytra \{length|width\} (marked with red line with white border) with respect to the scalebar (marked with blue line with green border) in cm? \\
    Relative distance
      & Among the four highlighted beetle specimens, which one has the largest elytral \{length|width\}? Beetle A (red dotted box/line), Beetle B (blue dotted box/line), Beetle C (green dotted box/line), and Beetle D (white dotted box/line) \\
    Localization
      & Provide bounding box coordinates for the subject animal (\{species\}) in this image. \\
    Relative position
      & This image is marked with six boxes: two red, two blue, one green and one cyan. Exactly one red box and exactly one blue box contain a specimen; the other red box and the other blue box are empty. The green and cyan boxes are irrelevant to the question. Where is the specimen inside the red box located with respect to the specimen inside the blue box? Options: \{four of the eight directions\} \\
    Viewpoint
      & What is the viewpoint of the animal (red bounding box) shown in this image? \textrm{(other tiers:} What is the viewpoint of the \{species\} in this image? \textrm{/} \dots\ marked in the \{left|middle|right\} bounding box?\textrm{)} \\
    OCR
      & What is the text under the barcode (starting with A0...) shown in this image? \\
    \bottomrule
  \end{tabular}
  \caption{The eight \cspat{} questions. Each names the target, so a non-localization question does not require the model to find the subject first.}
  \label{tab:supp-c3prompts}
\end{table}

\noindent\textit{Frames.}
\begin{quote}
  \footnotesize\ttfamily
  \textrm{counting, two-panel tiers:} This image consists of two separate photographs placed side by side. Count only within the photograph named in the question and ignore the other one. \\
  \textrm{relative counting:} This image is a 2x2 grid of four separate photographs: Image A - top-left, Image B - top-right, Image C - bottom-left, Image D - bottom-right. \\
  \textrm{absolute distance:} This image shows a plate of pinned beetle specimens, photographed at a single scale. A BLUE line marks the scale bar. The scale bar is exactly 1 cm long. A RED line marks the measurement asked about below. \\
  \textrm{relative distance:} This image shows a single plate of pinned beetle specimens, photographed at one magnification. Four specimens are marked, each with a dotted box and a coloured line drawn along its elytra: Beetle A - red, Beetle B - blue, Beetle C - green, Beetle D - white. \\
  \textrm{localization:} The whole image is \{W\} pixels wide and \{H\} pixels tall. \\
  \textrm{viewpoint, three-panel tiers:} This image consists of three separate photographs placed side by side. \\
  \textrm{OCR:} This image shows a specimen label with a barcode and a printed accession code beneath it.
\end{quote}

\noindent\textit{Answer-format instructions.}
\begin{quote}
  \footnotesize\ttfamily
  \textrm{counting:} Count every individual, including any that are partially visible or cut off at the edge of the photograph. Give a single best estimate even if you are uncertain. Respond with one integer and nothing else: no words, no units, no ranges. If you count seven, respond: 7 \\
  \textrm{relative counting:} Count the animals in each of the four photographs and compare the totals. The photographs are shown at different scales, so compare how many animals there are, not how large they appear. Exactly one photograph is the answer. Choose it even if the difference is small. Answer immediately with the letter. Do not describe the photographs and do not explain your reasoning. Respond with one letter and nothing else: A, B, C or D. \\
  \textrm{absolute distance:} Compare the pixel length of the red line to the pixel length of the blue scale bar. Because the scale bar is exactly 1 cm long, the red line's length in centimetres equals that ratio: if the red line is half as long as the blue line the answer is 0.50 cm, and if it is twice as long the answer is 2.00 cm. Measure the two lines as they appear in this image. Do not answer from what you know about typical beetle sizes. Respond with one number rounded to two decimal places followed by cm, and nothing else. No explanation, no range, no other units. Example: 0.87 cm \\
  \textrm{relative distance:} Compare the lengths of the four coloured lines as they appear in this image. Because all four specimens are in the same photograph at the same magnification, the line drawn on screen is directly proportional to the real measurement - no scale conversion is needed. Exactly one of the four is the largest. Choose it even if the difference is small. Respond with one letter and nothing else: A, B, C or D. \\
  \textrm{localization:} Respond with four numbers in this exact format and nothing else: [x, y, width, height] where x and y are the pixel coordinates of the top-left corner of the box, measured from the top-left corner of the WHOLE image described above, and width and height are the size of the box in pixels. Do not use fractions, percentages or normalised coordinates. Do not explain. Example: [1240, 88, 540, 410] \\
  \textrm{relative position:} Answer with one of the four options exactly as written above, in lower case, and nothing else. Do not describe the boxes, do not explain your reasoning, do not restate the question. For example, if the answer is the option bottom-left, reply: bottom-left \\
  \textrm{viewpoint:} Answer with the direction the animal is facing, relative to the camera. Choose exactly one of these accepted answers: front, back, left, right, frontleft, frontright, backleft, backright, up, down. Reply with one of those options, written exactly as above in lower case, and nothing else. No explanation, no punctuation. For example, if the animal faces away from the camera and to the left, reply: backleft \\
  \textrm{OCR:} Read the printed characters underneath the barcode. The code is always the letter A followed by eleven digits, twelve characters in total, with no spaces or punctuation. Transcribe exactly what is printed. Do not guess a plausible code and do not correct what you see. Respond with the twelve-character code and nothing else. Example: A00000012345
\end{quote}

\subsection{\creid{} prompt}

Every pair uses the same question and instruction.

\begin{quote}
  \footnotesize\ttfamily
  This image shows two images placed side by side. Is the main subject in the left image the very same individual animal as the main subject in the right image? \\
  * The image shows two photographs placed side by side. \\
  * Each photograph may contain more than one animal - consider only the main subject in the foreground. \\
  * Judge only from what is visible in the two photographs. \\
  * Answer with one word: Yes or No. Do not explain.
\end{quote}

Because we rely on a single phrasing, as discussed in \cref{sec:supp-limits}, this design cannot disentangle an architectural bias toward answering \emph{Yes} from artifacts introduced by the prompt's specific wording.

\subsection{The parser}

The \cattr{} scorer resolves every reply in the five steps below. The other components parse the answer forms listed in \cref{tab:supp-parse}.

\noindent\textbf{Step 1: Normalize.}
We convert text to lower-case, map underscores to spaces, standardize \emph{grey} and \emph{gray}, strip all punctuation (excluding colons, hyphens, and parentheses), and collapse whitespace. Consequently, \textit{Blue\_Jay}, \textit{blue jay}, and \textit{blue jay.} are treated as identical.

\noindent\textbf{Step 2: Detect the format.}
The literal token \textit{Options:} routes an item into the multiple-choice branch, which constructs an ordered letter-to-text mapping. All other inputs are treated as open-ended.

\noindent\textbf{Step 3: Parse the gold standard.}
We process gold answers stored either in the \textit{(A) text} format or as bare values.

\noindent\textbf{Step 4: Resolve the prediction.}
This step is highly consequential and deliberately permissive to ensure visual competence is not penalized for formatting quirks. In order, we: strip conversational scaffolding and retain the first line; detect any explicit letter markers; identify verbatim option text; and reconcile the two. If $\EM$ fail, we fall back to a fuzzy string match with a $0.86$ threshold; otherwise, we record a parsing failure and assign a score of zero. In cases where the letter and the text disagree, we record a mismatch and break the tie in favor of the letter, in accordance with our default policy.

\noindent\textbf{Step 5: Route to a family.}
We match template phrases in priority order. The marker phrase routes to \textit{marker\_localization}, while the salience phrase routes to
\textit{attr\_group\_value}. The visibility phrase routes to either \textit{visibility\_side} or \textit{visibility\_partset}, depending on whether
the options consist of the four side labels. Naming phrases route to the naming families, and a bare attribute question with provided options routes to \textit{attr\_bare\_value}. Both the graded and exact-match evaluators execute Steps 1 through 5 verbatim, ensuring both metrics partition the \num{824104} \cattr{} items identically. Consequently, \cref{tab:supp-c2fam,tab:supp-c2famem} are comparable cell-by-cell.

\subsection{Per-item outputs}

We save every per-item quantity to disk. For \cattr{}, this includes the family, category, gold and predicted text, resolution status, $\EM$, raw similarity $s$, chance baseline $b$, \cags{}, and severity. For \ctax{}, we record the hard match, soft match, resolution mechanism, LCA height, and both Wu--Palmer and Lin similarities. For \cspat{}, we log the parsed prediction, gold standard, per-task metric, and the floor computed on the same rows. Finally, for \creid{}, we store the confusion counts, parse tier, $\dpr$, criterion $c$, and the bootstrap interval. All results presented in this supplement are fully reproducible from these files.

\section{Reproducibility}
\label{sec:supp-repro}

\subsection{Determinism and Reproducibility}

\cref{tab:supp-det} delineates which stages of our evaluation pipeline reproduce exactly and which involve non-deterministic elements.

\begin{table}[t]
  \centering\footnotesize
  \setlength{\tabcolsep}{3pt}
  \begin{tabular}{p{2.0cm}cp{4.2cm}}
    \toprule
    Stage & Det. & Mechanism \\
    \midrule
    Image download & \xmark & Thread interleaving dictates which concurrent downloads fulfill the allocated budget. The finalized manifests serve as the artifact of record. \\
    QA generation & \cmark & Deterministically seeded per component. \\
    Manifest sharding & \cmark & Seed 42, applied subsequent to row identifier assignment. \\
    Inference & \cmark & Greedy decoding, evaluated zero-shot with one attempt per item. \\
    Difficulty tiers & \cmark & Strictly deterministic. Tie-breaking relies on taxon nomenclature. \\
    Scoring & \cmark & A pure function applied to the merged prediction files. Bootstraps are fixed-seed. \\
    \bottomrule
  \end{tabular}
  \caption{Determinism across the evaluation pipeline. Given a finalized manifest,
  all downstream operations are exactly reproducible. Consequently, we archive and distribute the
  manifests rather than expecting researchers to regenerate identical image sets byte-for-byte.}
  \label{tab:supp-det}
\end{table}

The sole non-deterministic stage is the \ctax{} image acquisition. We query the source catalog, isolate species with complete taxonomic hierarchies, and initiate concurrent downloads constrained by a per-group image budget. The specific subset of images that fulfills this budget depends entirely on asynchronous thread interleaving. Therefore, we treat the resulting manifest as the definitive dataset artifact and release it directly.

\subsection{The Verification Contract}

A data row is instantiated in a \ctax{} manifest if and only if its corresponding image is successfully written to disk and every taxonomic rank required by the generator is populated (non-null). An independent audit stage mathematically guarantees this contract holds prior to downstream execution, systematically scanning every manifest for missing, null, or empty ranks.

\subsection{Bootstrap Resampling Procedure}

All confidence intervals reported in this supplement are derived via percentile bootstrapping using fixed seeds. For \cattr{}, we resample the per-item \cags{} with replacement $B = 1000$ times, reporting the $2.5$ and $97.5$ percentiles of the resampled mean distribution. For \creid{}, we resample items within each model-dataset-condition stratum $B = 2000$ times. For the human baseline study, we resample items $B = 5000$ times (seed 42), employing paired resampling to evaluate comparative differences. For \cspat{} counting and absolute distance metrics, we perform cluster bootstrapping at the source-plate level rather than the individual row level, as multiple spatial questions reference the same underlying photograph. We apply a strict interpretation rule throughout our analysis: if the confidence intervals of two models overlap on a given metric, they are considered statistically tied, meaning raw ranking differences are not inherently meaningful.

\subsection{Codebase and Directory Structure}

Each task directory adheres to a standardized four-stage structure. The first stage synthesizes question-answer pairs from raw annotations. The second stage executes inference across the eighteen models, initiated by a preprocessing step that assembles the manifest and triggers the pre-run audit. The final stage computes metrics and generates summary statistics. 

\subsection{Metric Regeneration}

The scoring stage operates as a pure function parameterized by the stored model outputs. Given the released prediction files, every table and metric in this supplement can be regenerated by executing the four component scorers sequentially. Note that the \ctax{} scorer strictly requires the synonym map to be built beforehand; omitting this map silently degrades soft-match evaluation to hard-match evaluation rather than raising an explicit error. This dependency ordering is explicitly documented in the component README.

\section{Extended Limitations}
\label{sec:supp-limits}

The main manuscript briefly outlines nine limitations. Here, we expand upon those points and detail additional component-level constraints omitted for brevity.

\subsection{Benchmark-Level Limitations}

\paragraph{Exclusive Evaluation of Open-Weight Models.}
We evaluate 18 open-weight VLMs and no proprietary systems (e.g., GPT-4V, Gemini 1.5 Pro). Consequently, we cannot verify whether frontier closed models exhibit the same behavioral pathologies. This primarily impacts the scaling claims (our largest evaluated model is 38.4B) and the granularity claims, as a significantly larger proprietary model might resolve fine-grained taxonomic ranks that our evaluated models fail to capture. Claims derived from item floors and the adversarial conditions remain robust, as they reflect properties of the evaluation set rather than the model population.

\paragraph{Single Decoding Protocol.}
Our evaluation relies on greedy decoding without chain-of-thought prompting, few-shot exemplars, or multiple inference attempts. Models capable of self-correction or advanced reasoning modes may achieve higher accuracy than reported. However, we intentionally measure zero-shot, single-pass performance, as this aligns with the compute constraints and operational reality of most automated biodiversity deployments.

\paragraph{Template-Generated Prompts.}
Every query is synthesized from a template parameterized by an annotation. This introduces the risk that a model might infer the correct answer from linguistic cues rather than visual reasoning. However, the tasks that best separate model performance (\eg, open-ended naming, marker localization, and adversarial pair verification) cannot be deduced from the prompt text, as the answers are inherently visually contingent. This risk is most pronounced in the easier multiple-choice families, particularly \textit{category\_mcq}, which achieves a mean \cags{} of $.833$.

\paragraph{Potential Pretraining Data Contamination.}
Because our images are sourced from public biodiversity corpora, models may have encountered them during pretraining. It is infeasible to definitively audit the proprietary training mixtures of all 18 models. Critically, the bias introduced by data contamination is strictly upward (inflating scores). Because our primary findings center on where models systematically fail, these below-floor results survive the threat of contamination.

\paragraph{Small Sample Size in Cross-Component Analysis.}
Eighteen models represent a small statistical sample for a correlation matrix. Consequently, our intervals serve only to distinguish significantly correlated pairs from uncorrelated pairs, rather than establishing a strict ranking among correlations. The block-diagonal structure of the matrix is the primary finding, rather than individual pairwise magnitudes.

\subsection{\ctax{}}

\paragraph{Taxonomy Backbone Dependency.}
Our ground-truth hierarchy strictly follows the GBIF Backbone, which aligns with the Catalogue of Life. If a model accurately answers \emph{Reptilia} to a class-level question, it is technically reproducing valid textbook Linnaean taxonomy, yet the backbone has structurally evolved. We report the gap between strict and soft matching specifically to quantify this taxonomic lag, rather than unfairly penalizing the model's absolute accuracy.

\paragraph{Taxonomic Rank Inconsistencies.}
The \emph{order} column is natively unpopulated for every reptile record in our source data. Analytical tooling that rigidly assumes six fully populated ranks will silently discard the entire reptile class. To prevent this, our tree structure accommodates skip edges, attaching a node directly to its nearest populated ancestor.

\paragraph{Uniform Distractor Sampling.}
The distractor pools for multiple-choice questions are unioned across genus, family, order, and class, and then sampled uniformly. Consequently, the class-level set statistically dominates the distractors. The MCQ items should therefore be interpreted as coarse recognition probes. Analysts requiring graded discrimination should condition their evaluation on the provided confusability tiers.

\paragraph{Gymnosperm Representation.}
There are only roughly one thousand accepted gymnosperm species worldwide. Securing \num{10000} images required significantly relaxing our sampling constraints. Notably, gymnosperms represent the sole dissociation in the cross-group correlation matrix and exhibit the lowest loading on the shared \ctax{} factor. We retain and report this group for completeness, but attribute the dissociation to data coverage constraints.

\subsection{\cattr{}}

\paragraph{Clade-Specific Attribute Grounding.}
The CUB-200-2011 dataset consists entirely of avian species. The generalizability of these attribute grounding results to fundamentally different body plans (e.g., invertebrates, mammals) remains untested. 

\paragraph{Vocabulary Constraints.}
Color, pattern, and shape lexicons strictly adhere to the original CUB annotation schema. For example, the limitation to fifteen discrete color words restricts the granularity of the evaluation. Expanding this vocabulary would fundamentally alter the option distributions and the associated chance baselines.

\paragraph{Kernel Dependency in \cags{}.}
While the sensitivity sweep demonstrates ranking stability ($\rho \geq .9959$), absolute \cags{} values remain kernel-dependent. A reported \cags{} of $.593$ should therefore be interpreted as relative positioning within the evaluated field, not as a perfectly calibrated absolute probability.

\paragraph{Laterality Bias in the Anatomy Kernel.}
The metric inherently over-credits lateral proximity: confusing a left wing for a right wing yields a score of $.883$ due to their spatial closeness in the anatomy layout. This represents the metric's single systematic divergence from human annotator intuition, which we quantify via the human study. To mitigate this, we compute the laterality analysis using $\EM$ criteria and deliberately exclude \texttt{marker\_localization} from the primary aggregate score, where this bias would otherwise dominate.

\paragraph{Human Ceiling Estimation Constraints.}
The human performance ceiling ($.908$) is measured against a stratified subset of 306 items, not the entire corpus. This figure serves to validate the mathematical metric rather than establishing a definitive human benchmark for the entire dataset.

\paragraph{Approximated Open-Ended Baselines.}
The chance baseline for open-ended queries is established by sampling a fixed-seed pool of 40 alternative gold answers within that family. While stable under fixed-seed resampling, it remains a statistical approximation of the true full-distribution chance baseline.

\paragraph{Entanglement with \cspat{}.}
The disattenuated correlation between \cattr{} and \cspat{} is 1.015. At this level of statistical reliability, the two components cannot be differentiated by empirical model behavior alone; they are distinguished solely by their semantic task content.

\subsection{\cspat{}}

\paragraph{Dataset and Condition Specificity in Relative Position.}
The relative position task is executed exclusively on the beetle subset under the distractor condition. Lacking a clean (distractor-free) contrast baseline, we cannot statistically isolate the difficulty of the spatial relation reasoning from the visual interference introduced by the distractors.

\paragraph{Corpus Overfitting Risks.}
Absolute distance, relative distance, and relative position are evaluated solely on the beetle corpora, as this is the only subset possessing the requisite ground truth. Any idiosyncratic properties of this corpus (such as a fixed focal depth or highly uniform backgrounds) are unavoidably confounded with the spatial tasks themselves.

\paragraph{Indicative Nature of Counting Comparisons.}
Standard benchmarks evaluate counting capabilities using five-option multiple choice over sparse scenes. We challenge models with open numeric generation over dense, heavily occluded scenes of visually homogeneous animals. The observed three-fold performance gap is highly indicative of failure but is not perfectly controlled against standard sparse-scene performance.

\paragraph{OCR Control Image Sharing.}
The OCR control task utilizes the same beetle trays employed in the spatial evaluations, as these trays naturally feature specimen labels. A control task sourced from a completely independent corpus would provide a more robust decoupled baseline.

\subsection{\creid{}}

\paragraph{Embedding Circularity.}
The difficulty conditions are explicitly defined by DINOv2 cosine similarity margins. Several evaluated models share architectural lineage or self-supervised training objectives with DINOv2. Consequently, our difficulty manipulation partially indexes shared embedding geometries rather than universal properties of visual identity representation. While we do not replicate the construction across a second embedding space, the MegaDescriptor result tightly bounds this concern: as a system trained on an entirely disparate objective, it exhibits the exact same failure modes.

\paragraph{Absence of a Definitive Performance Ceiling.}
The specialist model effectively ties, rather than definitively exceeding, the best VLM on neutral pairs. Thus, we have not established the absolute theoretical upper bound for performance on these image pairs. Our claims remain strictly comparative across the evaluated models and conditions.

\paragraph{Single Prompt Formulation.}
The near-universal bias toward affirmative responses ("yes") in the hard condition may stem from architectural visual-matching tendencies, or it could be an artifact of our specific prompt wording. Testing a single prompt formulation prevents us from cleanly decoupling the two.

\paragraph{Composite Image Formatting.}
To test identity, we concatenate pairs into a single side-by-side image rather than passing two separate image tokens. This introduces an auxiliary spatial-parsing burden to the identity judgment. However, because this formatting is held constant across all models and all difficulty conditions, it does not confound the relative difficulty effect, even if it marginally suppresses absolute accuracy levels.

\paragraph{Bias Toward Pattern-Based Identity.}
The five selected species primarily communicate individual identity through strong pattern-based visual signals. Species whose identification relies strictly on morphological shape variance or incidental scarring remain unrepresented, with the beluga whale serving as the closest conceptual proxy in our set.

\paragraph{Single Operating Point Estimation.}
Because models output a discrete binary token, we cannot generate a full ROC curve; $\dpr$ is estimated from a single hit/false-alarm pair. While the applied log-linear correction is standard psychophysical practice, it can disproportionately influence estimates in extreme edge-case cells.

\paragraph{Multiple Hypothesis Testing.}
Evaluating eighteen models across five species and three pairwise contrasts yields several hundred statistical comparisons, guaranteeing that some nominally significant per-cell results will arise by chance. Consequently, per-cell results should be interpreted as exploratory, while the pooled, aggregate tests serve as our confirmatory findings.

\paragraph{Item Non-Independence.}
Certain source images are utilized as anchors in multiple pairwise comparisons. Bootstrap confidence intervals computed under the assumption of strict independence will therefore be marginally optimistic.

\paragraph{Assumptions Underlying $\dpr$.}
Our analysis relies on the assumption of equal-variance Gaussian evidence distributions. We utilize $\dpr$ primarily as a bias-corrected measure of effect size, rather than asserting it acts as a perfect proxy for internal model representations.

\subsection{Future Directions}

We highlight three critical augmentations that would significantly mature this benchmark protocol:

First, evaluating frontier proprietary models (e.g., GPT-4o, Gemini 1.5 Pro) on a stratified subset would conclusively demonstrate whether taxonomic granularity collapse is a universal limitation of current multimodal architectures, or merely a constraint of open-weight parameter scales.

Second, establishing a rigorous human ceiling on a stratified sample of \ctax{} and \cspat{} tasks would allow analysts to benchmark model deficits against genuine domain-expert performance, rather than solely against algorithmic specialists.

Finally, replicating the \creid{} adversarial pair construction within a disparate, non-DINO embedding space (e.g., CLIP, SigLIP) would definitively close the circularity concern, confirming that the observed failure modes stem from fundamental appearance-matching heuristics rather than shared geometric artifacts.

\section{Additional Tables}

The sections above report the statistic that carries each argument. This section
reports the rest. Every number here comes from the same scored files. Colors mark
the \bst{best}, \snd{second best}, \swst{second worst} and \wst{worst} over the
18 generalists, and the two shaded specialist rows are excluded from that
ranking.

\clearpage

\begin{table*}[tbp]
  \centering\scriptsize
  \setlength{\tabcolsep}{3pt}
  \begin{tabular}{@{}lrrrrrrrrr@{}}
    \toprule
    Model & phy & cls & ord & fam & gen & spc & sci & com & mcq \\
    \midrule
    \rowcolor{rowtint}
    BioCLIP-2 \texttt{kingdom}  & .985 & .871 & .816 & .772 & .691 & .547 & .546 & .502 & .966 \\
    \rowcolor{rowtint}
    BioCLIP-2 \texttt{tol}      & .925 & .874 & .821 & .775 & .691 & .545 & .544 & .500 & .966 \\
    \midrule
    Qwen3-VL-30B-A3B            & .864 & \bst{.731} & \bst{.757} & \bst{.612} & .417 & \bst{.072} & .221 & \bst{.107} & .772 \\
    Qwen3-VL-32B                & .831 & .615 & .712 & \bst{.612} & \bst{.444} & \snd{.066} & \snd{.244} & \snd{.078} & \bst{.809} \\
    Qwen3-VL-8B                 & \snd{.887} & .680 & \snd{.720} & \snd{.589} & .406 & .046 & .202 & .065 & \snd{.775} \\
    InternVL3-38B               & \bst{.911} & .683 & .638 & .484 & .355 & .054 & .189 & .044 & .752 \\
    Qwen2.5-VL-7B               & .763 & .504 & .673 & .541 & \snd{.420} & .047 & \bst{.269} & .062 & .753 \\
    Gemma-3-12B                 & .790 & \snd{.688} & .626 & .429 & .225 & .048 & .093 & .047 & .687 \\
    InternVL3.5-38B             & .746 & .636 & .624 & .363 & .271 & .038 & .128 & .025 & .707 \\
    Gemma-4-12B                 & .880 & .686 & .578 & .363 & .183 & .021 & .065 & .043 & .657 \\
    InternVL3-8B                & .827 & .497 & .478 & .322 & .262 & .032 & .124 & .020 & .643 \\
    Qwen2.5-VL-3B               & \swst{.416} & .457 & .477 & .454 & .358 & .040 & .206 & .064 & .689 \\
    Llama-3.2-Vision            & .545 & .482 & .518 & .434 & .234 & .031 & .104 & .054 & .655 \\
    InternVL3.5-8B              & .819 & .531 & .421 & .266 & .191 & .023 & .077 & .015 & .595 \\
    Molmo2-8B                   & .823 & .563 & .469 & .286 & .137 & .021 & .038 & .016 & .574 \\
    Pixtral-12B                 & .688 & .501 & .530 & .349 & .187 & .029 & .079 & .027 & .592 \\
    InternVL3.5-30B-A3B         & .486 & .469 & .545 & .310 & .240 & .034 & .118 & .024 & .653 \\
    Phi-3.5-Vision              & .748 & \swst{.359} & \wst{.267} & .191 & .066 & .013 & \swst{.023} & .009 & \swst{.443} \\
    LLaVA-NeXT-Mistral          & \wst{.364} & .392 & .408 & \swst{.162} & \swst{.029} & \swst{.005} & \wst{.015} & \swst{.004} & .465 \\
    LLaVA-NeXT-Vicuna           & .662 & \wst{.343} & \swst{.290} & \wst{.100} & \wst{.028} & \wst{.003} & \wst{.015} & \wst{.003} & \wst{.382} \\
    \bottomrule
  \end{tabular}
  \caption{$\EM$ by rank for every configuration. \cref{tab:supp-c1ranks} gives the column means. Every model falls monotonically from phylum to the species epithet, and the common-name MCQ column sits far above the open common-name column for all 20.}
  \label{tab:supp-t-emrank}
\end{table*}

\begin{table*}[tbp]
  \centering\scriptsize
  \setlength{\tabcolsep}{3pt}
  \begin{tabular}{@{}lrrrrrrrrr@{}}
    \toprule
    Model & phy & cls & ord & fam & gen & spc & sci & com & mcq \\
    \midrule
    \rowcolor{rowtint}
    BioCLIP-2 \texttt{kingdom}  & .985 & .871 & .816 & .772 & .692 & .547 & .547 & .502 & .966 \\
    \rowcolor{rowtint}
    BioCLIP-2 \texttt{tol}      & .925 & .874 & .821 & .775 & .692 & .545 & .545 & .500 & .966 \\
    \midrule
    Qwen3-VL-30B-A3B            & .882 & \bst{.829} & \bst{.758} & \bst{.615} & .420 & \bst{.072} & .233 & \bst{.107} & .772 \\
    Qwen3-VL-32B                & .861 & .713 & .718 & \bst{.615} & \bst{.448} & \snd{.067} & \snd{.258} & \snd{.078} & \bst{.809} \\
    Qwen3-VL-8B                 & .895 & .776 & \snd{.721} & \snd{.592} & .409 & .047 & .213 & .065 & \snd{.775} \\
    InternVL3-38B               & \snd{.932} & .779 & .640 & .488 & .359 & .054 & .199 & .044 & .752 \\
    Qwen2.5-VL-7B               & .763 & .598 & .674 & .548 & \snd{.421} & .047 & \bst{.283} & .062 & .753 \\
    Gemma-3-12B                 & .928 & \snd{.783} & .629 & .431 & .229 & .048 & .101 & .047 & .687 \\
    InternVL3.5-38B             & .855 & .729 & .625 & .378 & .273 & .038 & .135 & .025 & .707 \\
    Gemma-4-12B                 & .880 & .761 & .579 & .365 & .185 & .021 & .070 & .043 & .657 \\
    InternVL3-8B                & .903 & .565 & .478 & .330 & .264 & .032 & .131 & .020 & .643 \\
    Qwen2.5-VL-3B               & \swst{.438} & .553 & .477 & .467 & .362 & .040 & .220 & .064 & .689 \\
    Llama-3.2-Vision            & .565 & .578 & .519 & .439 & .239 & .031 & .113 & .054 & .655 \\
    InternVL3.5-8B              & .918 & .622 & .423 & .272 & .193 & .023 & .083 & .015 & .595 \\
    Molmo2-8B                   & \bst{.953} & .644 & .472 & .287 & .142 & .021 & .041 & .016 & .574 \\
    Pixtral-12B                 & .736 & .579 & .547 & .350 & .191 & .029 & .085 & .027 & .592 \\
    InternVL3.5-30B-A3B         & .554 & .558 & .545 & .314 & .242 & .034 & .126 & .024 & .653 \\
    Phi-3.5-Vision              & .768 & \wst{.424} & \wst{.267} & .191 & \swst{.069} & .013 & .026 & .009 & \swst{.443} \\
    LLaVA-NeXT-Mistral          & \wst{.368} & .484 & .409 & \swst{.163} & \wst{.032} & \swst{.005} & \swst{.017} & \swst{.004} & .465 \\
    LLaVA-NeXT-Vicuna           & .671 & \swst{.435} & \swst{.290} & \wst{.101} & \wst{.032} & \wst{.003} & \wst{.015} & \wst{.003} & \wst{.382} \\
    \bottomrule
  \end{tabular}
  \caption{$\EMs$ by rank for every configuration, on the same items as \cref{tab:supp-t-emrank}. The two specialist rows barely move, because a specialist emits current names.}
  \label{tab:supp-t-emsrank}
\end{table*}

\begin{table*}[tbp]
  \centering\scriptsize
  \setlength{\tabcolsep}{3pt}
  \begin{tabular}{@{}lrrrrrrrrr@{}}
    \toprule
    Model & phy & cls & ord & fam & gen & spc & sci & com & mcq \\
    \midrule
    \rowcolor{rowtint}
    BioCLIP-2 \texttt{kingdom}  & .000 & .000 & .000 & .000 & .001 & .000 & .001 & .000 & .000 \\
    \rowcolor{rowtint}
    BioCLIP-2 \texttt{tol}      & .000 & .000 & .000 & .000 & .001 & .000 & .001 & .000 & .000 \\
    \midrule
    Qwen3-VL-30B-A3B            & .018 & \bst{.098} & \swst{.001} & .003 & .003 & \snd{.000} & .012 & \bst{.000} & \bst{.000} \\
    Qwen3-VL-32B                & .030 & \bst{.098} & \snd{.006} & .003 & \snd{.004} & \bst{.001} & \bst{.014} & \bst{.000} & \bst{.000} \\
    Qwen3-VL-8B                 & .008 & .096 & \swst{.001} & .003 & .003 & \bst{.001} & .011 & \bst{.000} & \bst{.000} \\
    InternVL3-38B               & .021 & .096 & .002 & .004 & \snd{.004} & \snd{.000} & .010 & \bst{.000} & \bst{.000} \\
    Qwen2.5-VL-7B               & \wst{.000} & .094 & \swst{.001} & .007 & \wst{.001} & \snd{.000} & \snd{.014} & \bst{.000} & \bst{.000} \\
    Gemma-3-12B                 & \bst{.138} & .095 & .003 & .002 & \snd{.004} & \snd{.000} & .008 & \bst{.000} & \bst{.000} \\
    InternVL3.5-38B             & .109 & .093 & \swst{.001} & \bst{.015} & \swst{.002} & \snd{.000} & .007 & \bst{.000} & \bst{.000} \\
    Gemma-4-12B                 & \wst{.000} & .075 & \swst{.001} & .002 & \swst{.002} & \snd{.000} & .005 & \bst{.000} & \bst{.000} \\
    InternVL3-8B                & .076 & \swst{.068} & \wst{.000} & .008 & \swst{.002} & \snd{.000} & .007 & \bst{.000} & \bst{.000} \\
    Qwen2.5-VL-3B               & .022 & \snd{.096} & \wst{.000} & \snd{.013} & \snd{.004} & \snd{.000} & \bst{.014} & \bst{.000} & \bst{.000} \\
    Llama-3.2-Vision            & .020 & .096 & \swst{.001} & .005 & \bst{.005} & \snd{.000} & .009 & \bst{.000} & \bst{.000} \\
    InternVL3.5-8B              & .099 & .091 & .002 & .006 & \swst{.002} & \snd{.000} & .006 & \bst{.000} & \bst{.000} \\
    Molmo2-8B                   & \snd{.130} & .081 & .003 & \swst{.001} & \bst{.005} & \snd{.000} & .003 & \bst{.000} & \bst{.000} \\
    Pixtral-12B                 & .048 & .078 & \bst{.017} & \swst{.001} & \snd{.004} & \snd{.000} & .006 & \bst{.000} & \bst{.000} \\
    InternVL3.5-30B-A3B         & .068 & .089 & \wst{.000} & .004 & \swst{.002} & \snd{.000} & .008 & \bst{.000} & \bst{.000} \\
    Phi-3.5-Vision              & .020 & \wst{.065} & \wst{.000} & \wst{.000} & .003 & \snd{.000} & .003 & \bst{.000} & \bst{.000} \\
    LLaVA-NeXT-Mistral          & \swst{.004} & .092 & \swst{.001} & \swst{.001} & .003 & \snd{.000} & \swst{.002} & \bst{.000} & \bst{.000} \\
    LLaVA-NeXT-Vicuna           & .009 & .092 & \wst{.000} & \swst{.001} & .004 & \snd{.000} & \wst{.000} & \bst{.000} & \bst{.000} \\
    \bottomrule
  \end{tabular}
  \caption{$\EMs - \EM$ by rank, the difference between the two tables above. The gain concentrates at class and phylum for nearly every model and is zero from the species epithet rightward. \cref{tab:supp-c1lag} pools this by rank and by group.}
  \label{tab:supp-t-gaprank}
\end{table*}

\begin{table*}[tbp]
  \centering\scriptsize
  \setlength{\tabcolsep}{3pt}
  \begin{tabular}{@{}lrrrrrrrrrr@{}}
    \toprule
    Model & Amph & Angio & Arach & Birds & Bryo & Fish & Gymno & Insect & Mamm & Rept \\
    \midrule
    \rowcolor{rowtint}
    BioCLIP-2 \texttt{kingdom}  & .766 & .875 & .781 & .887 & .659 & .719 & .215 & .921 & .819 & .795 \\
    \rowcolor{rowtint}
    BioCLIP-2 \texttt{tol}      & .765 & .864 & .777 & .886 & .653 & .718 & .182 & .921 & .817 & .793 \\
    \midrule
    Qwen3-VL-30B-A3B            & \bst{.495} & \snd{.561} & \snd{.497} & \bst{.691} & \snd{.322} & \bst{.345} & .482 & \snd{.605} & \bst{.614} & \snd{.406} \\
    Qwen3-VL-32B                & \snd{.485} & .536 & \bst{.525} & \snd{.653} & \bst{.328} & \snd{.334} & .388 & \bst{.645} & .567 & \bst{.407} \\
    Qwen3-VL-8B                 & .460 & \bst{.595} & .478 & .645 & .285 & .327 & .469 & .601 & \snd{.574} & .385 \\
    InternVL3-38B               & .457 & .480 & .444 & .640 & .265 & .297 & .483 & .573 & .552 & .346 \\
    Qwen2.5-VL-7B               & .425 & .454 & .459 & .636 & .264 & .278 & \snd{.488} & .573 & .506 & .362 \\
    Gemma-3-12B                 & .420 & .367 & .382 & .539 & .256 & .229 & .470 & .487 & .544 & .310 \\
    InternVL3.5-38B             & .400 & .332 & .408 & .555 & .239 & .205 & .400 & .528 & .521 & .306 \\
    Gemma-4-12B                 & .361 & .453 & .411 & .474 & .211 & .194 & \bst{.516} & .468 & .478 & .263 \\
    InternVL3-8B                & .385 & .310 & .403 & .497 & .183 & .206 & .320 & .480 & .448 & .309 \\
    Qwen2.5-VL-3B               & .334 & .361 & .421 & .512 & .255 & .212 & .298 & .522 & .357 & .212 \\
    Llama-3.2-Vision            & .324 & .280 & .430 & .544 & .148 & .154 & .310 & .484 & .466 & .225 \\
    InternVL3.5-8B              & .368 & .293 & .359 & .474 & .156 & .195 & .266 & .445 & .447 & .243 \\
    Molmo2-8B                   & .387 & .297 & .396 & .463 & .156 & .192 & .190 & .437 & .468 & .241 \\
    Pixtral-12B                 & .312 & .289 & .395 & .480 & .196 & .219 & .280 & .460 & .442 & .207 \\
    InternVL3.5-30B-A3B         & .328 & .290 & .400 & .420 & .168 & .221 & .274 & .471 & .357 & .234 \\
    Phi-3.5-Vision              & \wst{.199} & .206 & .371 & \wst{.319} & \swst{.061} & \swst{.148} & .198 & .338 & .345 & \swst{.156} \\
    LLaVA-NeXT-Mistral          & \swst{.203} & \swst{.104} & \wst{.354} & .342 & \wst{.047} & .151 & \swst{.116} & \wst{.290} & \swst{.312} & \wst{.096} \\
    LLaVA-NeXT-Vicuna           & .215 & \wst{.066} & \swst{.356} & \swst{.329} & .073 & \wst{.144} & \wst{.071} & \swst{.304} & \wst{.295} & .160 \\
    \bottomrule
  \end{tabular}
  \caption{$\EM$ by organismal group for every configuration. Birds and Insects are the easiest groups for almost every model, Bryophytes the hardest. Gymnosperms is the one column where the specialist loses to most generalists, which is the same dissociation \cref{tab:supp-c1groupcorr} reports.}
  \label{tab:supp-t-group}
\end{table*}

\begin{table}[htbp]
  \centering\footnotesize
  \setlength{\tabcolsep}{4pt}
  \begin{tabular}{@{}lrrrr@{}}
    \toprule
    Model & Easy & Medium & Hard & Degen. \\
    \midrule
    \rowcolor{rowtint}
    BioCLIP-2 \texttt{kingdom}  & .800 & .698 & .616 & .985 \\
    \rowcolor{rowtint}
    BioCLIP-2 \texttt{tol}      & .798 & .699 & .619 & .925 \\
    \midrule
    Qwen3-VL-30B-A3B            & \snd{.606} & \bst{.429} & \bst{.295} & .864 \\
    Qwen3-VL-32B                & \bst{.613} & \snd{.428} & .245 & .831 \\
    Qwen3-VL-8B                 & .592 & .412 & \snd{.246} & \snd{.887} \\
    InternVL3-38B               & .562 & .375 & .207 & \bst{.911} \\
    Qwen2.5-VL-7B               & .542 & .408 & .227 & .763 \\
    Gemma-3-12B                 & .499 & .330 & .185 & .790 \\
    InternVL3.5-38B             & .501 & .331 & .160 & .746 \\
    Gemma-4-12B                 & .468 & .307 & .148 & .880 \\
    InternVL3-8B                & .444 & .300 & .098 & .827 \\
    Qwen2.5-VL-3B               & .489 & .332 & .161 & \swst{.416} \\
    Llama-3.2-Vision            & .463 & .297 & .131 & .545 \\
    InternVL3.5-8B              & .423 & .244 & .077 & .819 \\
    Molmo2-8B                   & .432 & .240 & .061 & .823 \\
    Pixtral-12B                 & .432 & .270 & .107 & .688 \\
    InternVL3.5-30B-A3B         & .456 & .280 & .106 & .486 \\
    Phi-3.5-Vision              & \swst{.284} & \swst{.152} & .041 & .748 \\
    LLaVA-NeXT-Mistral          & .300 & .174 & \swst{.039} & \wst{.364} \\
    LLaVA-NeXT-Vicuna           & \wst{.245} & \wst{.128} & \wst{.029} & .662 \\
    \bottomrule
  \end{tabular}
  \caption{$\EM$ by difficulty tier for every configuration. All 20 satisfy Easy $>$ Medium $>$ Hard without exception. The Degenerate column holds items in cells too thin to tier, described in \cref{sec:supp-tiers}.}
  \label{tab:supp-t-diff}
\end{table}

\begin{table}[htbp]
  \centering\footnotesize
  \setlength{\tabcolsep}{4pt}
  \begin{tabular}{@{}lrrr@{}}
    \toprule
    Rank & Easy & Medium & Hard \\
    \midrule
    class             & .796 & .691 & .233 \\
    order             & .786 & .605 & .299 \\
    family            & .580 & .387 & .270 \\
    genus             & .444 & .265 & .165 \\
    species epithet   & .101 & .083 & .074 \\
    scientific name   & .236 & .148 & .116 \\
    common name       & .113 & .075 & .065 \\
    common name MCQ   & .705 & .618 & .573 \\
    \bottomrule
  \end{tabular}
  \caption{Mean $\EM$ by rank and difficulty tier, pooled over all 20 configurations. The tier gap is widest at class and order and narrowest at the species epithet, where every tier is already near the floor. Phylum carries no tier, because a single group holds too few phyla to cut tertiles.}
  \label{tab:supp-t-rankdiff}
\end{table}

\begin{table}[htbp]
  \centering\footnotesize
  \setlength{\tabcolsep}{4pt}
  \begin{tabular}{@{}lrrr@{}}
    \toprule
    Model & open-ended & 4-way MCQ & gap \\
    \midrule
    Qwen3-VL-32B                & .078 & .809 & $+$.731 \\
    Qwen3-VL-8B                 & .065 & .775 & $+$.710 \\
    InternVL3-38B               & .044 & .752 & $+$.708 \\
    Qwen2.5-VL-7B               & .062 & .753 & $+$.691 \\
    InternVL3.5-38B             & .025 & .707 & $+$.682 \\
    Qwen3-VL-30B-A3B            & .107 & .772 & $+$.665 \\
    Gemma-3-12B                 & .047 & .687 & $+$.640 \\
    InternVL3.5-30B-A3B         & .024 & .653 & $+$.629 \\
    Qwen2.5-VL-3B               & .064 & .689 & $+$.625 \\
    InternVL3-8B                & .020 & .643 & $+$.623 \\
    Gemma-4-12B                 & .043 & .657 & $+$.614 \\
    Llama-3.2-Vision            & .054 & .655 & $+$.601 \\
    InternVL3.5-8B              & .015 & .595 & $+$.580 \\
    Pixtral-12B                 & .027 & .592 & $+$.565 \\
    Molmo2-8B                   & .016 & .574 & $+$.558 \\
    \rowcolor{rowtint}
    BioCLIP-2 \texttt{tol}      & .500 & .966 & $+$.466 \\
    \rowcolor{rowtint}
    BioCLIP-2 \texttt{kingdom}  & .502 & .966 & $+$.464 \\
    LLaVA-NeXT-Mistral          & .004 & .465 & $+$.461 \\
    Phi-3.5-Vision              & .009 & .443 & $+$.434 \\
    LLaVA-NeXT-Vicuna           & .003 & .382 & $+$.379 \\
    \bottomrule
  \end{tabular}
  \caption{Open-ended against four-way multiple choice on the common name, all 20 configurations, sorted by gap. Every configuration recognizes far better than it recalls, including both specialists. \cref{tab:supp-c1recog} gives the five widest and the two means.}
  \label{tab:supp-t-recog}
\end{table}

\begin{table}[htbp]
  \centering\footnotesize
  \setlength{\tabcolsep}{4pt}
  \begin{tabular}{@{}lrrr@{}}
    \toprule
    Model & unresolved & severity & $\EM$ \\
    \midrule
    LLaVA-NeXT-Mistral          & .233 & 2.739 & .203 \\
    LLaVA-NeXT-Vicuna           & .220 & 3.135 & .202 \\
    Llama-3.2-Vision            & .217 & 2.493 & .338 \\
    Phi-3.5-Vision              & .179 & 2.992 & .235 \\
    Qwen2.5-VL-3B               & .177 & 2.221 & .350 \\
    InternVL3-8B                & .160 & 2.692 & .355 \\
    InternVL3.5-30B-A3B         & .135 & 2.667 & .317 \\
    Molmo2-8B                   & .122 & 2.746 & .324 \\
    Pixtral-12B                 & .121 & 2.775 & .329 \\
    Qwen2.5-VL-7B               & .110 & 2.444 & .445 \\
    InternVL3.5-38B             & .108 & 2.606 & .390 \\
    Gemma-4-12B                 & .107 & 2.763 & .384 \\
    Qwen3-VL-32B                & .102 & 2.459 & .488 \\
    Gemma-3-12B                 & .102 & 2.574 & .401 \\
    InternVL3.5-8B              & .100 & 2.740 & .325 \\
    Qwen3-VL-8B                 & .088 & 2.480 & .483 \\
    InternVL3-38B               & .083 & 2.614 & .455 \\
    Qwen3-VL-30B-A3B            & .081 & 2.472 & .503 \\
    \rowcolor{rowtint}
    BioCLIP-2 \texttt{tol}      & .026 & 2.638 & .737 \\
    \rowcolor{rowtint}
    BioCLIP-2 \texttt{kingdom}  & .025 & 2.523 & .743 \\
    \bottomrule
  \end{tabular}
  \caption{Unresolved share, severity and $\EM$ for every configuration, sorted by unresolved. An unresolved answer names no node in the tree, so it carries no severity. The share correlates with $\EM$ at $\rho = -0.854$.}
  \label{tab:supp-t-unres}
\end{table}

\begin{table}[htbp]
  \centering\footnotesize
  \setlength{\tabcolsep}{4pt}
  \begin{tabular}{@{}lrrrrr@{}}
    \toprule
    Model & $\EM$ & sev & $\EM$ rk & sev rk & shift \\
    \midrule
    \rowcolor{rowtint}
    BioCLIP-2 \texttt{kingdom}  & .743 & 2.523 & 1 & 7 & $-6$ \\
    \rowcolor{rowtint}
    BioCLIP-2 \texttt{tol}      & .737 & 2.638 & 2 & 11 & $-9$ \\
    Qwen3-VL-30B-A3B            & .503 & 2.472 & 3 & 4 & $-1$ \\
    Qwen3-VL-32B                & .488 & 2.459 & 4 & 3 & $+1$ \\
    Qwen3-VL-8B                 & .483 & 2.480 & 5 & 5 & $+0$ \\
    InternVL3-38B               & .455 & 2.614 & 6 & 10 & $-4$ \\
    Qwen2.5-VL-7B               & .445 & 2.444 & 7 & 2 & $+5$ \\
    Gemma-3-12B                 & .401 & 2.574 & 8 & 8 & $+0$ \\
    InternVL3.5-38B             & .390 & 2.606 & 9 & 9 & $+0$ \\
    Gemma-4-12B                 & .384 & 2.763 & 10 & 17 & $-7$ \\
    InternVL3-8B                & .355 & 2.692 & 11 & 13 & $-2$ \\
    Qwen2.5-VL-3B               & .350 & 2.221 & 12 & 1 & $+11$ \\
    Llama-3.2-Vision            & .338 & 2.493 & 13 & 6 & $+7$ \\
    Pixtral-12B                 & .329 & 2.775 & 14 & 18 & $-4$ \\
    InternVL3.5-8B              & .325 & 2.740 & 15 & 15 & $+0$ \\
    Molmo2-8B                   & .324 & 2.746 & 16 & 16 & $+0$ \\
    InternVL3.5-30B-A3B         & .317 & 2.667 & 17 & 12 & $+5$ \\
    Phi-3.5-Vision              & .235 & 2.992 & 18 & 19 & $-1$ \\
    LLaVA-NeXT-Mistral          & .203 & 2.739 & 19 & 14 & $+5$ \\
    LLaVA-NeXT-Vicuna           & .202 & 3.135 & 20 & 20 & $+0$ \\
    \bottomrule
  \end{tabular}
  \caption{$\EM$ rank against severity rank for every configuration. A positive shift means the model ranks better on severity than on accuracy. Qwen2.5-VL-3B moves 11 places and is the clearest case of a model that is inaccurate and still misses closely.}
  \label{tab:supp-t-dissoc}
\end{table}

\begin{table}[htbp]
  \centering\footnotesize
  \setlength{\tabcolsep}{4pt}
  \begin{tabular}{@{}llrrr@{}}
    \toprule
    Model & Family & Params & $\EM$ & sev \\
    \midrule
    Qwen2.5-VL-3B               & Qwen2.5-VL  &   3.0 & .350 & 2.221 \\
    Phi-3.5-Vision              & Phi-3.5     &   4.2 & .235 & 2.992 \\
    Qwen2.5-VL-7B               & Qwen2.5-VL  &   7.0 & .445 & 2.444 \\
    LLaVA-NeXT-Mistral          & LLaVA-1.6   &   7.0 & .203 & 2.739 \\
    LLaVA-NeXT-Vicuna           & LLaVA-1.6   &   7.0 & .202 & 3.135 \\
    InternVL3.5-8B              & InternVL3.5 &   8.0 & .325 & 2.740 \\
    InternVL3-8B                & InternVL3   &   8.0 & .355 & 2.692 \\
    Qwen3-VL-8B                 & Qwen3-VL    &   8.0 & .483 & 2.480 \\
    Molmo2-8B                   & Molmo2      &   8.0 & .324 & 2.746 \\
    Llama-3.2-Vision            & Llama-3.2   &  11.0 & .338 & 2.493 \\
    Gemma-4-12B                 & Gemma-4     &  12.0 & .384 & 2.763 \\
    Gemma-3-12B                 & Gemma-3     &  12.0 & .401 & 2.574 \\
    Pixtral-12B                 & Pixtral     &  12.0 & .329 & 2.775 \\
    Qwen3-VL-30B-A3B            & Qwen3-VL    &  30.0 & .503 & 2.472 \\
    InternVL3.5-30B-A3B         & InternVL3.5 &  30.0 & .317 & 2.667 \\
    Qwen3-VL-32B                & Qwen3-VL    &  32.0 & .488 & 2.459 \\
    InternVL3-38B               & InternVL3   &  38.0 & .455 & 2.614 \\
    InternVL3.5-38B             & InternVL3.5 &  38.0 & .390 & 2.606 \\
    \bottomrule
  \end{tabular}
  \caption{Parameters against pooled $\EM$ and severity for the 18 generalists, sorted by size. Within Qwen3-VL, 8B scores $.483$ and 32B scores $.488$. Within InternVL3, 8B scores $.355$ and 38B scores $.455$. Scale helps one family and not the other, and it never closes the gap to the specialist.}
  \label{tab:supp-t-scale}
\end{table}

\begin{table}[htbp]
  \centering\footnotesize
  \setlength{\tabcolsep}{4pt}
  \begin{tabular}{@{}lrl@{}}
    \toprule
    Metric & $\rho$ with $\EM$ & reading \\
    \midrule
    Wu--Palmer  & $+0.959$ & near-redundant with $\EM$ \\
    Lin         & $+0.965$ & near-redundant with $\EM$ \\
    severity    & $-0.662$ & orders the field differently \\
    unresolved  & $-0.854$ & tracks $\EM$, reported beside it \\
    \bottomrule
  \end{tabular}
  \caption{Spearman correlation of each \ctax{} metric with $\EM$ over the 20 configurations. Wu--Palmer and Lin reorder almost nothing, so we keep them as corroboration. Severity is the one metric that carries information $\EM$ does not.}
  \label{tab:supp-t-redund}
\end{table}

\begin{table*}[tbp]
  \centering\scriptsize
  \setlength{\tabcolsep}{3.5pt}
  \begin{tabular}{@{}lrrrrrrrrrr@{}}
    \toprule
    Model & attr\_bare & attr\_group & cat\_mcq & marker & common & genus & kind\_of & sci & vis\_set & vis\_side \\
    \midrule
    Qwen3-VL-32B          & 1.003 & 1.015 & \bst{1.000} & .968 & \snd{.503} & .995 & \bst{.374} & \snd{.769} & .857 & 1.069 \\
    Qwen3-VL-30B-A3B      & 1.001 & .993 & \bst{1.000} & .919 & .522 & \bst{.977} & .398 & .775 & .833 & 1.034 \\
    InternVL3-38B         & .998 & 1.008 & \bst{1.000} & .983 & .592 & .990 & .442 & .777 & .814 & 1.031 \\
    Gemma-4-12B           & \wst{1.031} & \swst{1.032} & \bst{1.000} & \wst{1.078} & .790 & 1.005 & .470 & .939 & .763 & \swst{1.087} \\
    Qwen3-VL-8B           & 1.010 & 1.006 & \bst{1.000} & .940 & .555 & \snd{.985} & .411 & \bst{.765} & .861 & 1.049 \\
    InternVL3.5-38B       & \swst{1.019} & 1.012 & \bst{1.000} & .951 & .695 & .987 & .593 & .832 & .881 & 1.071 \\
    Molmo2-8B             & 1.007 & 1.015 & \bst{1.000} & .949 & .761 & 1.003 & .473 & .921 & .967 & \snd{.878} \\
    Qwen2.5-VL-7B         & 1.006 & 1.017 & \bst{1.000} & .916 & \bst{.495} & .990 & .393 & .803 & .809 & 1.022 \\
    InternVL3.5-30B-A3B   & .989 & 1.017 & \bst{1.000} & .986 & .770 & .992 & .475 & .850 & .878 & 1.035 \\
    InternVL3-8B          & .994 & \wst{1.043} & \bst{1.000} & .991 & .702 & .987 & .441 & .811 & .858 & 1.009 \\
    InternVL3.5-8B        & .976 & 1.002 & \bst{1.000} & .905 & .779 & 1.004 & .495 & .866 & \bst{.756} & 1.055 \\
    Gemma-3-12B           & .993 & .996 & \bst{1.000} & \snd{.899} & .649 & 1.001 & \snd{.392} & .874 & .999 & .936 \\
    Qwen2.5-VL-3B         & \snd{.968} & \snd{.985} & \bst{1.000} & .981 & .557 & 1.001 & .453 & .780 & 1.022 & 1.000 \\
    Pixtral-12B           & .994 & 1.024 & \bst{1.000} & \bst{.889} & .652 & .992 & .522 & .903 & .820 & \wst{1.150} \\
    LLaVA-NeXT-Vicuna     & .999 & \bst{.962} & \bst{1.000} & \swst{.993} & \wst{.868} & 1.003 & \swst{.796} & \swst{.969} & \snd{.759} & \bst{.721} \\
    Phi-3.5-Vision        & 1.004 & 1.013 & \bst{1.000} & .975 & .845 & \wst{1.007} & \wst{.918} & \wst{.979} & \wst{1.088} & 1.024 \\
    Llama-3.2-Vision      & .971 & 1.016 & \bst{1.000} & .926 & .575 & 1.004 & .602 & .869 & 1.048 & 1.003 \\
    LLaVA-NeXT-Mistral    & \bst{.959} & 1.026 & \bst{1.000} & .958 & \swst{.858} & \swst{1.007} & .701 & .957 & \swst{1.052} & 1.024 \\
    \midrule
    \textbf{mean}         & .996 & 1.010 & 1.000 & .956 & .676 & .996 & .519 & .858 & .893 & 1.011 \\
    \bottomrule
  \end{tabular}
  \caption{Mistake severity by question family, the mean $1-\cags{}$ over wrong items only. Lower is better. A value above 1 means the average miss lands farther from the gold than a random distractor. \cref{tab:supp-c2lb} gives the pooled column. Colors mark the \bst{best}, \snd{second best}, \swst{second worst} and \wst{worst} in each family.}
  \label{tab:supp-t-c2sev}
\end{table*}

\begin{table*}[tbp]
  \centering\scriptsize
  \setlength{\tabcolsep}{3.5pt}
  \begin{tabular}{@{}lrrrrrrrrrr@{}}
    \toprule
    Model & attr\_bare & attr\_group & cat\_mcq & marker & common & genus & kind\_of & sci & vis\_set & vis\_side \\
    \midrule
    Qwen3-VL-32B          & \bst{.725} & \snd{.721} & \snd{.959} & \snd{.787} & .624 & \bst{.699} & \bst{.766} & \bst{.655} & \bst{.480} & \snd{.613} \\
    Qwen3-VL-30B-A3B      & .714 & .697 & .951 & .755 & .613 & .426 & .737 & .622 & \snd{.479} & .554 \\
    InternVL3-38B         & \snd{.724} & .704 & .948 & .721 & .501 & .638 & .670 & .641 & .423 & .572 \\
    Gemma-4-12B           & .710 & \bst{.728} & .772 & .756 & .294 & .131 & .621 & .168 & .432 & .575 \\
    Qwen3-VL-8B           & .706 & .681 & .958 & .732 & .549 & .489 & .695 & .584 & .475 & .546 \\
    InternVL3.5-38B       & .724 & .704 & .920 & .754 & .396 & .422 & .524 & .492 & .366 & .584 \\
    Molmo2-8B             & .721 & .660 & \wst{.532} & \bst{.823} & .291 & .154 & .583 & .189 & .419 & \bst{.807} \\
    Qwen2.5-VL-7B         & .670 & .688 & \bst{.962} & .720 & \bst{.631} & \snd{.639} & \snd{.739} & \snd{.642} & .383 & .507 \\
    InternVL3.5-30B-A3B   & .694 & .688 & .869 & .700 & .309 & .510 & .611 & .473 & .388 & .481 \\
    InternVL3-8B          & .673 & .672 & .884 & .689 & .381 & .527 & .647 & .533 & .389 & .480 \\
    InternVL3.5-8B        & .685 & .656 & .783 & .697 & .284 & .354 & .562 & .388 & .399 & .524 \\
    Gemma-3-12B           & .646 & .639 & .914 & .716 & .444 & .304 & .692 & .295 & .472 & .468 \\
    Qwen2.5-VL-3B         & .599 & .610 & .926 & \swst{.632} & .572 & .570 & .669 & .625 & .461 & .504 \\
    Pixtral-12B           & .597 & .632 & .847 & .690 & .402 & .266 & .618 & .288 & .383 & \wst{.386} \\
    LLaVA-NeXT-Vicuna     & \wst{.564} & .625 & \swst{.550} & \wst{.608} & \wst{.150} & \swst{.038} & \swst{.240} & \swst{.122} & .409 & .525 \\
    Phi-3.5-Vision        & .634 & .633 & .716 & .638 & .186 & .076 & \wst{.107} & \wst{.115} & \wst{.203} & \swst{.402} \\
    Llama-3.2-Vision      & \swst{.583} & \wst{.547} & .859 & .690 & \snd{.628} & .195 & .707 & .305 & .335 & .443 \\
    LLaVA-NeXT-Mistral    & .585 & \swst{.584} & .636 & .671 & \swst{.163} & \wst{.033} & .344 & .128 & \swst{.289} & .493 \\
    \midrule
    \textbf{mean}         & .664 & .659 & .833 & .710 & .412 & .360 & .585 & .404 & .399 & .526 \\
    \bottomrule
  \end{tabular}
  \caption{Raw graded similarity $\mathrm{GS}_{\text{raw}}$ by question family, before the chance adjustment. Reading this against \cref{tab:supp-c2fam} shows what the adjustment removes. The two are close on the open families, where distractors are sampled and $b$ is small, and far apart on \texttt{visibility\_partset} and \texttt{attr\_group\_value}, where the option sets sit close to the gold and a raw score flatters every model.}
  \label{tab:supp-t-c2gsraw}
\end{table*}

\begin{table}[htbp]
  \centering\footnotesize
  \setlength{\tabcolsep}{4pt}
  \begin{tabular}{@{}lrrr@{}}
    \toprule
    Model & easy & random & hard \\
    \midrule
    Qwen3-VL-32B          & $-$.270 & \snd{$-$.076} & $-$.382 \\
    InternVL3-8B          & $-$.652 & $-$.765 & $-$1.024 \\
    Qwen2.5-VL-3B         & $-$.843 & $-$1.264 & $-$1.440 \\
    InternVL3.5-30B-A3B   & \bst{$-$.003} & \bst{$-$.066} & \snd{$-$.263} \\
    Phi-3.5-Vision        & $-$.740 & $-$.885 & $-$1.133 \\
    Qwen3-VL-30B-A3B      & \snd{$-$.010} & .118 & \bst{$-$.052} \\
    Qwen3-VL-8B           & $-$.611 & $-$.776 & $-$.986 \\
    Molmo2-8B             & $-$.378 & $-$.660 & $-$.839 \\
    LLaVA-NeXT-Mistral    & \swst{$-$2.221} & \swst{$-$2.241} & \swst{$-$2.301} \\
    InternVL3-38B         & $-$.303 & $-$.494 & $-$.767 \\
    Qwen2.5-VL-7B         & $-$1.020 & $-$1.242 & $-$1.352 \\
    Gemma-4-12B           & $-$.257 & $-$.292 & $-$.434 \\
    InternVL3.5-38B       & $-$.397 & $-$.640 & $-$.759 \\
    InternVL3.5-8B        & $-$.170 & $-$.355 & $-$.401 \\
    Gemma-3-12B           & $-$1.257 & $-$1.818 & $-$1.767 \\
    Pixtral-12B           & $-$.570 & $-$.665 & $-$.696 \\
    Llama-3.2-Vision      & $-$.999 & $-$1.340 & $-$1.429 \\
    LLaVA-NeXT-Vicuna     & \wst{$-$3.540} & \wst{$-$3.540} & \wst{$-$3.540} \\
    \bottomrule
  \end{tabular}
  \caption{Criterion $c$ for every model in all three conditions, pooled over the five species by summing the confusion counts. Negative means a standing preference for \emph{yes}. Colors mark the smallest and largest $|c|$, so \bst{best} is the least biased. Every model leans toward \emph{yes}, and 17 of 18 lean further as the pairs harden.}
  \label{tab:supp-t-c4crit}
\end{table}

\begin{table*}[tbp]
  \centering\scriptsize
  \setlength{\tabcolsep}{5pt}
  \begin{tabular}{@{}lrrrrr@{}}
    \toprule
    Model & Whale shark & Hyena & Leopard & Grevy's zebra & Beluga \\
    \midrule
    Qwen3-VL-32B          & .496 & .814 & .517 & \snd{.931} & \bst{1.027} \\
    InternVL3-8B          & \bst{.611} & \bst{1.097} & \snd{.981} & .798 & \snd{.882} \\
    Qwen2.5-VL-3B         & .287 & 1.048 & \bst{1.243} & .203 & .470 \\
    InternVL3.5-30B-A3B   & .412 & .968 & .443 & .506 & .700 \\
    Phi-3.5-Vision        & .364 & .880 & .896 & .344 & .709 \\
    Qwen3-VL-30B-A3B      & .332 & .861 & .593 & \bst{1.088} & .061 \\
    Qwen3-VL-8B           & .352 & .780 & .417 & .516 & .821 \\
    Molmo2-8B             & .232 & .963 & .644 & .472 & .228 \\
    LLaVA-NeXT-Mistral    & .179 & \snd{1.052} & .714 & \swst{0.000} & .633 \\
    InternVL3-38B         & .234 & .729 & .564 & .519 & .337 \\
    Qwen2.5-VL-7B         & .381 & .709 & .323 & .338 & .332 \\
    Gemma-4-12B           & .399 & .564 & .387 & .670 & \swst{$-$.042} \\
    InternVL3.5-38B       & .126 & .287 & .695 & .409 & .335 \\
    InternVL3.5-8B        & .398 & .401 & \wst{$-$.075} & .204 & .294 \\
    Gemma-3-12B           & \snd{.501} & .143 & .413 & \wst{$-$.379} & .172 \\
    Pixtral-12B           & .077 & \swst{.074} & .141 & .127 & .441 \\
    Llama-3.2-Vision      & \wst{$-$.406} & .728 & .414 & .286 & \wst{$-$.499} \\
    LLaVA-NeXT-Vicuna     & \swst{0.000} & \wst{0.000} & \swst{0.000} & \swst{0.000} & 0.000 \\
    \bottomrule
  \end{tabular}
  \caption{$\dpr$ by model and species under random pairing, the neutral condition. \cref{tab:supp-c4species} gives the column means. No model reaches $\dpr = 1.5$ on any species, and beluga is the hardest column for almost every model.}
  \label{tab:supp-t-c4species}
\end{table*}

\begin{table}[htbp]
  \centering\footnotesize
  \setlength{\tabcolsep}{4pt}
  \begin{tabular}{@{}lrrrr@{}}
    \toprule
    Model & acc & bal.\ acc & MCC & $\kappa$ \\
    \midrule
    Qwen3-VL-32B          & \bst{.636} & \bst{.636} & \bst{.284} & \bst{.272} \\
    InternVL3-8B          & .601 & .601 & \snd{.256} & .202 \\
    Qwen2.5-VL-3B         & .559 & .559 & .176 & .119 \\
    InternVL3.5-30B-A3B   & \snd{.612} & \snd{.612} & .231 & \snd{.223} \\
    Phi-3.5-Vision        & .576 & .576 & .198 & .151 \\
    Qwen3-VL-30B-A3B      & .602 & .602 & .216 & .203 \\
    Qwen3-VL-8B           & .572 & .572 & .181 & .145 \\
    Molmo2-8B             & .578 & .578 & .179 & .157 \\
    LLaVA-NeXT-Mistral    & .508 & .508 & .080 & .016 \\
    InternVL3-38B         & .581 & .581 & .177 & .162 \\
    Qwen2.5-VL-7B         & .541 & .541 & .121 & .082 \\
    Gemma-4-12B           & .567 & .567 & .147 & .134 \\
    InternVL3.5-38B       & .542 & .542 & .107 & .084 \\
    InternVL3.5-8B        & .544 & .544 & .094 & .088 \\
    Gemma-3-12B           & \swst{.505} & \swst{.505} & \swst{.028} & .009 \\
    Pixtral-12B           & .519 & .519 & .054 & .038 \\
    Llama-3.2-Vision      & \wst{.500} & \wst{.500} & \wst{.019} & \swst{.000} \\
    LLaVA-NeXT-Vicuna     & \wst{.500} & \wst{.500} & nan & \wst{0.000} \\
    \bottomrule
  \end{tabular}
  \caption{Accuracy, balanced accuracy, Matthews correlation and Cohen's $\kappa$ under random pairing, averaged over the five species. We report $\dpr$ as the headline because these four all conflate discrimination with response bias, and \cref{tab:supp-t-c4crit} shows the bias is large. The four agree with each other and with $\dpr$ on the broad ordering.}
  \label{tab:supp-t-c4alt}
\end{table}

\begin{table}[htbp]
  \centering\footnotesize
  \setlength{\tabcolsep}{4pt}
  \begin{tabular}{@{}lrrr@{}}
    \toprule
    Model & easy & hard & easy $-$ hard \\
    \midrule
    InternVL3-8B          & 80.1 & 41.3 & \bst{38.8} \\
    InternVL3-38B         & 77.7 & 39.3 & \snd{38.4} \\
    InternVL3.5-30B-A3B   & 78.3 & 39.9 & 38.4 \\
    Phi-3.5-Vision        & 77.1 & 40.1 & 37.0 \\
    Qwen3-VL-32B          & 75.9 & 38.9 & 36.9 \\
    Molmo2-8B             & 80.2 & 44.2 & 36.0 \\
    Qwen2.5-VL-3B         & 79.8 & 44.5 & 35.3 \\
    Qwen3-VL-8B           & 75.3 & 42.8 & 32.5 \\
    Gemma-4-12B           & 73.6 & 42.5 & 31.1 \\
    InternVL3.5-38B       & 72.3 & 43.1 & 29.1 \\
    Qwen3-VL-30B-A3B      & 73.9 & 45.1 & 28.8 \\
    InternVL3.5-8B        & 67.2 & 44.7 & 22.5 \\
    Qwen2.5-VL-7B         & 65.8 & 46.8 & 19.0 \\
    Pixtral-12B           & 60.9 & 46.2 & 14.7 \\
    Llama-3.2-Vision      & 60.5 & 48.5 & 12.0 \\
    Gemma-3-12B           & 58.7 & 49.6 & 9.1 \\
    LLaVA-NeXT-Mistral    & 54.4 & 49.2 & \swst{5.2} \\
    LLaVA-NeXT-Vicuna     & 50.0 & 50.0 & \wst{0.0} \\
    \bottomrule
  \end{tabular}
  \caption{The difficulty manipulation for every model, in percentage points of accuracy averaged over the five species. Seventeen models confirm the ordering, one is flat, and none reverses. The flat model emits a constant answer and cannot respond to any manipulation, which is what makes it a useful control.}
  \label{tab:supp-t-c4manip}
\end{table}

\end{document}